\documentclass{article}

\PassOptionsToPackage{sort, numbers, compress}{natbib}

\usepackage[main, final]{neurips_2026}

\usepackage[utf8]{inputenc} 
\usepackage[T1]{fontenc}    
\usepackage{hyperref}       
\hypersetup{breaklinks=true,colorlinks,urlcolor={red!90!black},linkcolor={red!90!black},citecolor={blue!90!black}}

\usepackage{url}            
\usepackage{booktabs}       
\usepackage{amsfonts}       
\usepackage{nicefrac}       
\usepackage{microtype}      
\usepackage{xcolor}         

\usepackage{algorithm}
\usepackage{algpseudocode}
\usepackage{amsmath}
\usepackage{graphicx}
\usepackage{makecell}
\usepackage{multirow}
\usepackage{placeins}
\usepackage{subcaption}
\usepackage{wrapfig}
\usepackage[table]{xcolor}
\newcommand{\algphase}[1]{%
    \Statex
    \hspace{-0.25em}%
    {\color{blue}\ttfamily\bfseries\# #1}%
}

\title{NEAT: Neighborhood-Guided, Efficient, Autoregressive Set Transformer\\for 3D Molecular Generation}

\author{
  Daniel~Rose$^{1,2,3}$\thanks{Equal contribution, correspondence to: \texttt{\{first.middle.last\}@univie.ac.at}}
  \quad 
  Roxane~Axel~Jacob$^{1,2,3}$\footnotemark[1]
  \quad
  Johannes~Kirchmair$^{1,2}$
  \quad
  Thierry~Langer$^{1,2}$\\
  $^{1}$Department of Pharmaceutical Sciences, Faculty of Life Sciences, University of Vienna\\
  $^{2}$Christian Doppler Laboratory for Molecular Informatics in the Biosciences, \\Department of Pharmaceutical Sciences, University of Vienna\\
  $^{3}$Vienna Doctoral School of Pharmaceutical, Nutritional and Sport Sciences, University of Vienna\\
}

\begin{document}

\maketitle

\begin{abstract}
    Transformer-based autoregressive models offer an efficient alternative to diffusion- and flow-based approaches for generating 3D molecules. 
One challenge remains: standard transformer architectures require a sequential ordering of tokens, which is not inherently defined for the atoms in a molecule. 
Prior works have addressed this by using canonical atom orderings. 
However, these approaches are not permutation invariant w.r.t. atoms and bias next-token prediction towards ordering conventions. 
We overcome this limitation by introducing a novel neighborhood-guided training strategy. 
Our model, NEAT (\underline{N}eighborhood-Guided, \underline{E}fficient, \underline{A}utoregressive Set \underline{T}ransformer) treats molecular graphs as sets of atoms and learns an order-agnostic distribution over admissible tokens at the graph boundary, thereby ensuring atom-level permutation invariance. 
NEAT achieves state-of-the-art generation quality on the QM9 and GEOM-Drugs datasets, enables seamless completion of molecular prefixes, and offers a significant speed advantage over existing baselines.
\end{abstract}

\section{Introduction}

Generative modeling offers a promising alternative to virtual screening by enabling the construction of new molecules based on user-defined constraints.
Molecular generation can be approached through the generation of (1D) SMILES strings \citep{GomezBombarelli2018CVAE, Kusner2017GVAE}, (2D) molecular graphs \citep{Simonovsky2018GraphVAE, DeCao2022MolGAN}, (3D) atomic point clouds \citep{Gebauer2019GSchnNet, Simm2020, simm2021symmetryaware, Hoogeboom2022EDM, Luo2022GSphereNet, Xu2023GeoLDM, Daigavane2024Symphony, Zhang2025SymDiff, Cheng2025Quetzal}, or a combination thereof \citep{Hua2023MuDiff, Peng2023MolDiff, Vignac2023MiDi}. 
While modeling 3D structures is challenging, the additional spatial information is useful for many downstream applications.
In this work, we focus on \textit{unconditional} 3D generation, which is a necessary prerequisite to condition the generation process towards compounds with specific biological functions later on \citep{peng2022pocket2mol, schneuing2024structure}.

State-of-the-art 3D molecular generation models leverage diffusion and flow matching techniques \citep{Hoogeboom2022EDM, Xu2023GeoLDM, Song2023EquiFM, Song2024GeoBFN, Zhang2025SymDiff, Feng2025UniGEM, Dunn2025FlowMol}. 
At each integration step, these methods rely on  quadratically scaling message-passing updates to capture atom interdependencies. 
Furthermore, most of these models are not inherently adaptive to molecule size, as the number of atoms is typically fixed by the initial nodes sampled from the noise distribution. 
Trans-dimensional jump-diffusion \citep{campbell2023trans} enables atom insertions during denoising, but managing jump points in the trajectory is challenging. 
The use of virtual atoms \citep{Schneuing2025, Dunn2025FlowMol} allows dynamic masking of atoms during denoising, yet the maximum number of atoms in the generated molecule remains constrained by the initialization.

Autoregressive models offer a flexible and efficient alternative. 
They naturally support variable atom counts and require only one encoder call per generated token \citep{Li2024AutoregressiveImageGenerationWithoutVQ}.
For 3D autoregressive molecular generation, there currently exist three lines of work: (1) using graph neural networks (GNNs) in combination with focus atoms \citep{Gebauer2019GSchnNet, Luo2022GSphereNet, Daigavane2024Symphony}; (2) using standard large language models (LLMs) for generating XYZ, CIF or PDB files \citep{flamshepherd2023llm2mol} or SE(3)-invariant 1D tokens \citep{Li2025Geo2Seq, Fu2025Frag2Seq, Lu2025Uni-3DAR}; (3) combining a causal transformer with a diffusion loss to generate continuous tokens (atomic coordinates) \citep{Cheng2025Quetzal, Li2025InertialAR}.
Although current GNN-based methods ensure permutation invariance w.r.t. atom ordering, they generate molecules with low chemical validity. 
Conversely, sequence-based methods perform well in terms of validity but are not permutation-invariant w.r.t. atom ordering.

\begin{wrapfigure}{r}{0.50\textwidth}
  \vspace{-17pt}
  \centering
    \includegraphics[width=0.48\textwidth]{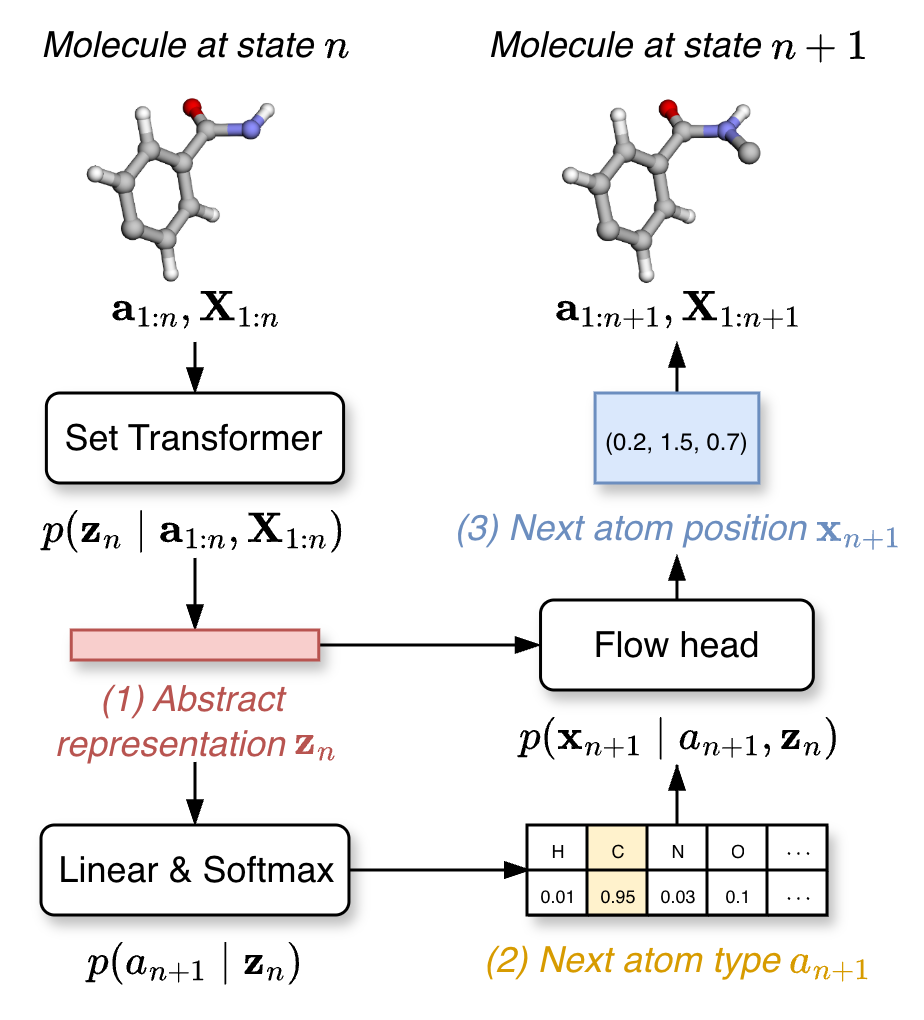}
    \caption{Autoregressive generation of 3D molecules with NEAT. (1)~A set transformer produces an abstract representation $\mathbf{z}_{n}$ of the molecule at state $n$, given the atom types $\mathbf{a}_{1:n}$ and the coordinates $\mathbf{X}_{1:n}$. (2)~The next atom type $a_{n+1}$ is inferred from $\mathbf{z}_{n}$. (3)~The next atom position $\mathbf{x}_{n+1}$ is modeled \textit{via} a flow head, which concludes the molecule update at state $n+1$.}
    \label{fig:thumbnail}
  \vspace{-15pt}
\end{wrapfigure}

In this work, we present the first autoregressive transformer model for 3D molecular generation that respects the atom permutation-invariance of molecular graphs. 
Working without canonical atom orderings introduces two challenges. 
First, it precludes the use of sequence embeddings, which are an essential component of standard autoregressive models. 
Second, it changes the autoregressive training set-up from supervision with a single token per step to supervision with a set of valid tokens: without a canonical atom order, all atoms in the 1-hop neighborhood of the current graph are valid continuations. 

Our solution to these challenges is simple yet effective. 
Given a partial molecular graph, a permutation-invariant set transformer \cite{Lee2019SetTransformer} is trained to produce a token distribution that matches the boundary nodes of the subgraph (\textit{i.e.}, all 1-hop neighbors of existing nodes). 
This training signal, which we refer to as neighborhood guidance, yields a probabilistic next-token distribution without relying on \textit{ad hoc} canonicalization schemes. 
We combine it with a flow matching loss to model continuous atomic coordinates.
Our approach enables training a standard transformer architecture for 3D molecular generation while maintaining permutation invariance to atom ordering.

NEAT outperforms prior autoregressive models and matches state-of-the-art flow models on the QM9 and GEOM-Drug benchmarks, while sampling $5\times$ faster than SemlaFlow \cite{Irwin2025} and $16\times$ faster than FlowMol3 \cite{Dunn2025FlowMol}. 
Its autoregressive design enables molecular prefix completion with no changes to the model or inference procedure---a task diffusion and flow models cannot perform out of the box. 
By establishing order-agnostic autoregressive generation as a competitive alternative to the SE(3)-equivariant diffusion and flow models that dominate the field, NEAT widens the space of viable approaches for 3D molecule generation. 
We hope to encourage the community to adopt and refine this direction for drug design tasks such as hit identification and lead optimization.
\section{Related works}

Autoregressive models generate sequences of tokens by conditioning each token on its predecessors. 
Formally, given a token sequence $x = (x_1, \dots, x_N)$, the joint distribution $p(x)$ is factorized as:  
\begin{equation}
    p(x) = p(x_1)\prod_{n=1}^{N-1} p(x_{n+1} \mid x_{1:n}),
\end{equation}  
where $x_{1:n} = (x_1, \dots, x_n)$ represents the tokens up to step $n$.
Sampling is performed by initializing with $x_1$ and generating tokens sequentially until $x_N$.
This is straightforward for sequentially ordered data such as text or images.
In the graph domain, the generation order is less obvious, and has been an active focus of research.
One way to model graphs as sequences is through canonicalization.
However, predefined canonical orderings impose limitations, as the optimal generation order is often context-dependent.  
To alleviate strict assumptions, \citet{you2018graphrnn} proposed a breadth-first search node ordering for autoregressive graph generation, while \citet{wang2025learningorder} introduced order policies to dynamically assign generation orders based on the current state. 
Although these approaches reduce reliance on fixed orderings, they do not fully eliminate order dependence.  
3D molecules are geometric graphs whose structure is determined by connectivity and geometry. 
The generation of valid atom positions poses additional constraints on the graph generation process and recent work on this specific task is discussed in the following. 

\paragraph{GNN models with focus atoms}
G-SchNet \citep{Gebauer2019GSchnNet} is an $E(3)$-invariant GNN that uses the concept of focus atoms for generation order. 
In each generation step, the model selects a focus atom as the center of a 3D grid, chooses the next atom type, and infers its position probability from radial distances to previously placed atoms. 
G-SphereNet \citep{Luo2022GSphereNet} adopts this approach but places atoms relative to the focus atom by generating atom distances, bond angles, and torsion angles with a normalizing flow.  
The $E(3)$-equivariant Symphony model \citep{Daigavane2024Symphony} parameterizes the 3D probability around the focus atom using spherical harmonic projections. 
While all these models respect the permutation-invariance of atoms, they produce molecules with comparably low chemical validity.

\paragraph{Transformer models with discrete tokenization}
Transformers can be used for 3D molecule generation when combined with a suitable text format.  
\citet{flamshepherd2023llm2mol} trained an LLM for 3D molecular generation directly on XYZ, CIF, or PDB files.
An active line of research are tokenization schemata to allow LLMs more fine-grained control in the generation process.
Geo2Seq \citep{Li2025Geo2Seq} converts molecular geometries into $SE(3)$-invariant 1D discrete sequences for unconditional modeling, while Frag2Seq \citep{Fu2025Frag2Seq} uses a related technique for structure-based modeling. 
Uni-3DAR \citep{Lu2025Uni-3DAR} employs a voxel-based discretization scheme with an octree-based compression technique, yielding sparse sequences.

\paragraph{Transformer models for continuous token generation}
While standard transformer architectures were initially developed for categorical data, they can be adapted for the generation of continuous values. 
\citet{Li2024AutoregressiveImageGenerationWithoutVQ} modeled the next-token distribution $p(x_{n+1} \mid x_{1:n})$ without vector quantization using a diffusion-based approach, factorized as:  
\begin{equation}
    p(x_{n+1} \mid x_{1:n}) = p(x_{n+1} \mid \mathbf{z}_n) \cdot p(\mathbf{z}_n \mid x_{1:n}),
\end{equation}  
where $\mathbf{z}_n \in \mathbb{R}^k$ represents an abstract latent variable for the next token. 
This factorization enables to use a transformer to model $p(\mathbf{z}_n \mid x_{1:n})$, capturing interdependencies among tokens, followed by a lightweight diffusion-based multilayer perceptron (MLP) to map $\mathbf{z}_n$ to the continuous token space. 
Their approach enabled high-quality autoregressive image generation with improved inference speed. 
\citet{Cheng2025Quetzal} applied this idea to the generation of atomic positions, enabling autoregressive 3D molecular generation without spatial discretization.
They employed a causal transformer with sequential embeddings derived from the atom order in XYZ coordinate files to predict discrete atom types, and a diffusion model, conditioned on these predictions, to predict the 3D atom positions.

\paragraph{Atom order matters to causal transformer models}
\citet{Cheng2025Quetzal} acknowledged that a limitation of a causal transformer formulation to 3D molecular generation is its sensitivity to atom ordering. 
They showed that replacing XYZ orderings with random atom-order augmentation during training substantially degrades model performance. 
This might seem surprising, since random token-order augmentation has been proposed as a way to address permutation invariance in autoregressive transformers for image generation \cite{Yu2025RandomAugmentations}. 
In 3D molecular generation, however, this strategy has not yielded satisfactory results, motivating the need for an order-agnostic approach. 

\paragraph{Our solution: an order-agnostic set transformer with neighborhood-guidance}
We propose that a set transformer \citep{Lee2019SetTransformer} with bidirectional attention is better suited to autoregressive 3D molecular generation, as it naturally respects the permutation invariance of atom sets. 
However, this induces a new problem: \emph{Without sequence information, how do we define the next token for model supervision?}
Our neighborhood-guidance strategy addresses it by introducing a chemical inductive bias: the model is supervised to generate atoms that form bonds with the current partial molecule, encouraging connectedness throughout generation. 
\section{Methods}\label{sec:Methods}

\begin{figure*}[t]
    \centering
    \includegraphics[width=1.\linewidth]{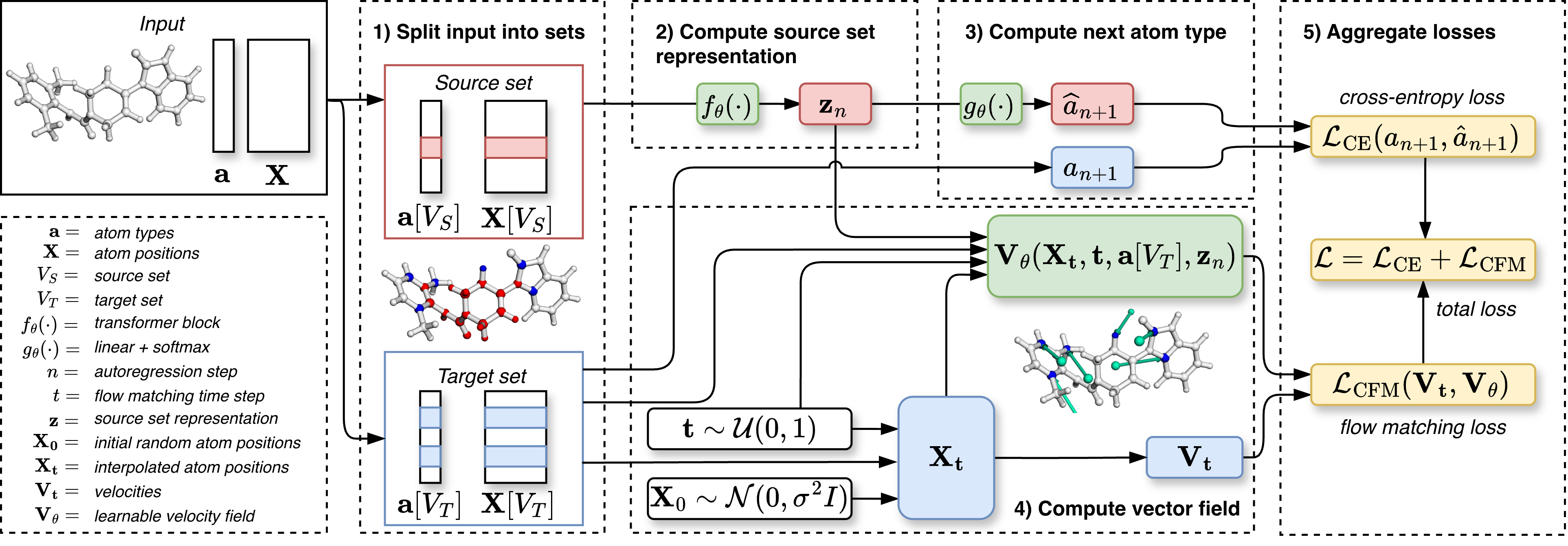}
    \caption[]{Overview of NEAT's training workflow through neighborhood guidance. The model takes molecular graphs as input. (1) For each training sample, we randomly construct a connected subgraph \textit{via} Algorithm~\ref{alg:source_target}. The selected nodes form the source set, and their neighboring boundary nodes define the target set. (2) The source set is encoded using a set transformer, (3) and its representation is used to predict the distribution over the next atom type. (4) For each target-set node, we sample a position from a normal distribution and construct linear interpolation paths between the sampled and true positions. These interpolation paths provide supervision for learning the vector field \textit{via} flow matching.  (5) The cross-entropy and flow matching losses are summed to yield the total loss.}
    \label{fig:overview_training}
\end{figure*}

\paragraph{Overview}
Our architecture consists of three main components: 
(1) a data-augmentation module for neighborhood-guided training, (2) a set transformer trunk that produces permutation-invariant encodings of atom sets, and (3) a flow matching head that generates continuous atomic coordinates.
During training, we split molecules into a \emph{source set}, representing the state of the molecular graph at step $n$, and a \emph{target set}, representing the set of possible token continuations at step $n+1$. 
The source set is processed by the set transformer to obtain a learned representation of the current state. 
This source set representation is then passed through a simple linear layer followed by a softmax to yield a categorical distribution over the next atoms' types.
The continuous next atoms' positions are modeled by the flow matching head, conditioned on the source set representation and the next atom type. 
During training, we use teacher forcing to decouple the prediction of atom types from the prediction of atom positions.
Because the expensive source set representation is computed only once per generated token, our model is computationally more efficient than non-autoregressive frameworks that rely entirely on diffusion or flow matching.
The training procedure is summarized in Figure~\ref{fig:overview_training}.

\paragraph{Molecular representation}
Molecules are represented as undirected graphs $G = (V, E, \mathbf{a}, \mathbf{X})$, where $V=[N]$ is the set of node indices, $E \subseteq \{\{i,j\} : 1 \leq i < j \leq N \}$ is the set of undirected edges, 
atom types $\mathbf{a} \in \mathcal{A}^N$ are elements of vocabulary $\mathcal{A}$ and positions $\mathbf{X} \in \mathbb{R}^{N \times 3}$ are the Cartesian coordinates after zero-centering the molecule. 
The vocabulary $\mathcal{A}$ is the set of possible atom types (hydrogen included), and a stop token, which stops the generation process. 
A subgraph $G' = (V', E')$ of $G$, denoted by $G' \subseteq G$, is defined as a subset of the nodes $V' \subseteq V$ and edges $E' \subseteq E$.
Furthermore, $G' = G[V']$ is a node-induced subgraph of $G$, if $E' = (V' \times V') \cap E$.
A connected graph contains at least one path between any pair of vertices, and a path is defined as a sequence of vertices $(v_0, \dots, v_n)$ with $(v_i, v_{i+1}) \in E$ for $0 \leq i < n$. 
The neighborhood of a node index $i$ is the set of all adjacent node indices $\mathcal{N}(i) = \{j \in V : (i, j) \in E\}$.
Given a node-induced subgraph $G[V']$, we further define the set of boundary nodes $\partial V' = (\bigcup_{i \in V'} \mathcal{N}(i)) \setminus V'$. 
Finally, given a node subset $V'$, we denote the corresponding subset of atom types as $\mathbf{a}[V']$ and the subset of atom positions as $\mathbf{X}[V']$.

\paragraph{Autoregressive 3D molecular graph generation}
In the build-up phase, the model iteratively predicts atom type $a_{n+1}$ and position $\mathbf{x}_{n+1}$, given the current set of atoms and positions at step $n$.
We will use the following compact notation $\mathbf{a}_{1:n} = (a_1, \dots, a_n)$ to refer to the generated atom types and positions $\mathbf{X}_{1:n} = (\mathbf{x}_1, \dots, \mathbf{x}_n)$ at step $n$.
The autoregressive build-up stops when the model predicts a stop token. 
Edges are then inferred from the $N$ generated atoms \textit{via} a predefined set of rules $E = \mathcal{R}(\mathbf{a}, \mathbf{X})$, which completes inference of the molecular graph $G$.
While edges are inferred from atom types and positions after the build-up phase, the edge information can still be used for teacher guidance in the training phase. In the following, we describe how we use this information for model training.

\begin{wrapfigure}{r}{0.51\textwidth}
  \vspace{-22pt} 
  \centering
  \begin{minipage}{\linewidth}
    \begin{algorithm}[H] 
  \caption{Source-target split augmentation}
  \label{alg:source_target}
  \begin{algorithmic}[1]
    \State {\bfseries Input:} Graph $G = (V, E)$, $\beta$, $\gamma$
    \algphase{Source set construction}
    \State $v_0 = \text{RandomSeedNode}(V)$
    \State $V_S = \{v_0\}$
    \State $m = \lfloor {\beta \cdot \epsilon(v_0) \cdot u_0} \rfloor + 1, \; u_0 \sim \mathcal{U}_{[0, 1)}$
    \For{$i=1$ {\bfseries to} $m$}
        \State $\partial V_S \gets (\bigcup_{j \in V_S} \mathcal{N}(j)) \setminus V_S$
        \State $\partial V_S' \gets \{v \in \partial V_S \mid u_v > \gamma, u_v \sim \mathcal{U}_{(0, 1)} \}$
        \State $V_S \gets V_S \cup \partial V_S'$
    \EndFor
    \algphase{Target set construction}
    \State $V_T \gets (\bigcup_{j \in V_S} \mathcal{N}(j)) \setminus V_S$
    \State {\bfseries Return:} $V_S, V_T$
  \end{algorithmic}
\end{algorithm}
  \end{minipage}
  \vspace{-5pt}
\end{wrapfigure}

\paragraph{Neighborhood guidance}
We propose a sampling strategy that mimics molecular growth during the autoregressive build-up phase. 
Given a molecular graph $G$ in the training set, we sample a node-induced, connected subgraph $G[V_S]$ of $G$. We call $V_S$ the \emph{source set}. 
$G[V_S]$ simulates a molecular fragment with atom types $\mathbf{a}[V_S] = a_{1:n}$ and positions $\mathbf{X}[V_S] = \mathbf{x}_{1:n}$ in the autoregressive build-up at step $n = |V_S|$. 
The \emph{target set} $V_T$ is the set of valid token continuations at step $n+1$. 
It is the set of boundary nodes $\partial V_S$, given $V_S$, and induces atom types $\mathbf{a}[V_T]$ and positions $\mathbf{X}[V_T]$ for model supervision.
The sampling of connected subgraphs is performed dynamically during training. 
First, we compute the eccentricity $\epsilon(v)$ (the maximum shortest-path distance from $v$ to all other nodes in $G$) for each node $v \in V$. 
During runtime, a random seed node $v_0$ is selected from each molecular graph in the batch. 
Starting from this seed node, we perform $m$ rounds of 1-hop neighborhood aggregation, where $m$ is a random integer between $1$ and $\epsilon(v_0)$. 
To avoid biasing the sampling process toward globular subgraphs, we introduce stochasticity by randomly dropping a subset of the boundary nodes in each aggregation round, controlled by a dropout factor $\gamma$. 
To counteract the resulting bias toward smaller subgraphs, we introduce a scaling parameter $\beta$, which adjusts the number of aggregation rounds per graph. 
By balancing these two parameters, our sampling strategy generates connected subgraphs of varying sizes and shapes, from a single node to the full molecular graph, thereby simulating molecular growth in an autoregressive manner.

The pseudo-code for our sampling strategy is provided in Algorithm~\ref{alg:source_target}.
Additional information on the impact of $\gamma$ and $\beta$ on the sampling strategy and the values used in this work are available in Appendix~\ref{subsec:source_target_split}.
In practice, this sampling strategy is efficiently implemented using batched operations and introduces only minimal overhead to the training process, which is dominated by the forward passes through the transformer.

The geometry of generated molecules can deviate from training examples, which can lead to error accumulation at inference. 
To improve robustness, we inject Gaussian noise $\mathcal{N}(0,\sigma^2 I)$ into a $\delta$-fraction of source-set atoms during training, where $\sigma$ and $\delta$ are model-specific hyperparameters (details in Appendix~\ref{subsec:hyperparameters}). 
We hypothesize that this augmentation trains the model to tolerate inaccuracies from earlier generation stages. 
We also re-center atom positions in both the source and target sets w.r.t. the noised source-set atoms prior to computing the source-set representation. 
Both design choices improve NEAT’s predictive performance. Their impact is quantified in Appendix~\ref{subsec:scaling_and_ablations}.

\paragraph{Transformer trunk}
The source set representation $f_\theta : (\mathbf{a}_{1:n}, \mathbf{X}_{1:n}) \rightarrow \mathbf{z}_n$ at step $n$ is computed by encoding atom types $\mathbf{a}_{1:n}$ with a linear embedding layer and positions $\mathbf{X}_{1:n}$ with Fourier features \citep{tancik2020fourier}. 
Summing these encodings yields the initial node representations. 
For the transformer backbone $f_\theta$, we adopt the NanoGPT implementation \citep{Karpathy2025nanoGPT} of GPT-2 \citep{radford2019language}, removing positional embeddings and using bidirectional attention. 
After transformer updates, additive pooling \citep{zaheer2017deep} aggregates node representations into the set-level representation $\mathbf{z}_n$, characterizing the source set at step $n$.
Due to conceptual similarity with the \emph{set attention block} (SAB) by \citet{Lee2019SetTransformer}, we refer to this setup as \emph{set transformer}. 
Using a set transformer ensures permutation invariance to the order of atoms in the \emph{source set}.

\paragraph{Next token prediction}  
The set-level representation $\mathbf{z}_n$ is used to model the distribution of possible next tokens, represented as the joint distribution $p(a_{n+1}, \mathbf{x}_{n+1} \mid \mathbf{z}_n)$, which is factorized as follows:
\begin{equation}
    p(a_{n+1}, \mathbf{x}_{n+1} \mid \mathbf{z}_n) = p(a_{n+1} \mid \mathbf{z}_n) \cdot p(\mathbf{x}_{n+1} \mid a_{n+1}, \mathbf{z}_n),
\end{equation}
where $p(a_{n+1} \mid \mathbf{z}_n)$ represents the distribution over the next atom types, and $p(\mathbf{x}_{n+1} \mid a_{n+1}, \mathbf{z}_n)$ models the distribution over the next atom positions given the next atom type. 
The next atom type probability $p(a_{n+1} \mid \mathbf{z}_n)$ is modeled as a categorical distribution over the vocabulary of atom types and is approximated with a linear layer followed by a softmax operation $g_\theta : \mathbf{z}_n \rightarrow \hat{a}_{n+1}$. 
For model supervision, we compute the cross-entropy loss between $\hat{a}_{n+1}$ and the observed distribution over target atom types, which is obtained by computing the mean over the one-hot-encoded atom types in the target set $a_{n+1} = \text{mean}(\text{onehot}(\mathbf{a}[V_T]))$.
The next atom position is modeled as a conditional distribution given the next atom type. 
We model this distribution \textit{via} conditional flow matching (CFM, see Appendix~\ref{sec:flow_matching} for a more detailed description of the formalism) \citep{Lipman2023FlowMatching, Albergo2023NormalizingFlows}.
During training, we employ teacher forcing to decouple the atom type prediction from the task of positional recovery. 
For the set of possible next positions $\mathbf{X}_1 =\mathbf{X}[V_T]$ corresponding to the target set $V_T$, we sample a set of random initial positions $\mathbf{X}_0 \sim p_0$, where $p_0 =\mathcal{N}(0, \sigma^2 I)$.
We construct interpolation paths between the positions at the random position at $t=0$ and the true position at $t=1$:
\begin{equation}
    \mathbf{X_t} = (1-\mathbf{t}) \cdot \mathbf{X}_{0} + \mathbf{t} \cdot \mathbf{X}_{1}
\end{equation}
with random time steps $\mathbf{t} \in \mathcal{U}(0,1)^{|V_T|}$, yielding a set of interpolated positions $\mathbf{X_t}$ with velocities $\mathbf{V_t} = \mathbf{X}_{1} - \mathbf{X}_{0}$.
We then train the conditional vector field $\mathbf{V}_\theta$ through atom-wise regression against the set of CFM path velocities:
\begin{equation}
    \mathcal{L_\text{CFM}} = \mathbb{E}_{\mathbf{t}, \mathbf{X}_0, \mathbf{X}_1} \left[  \frac{1}{|V_T|} \lVert \mathbf{V}_\theta(\mathbf{X_t}, \textbf{t}, \mathbf{a}[V_T], \mathbf{z}_n) - \mathbf{V_t} \rVert_F^2\right]
\end{equation}
where $\lVert \cdot \rVert_F^2$ denotes the squared Frobenius norm.
Aggregating atom-wise contributions \textit{via} summation ensures permutation invariance to the order of atoms in the \emph{target set}.
The vector field $\mathbf{V}_\theta$ is parametrized using an MLP with adaptive layer normalization (adaLN), as done in \citep{Li2024AutoregressiveImageGenerationWithoutVQ}.
Input to the MLP for velocity regression is the atom-wise summation of positional encoding, atom type embedding, time step embedding, and source set representation. 
The final loss is obtained by adding the cross-entropy loss and the flow matching loss.
In order to reduce the transport cost between the source distribution $p_0$ and the target distribution $p_1$, we employ a simple coupling scheme. 
Given the positions $\mathbf{X}_1$ and $\textbf{X}_0$, we solve the linear assignment problem on the Euclidean distance matrix $D(\textbf{X}_1,\textbf{X}_0)$ and permute the positions in $\textbf{X}_0$ accordingly. 
This procedure avoids crossing paths, resulting in improved performance.

\paragraph{Inference}
The generation process begins by randomly sampling an initial atom type $a_1$ from the atom type vocabulary $\mathcal{A}$ and drawing its initial position $\mathbf{x}_1 \in \mathbb{R}^{3}$ from a normal distribution $\mathcal{N}(0, \sigma^2I)$. 
At each subsequent autoregressive step, the current atom set $(\mathbf{a}_{1:n}, \mathbf{X}_{1:n})$ is passed through the transformer, producing the set-level representation $\mathbf{z}_n$. 
Note that the positions are re-centered between autoregressive steps. 
The distribution over the possible next atom types is predicted from $\mathbf{z}_n$ using a linear layer followed by the softmax function, from which the next atom type $a_{n+1}$ is sampled.
A random initial position $\mathbf{x}_{n+1}$ is then sampled from $\mathcal{N}(0,\sigma^2I)$, and the transport equation is integrated from $t=0$ to $t=1$ to compute the next atom position.
Details about atom type sampling and the integration scheme are provided in Appendix~\ref{subsec:inference_details}. 
This iterative process continues until a stop token is predicted, signaling the end of the molecular generation.
Molecular generation is efficiently implemented using batched operations, enabling parallel generation and high computational speed. 
\section{Experiments}

\paragraph{Data}
We used the QM9 \citep{Ramakrishnan2014QM9} and GEOM-Drugs \citep{Axelrod2022} datasets.
QM9 is a collection of small organic molecules composed of up to nine heavy atoms (C, N, O, F), with a total of up to 29 atoms when including hydrogen. 
We obtained the QM9 dataset from \citet{raw_qm9_data} and, following standard practice \citep{Hoogeboom2022EDM, Cheng2025Quetzal}, randomly split it into training, validation, and test sets, yielding 100,000 molecules for training, 11,619 for validation, and 12,402 for testing after preprocessing.
GEOM-Drugs contains 304,313 drug-like molecules, each with up to 30 conformers. 
Following \citet{Dunn2025FlowMol}, we loaded the split data from \citet{raw_geom_data}. 
Like \citet{Vignac2023MiDi}, we retained, for each molecule, the five lowest-energy conformers when available. 
After preprocessing, the dataset contained 1,172,436 conformers across 243,451 unique molecules for training, 146,394 conformers across 30,430 unique molecules for validation, and 146,788 conformers across 30,432 unique molecules for testing. 
The mean and median number of conformers per molecule are 4.8 and 5.0, respectively. 
In the following, we refer to GEOM-Drugs only as GEOM. 
Detailed information about data processing is provided in the Appendix~\ref{subsec:data_processing}.

\paragraph{Model training}
We trained on QM9 and GEOM for 10,000 and 1,000 epochs, respectively, keeping the checkpoint with the lowest validation loss for inference. 
System specifications, implementation details, hyperparameters of the final models, scaling analyses w.r.t. transformer size, and ablations of key design decisions are provided in Appendix~\ref{subsec:system_software_specification}-\ref{subsec:scaling_and_ablations}. 
Our best-performing models (QM9 and GEOM) both have a 12-layer transformer with 12 heads and hidden dimension 768, and an MLP projector comprising 6 AdaLN blocks with hidden dimension 1536 (165M parameters). 
This configuration offers a favorable trade-off between generation quality and model size.

\paragraph{Evaluation metrics}
To evaluate the quality of generated molecules, we adhere to standard practices \citep{Hoogeboom2022EDM, Daigavane2024Symphony, Zhang2025SymDiff, Cheng2025Quetzal} and report atomic and molecular stability, molecular validity, uniqueness, and novelty. 
An atom is considered stable if its valence satisfies the octet (heavy atoms) or duet (hydrogen atoms) rule. 
A molecule is stable if all its atoms are stable. 
Molecular uniqueness is the percentage of distinct canonical SMILES strings derived from valid molecules (\textit{i.e.}, the percentage of valid \underline{and} unique molecules). 
Novelty is the percentage of valid and unique molecules that are not present in the training set. 
A comparison of the distribution of molecular properties between the generated molecules and the training set can be found in Appendix~\ref{subsec:mol_props}.
Examples of molecules generated by NEAT are shown in Appendix~\ref{subsec:examples_generated_mols}. Failure modes are discussed in Appendix~\ref{subsec:failure_modes}.

\paragraph{Molecule building} 
For NEAT as well as for every other model that does not explicitly generate bonds, molecular validity depends on the algorithm used to construct molecules from 3D clouds of unconnected atoms. 
We investigate two approaches: (1) the widespread pipeline by \citet{Hoogeboom2022EDM} based on bond length lookup tables, (2) the pipeline introduced by \citet{Daigavane2024Symphony} and adopted by \citet{Cheng2025Quetzal}, where a molecule is considered valid if bonds can be assigned using RDKit's MolFromXYZBlock function and rdDetermineBonds module, which uses the xyz2mol package based on the work of \citet{Kim2015xyz2mol}. 
A notable limitation of the lookup-based pipeline is the absence of aromatic bond handling and a high sensitivity to bond lengths \citep{Daigavane2024Symphony}, which yields systematically lower scores for valid aromatic structures and thus reduces the informative value of these metrics.
We also observed that the output of RDKit's rdDetermineBonds module depends on the RDKit version.
Throughout this work, we use RDKit v2025.9.1. 
A detailed discussion of limitations arising from these bond-assignment-based evaluation pipelines is provided in Appendix~\ref{sec:observations_on_eval_metrics}.

\paragraph{Baselines}
We compare NEAT trained on QM9 to five autoregressive models (G-SchNet \citep{Gebauer2019GSchnNet}, G-SphereNet \citep{Luo2022GSphereNet}, Symphony \citep{Daigavane2024Symphony}, QUETZAL \citep{Cheng2025Quetzal}, Geo2Seq \citep{Li2025Geo2Seq}). 
The results are reported in Table~\ref{tab:results_unconditional}, with additional comparisons to diffusion-based 3D molecular generation models provided in Appendix~\ref{subsec:comparison_to_diffusion}. 
We compare NEAT trained on GEOM to three flow-based models (FlowMol3 \citep{Dunn2025FlowMol}, SemlaFlow \citep{Irwin2025}, Megalodon-FM \citep{Reidenbach2025}) and two autoregressive models (QUETZAL \citep{Cheng2025Quetzal}, Geo2Seq \citep{Li2025Geo2Seq}) and report the results in Table~\ref{tab:results_unconditional}. 
Since the evaluation metrics are sensitive to the procedure used to infer bonds from atomic point clouds, we generated molecules with the source code and model weights of each baseline wherever possible and evaluated them using the same pipeline.
For G-SchNet, G-SphereNet, and Symphony, we used the pre-computed XYZ files released by the authors of Symphony \citep{XYZ_by_symphony} and computed metrics with our pipeline. 
For Geo2Seq, we report the metrics from the original publication. 

\begin{table}[t!]
    \centering
    \caption{Benchmarking of NEAT against 3D molecular generators trained on QM9 and GEOM. Metrics are reported as percentages. The first row reports results on the QM9 test set (12,402 molecules). The eighth row reports results on the GEOM test set (146,788 conformers, 30,432 unique molecules). Other rows report results computed from 10,000 generated molecules. Entries for which generated molecules could be retrieved and re-evaluated with our pipeline are marked with a single star (*). Entries for which the source code and model weights were available to generate molecules are marked with two stars (**). For these models and for NEAT, we report the mean $\pm$ one standard deviation over three runs. Otherwise, values are taken directly from original publications.}
    \resizebox{\linewidth}{!}{
    \begin{tabular}{llllllll}
    \toprule
    Model & 
    \makecell[l]{atom\\stable} &
    \makecell[l]{mol\\stable} &
    \makecell[l]{lookup\\valid}&
    \makecell[l]{lookup\\unique}&
    \makecell[l]{xyz2mol\\valid} &
    \makecell[l]{xyz2mol\\unique} &
    \makecell[l]{xyz2mol\\novel} \\
    \midrule
    \midrule
    QM9                & 99.3 & 94.8 & 97.5 & 97.5 & 100.0 & 100.0 & 99.8 \\
    \midrule
    G-SchNet*          & 93.7 & 63.6 & 80.1 & 74.7 & 73.0 & 70.6 & \textbf{55.1} \\
    G-SphereNet*       & 67.8 & 14.0 & 16.4 & 3.8  & 37.7 & 7.2  & 6.7  \\
    Symphony*          & 90.8 & 43.9 & 68.1 & 66.5 & 77.0 & 75.2 & 41.3 \\
    Geo2Seq            & \textbf{98.9} & \textbf{93.2} & \textbf{97.1} & 81.7 & n.a. & n.a. & n.a. \\
    QUETZAL**          & \underline{98.3} $\pm$ 0.1 & \underline{87.5} $\pm$ 0.1 & \underline{94.6} $\pm$ 0.1 & \textbf{90.1} $\pm$ 0.2 & \underline{93.7} $\pm$ 0.3 & \underline{89.1} $\pm$ 0.2 & 24.7 $\pm$ 0.3 \\
    \rowcolor{gray!15} NEAT (ours)        & 97.8 $\pm$ 0.1 & 83.1 $\pm$ 0.1 & 93.3 $\pm$ 0.1 & \underline{88.8} $\pm$ 0.3 & \textbf{95.2} $\pm$ 0.2 & \textbf{91.6} $\pm$ 0.3 & \underline{53.5} $\pm$ 0.6 \\
    \midrule
    \midrule
    GEOM & 86.2 & 2.8 & 96.6 & 96.6 & 93.5 & 93.5 & 93.5 \\
    \midrule
    FlowMol3** & \textbf{88.8} $\pm$ 0.1 & \textbf{5.0} $\pm$ 0.8 & 96.5 $\pm$ 0.6 & 96.5 $\pm$ 0.5& \textbf{95.8} $\pm$ 0.7 & \textbf{95.7} $\pm$ 0.7 & \textbf{95.7} $\pm$ 0.7 \\
    SemlaFlow** & 84.1 $\pm$ 0.1 & 2.2 $\pm$ 0.1 & 92.9 $\pm$ 0.1 & 92.9 $\pm$ 0.1 & 86.6 $\pm$ 0.5 & 86.5 $\pm$ 0.5 & 86.5 $\pm$ 0.5 \\
    Megalodon-FM** & 81.5 $\pm$ 0.1 & 1.8 $\pm$ 0.1 & \underline{97.9} $\pm$ 0.2 & \textbf{97.8} $\pm$ 0.2 & 88.2 $\pm$ 0.2 & 88.1 $\pm$ 0.2 & 88.1 $\pm$ 0.2 \\
    Geo2Seq & 82.5 & n.a. & 96.1 & n.a. & n.a. & n.a. & n.a. \\
    QUETZAL** & \underline{86.7} $\pm$ 0.1 & \underline{2.5} $\pm$ 0.1 & 95.7 $\pm$ 0.1 & 95.4 $\pm$ 0.1 & 90.1 $\pm$ 0.1 & 89.8 $\pm$ 0.1 & 89.8 $\pm$ 0.1 \\
    \rowcolor{gray!15} NEAT (ours) & 81.5 $\pm$ 0.1 & 0.6 $\pm$ 0.1 & \textbf{99.7} $\pm$ 0.1 & \underline{96.9} $\pm$ 0.1 & \underline{93.9} $\pm$ 0.2 & \underline{91.6} $\pm$ 0.2 & \underline{91.6} $\pm$ 0.2 \\
    \bottomrule
    \end{tabular}
    }
    \label{tab:results_unconditional}
\end{table}

\paragraph{Unconditional generation performance}
On QM9, NEAT outperformed autoregressive models in terms of xyz2mol validity and uniqueness.
NEAT ranked second to QUETZAL on lookup uniqueness, and third on lookup validity, atom stability, and molecular stability.
While Geo2Seq achieved the highest scores for atom stability, molecular stability, and lookup validity, it performed poorly on molecular uniqueness.
On the GEOM dataset, NEAT achieved the highest lookup validity and second-best lookup uniqueness and xyz2mol validity, uniqueness and novelty. 
All models yielded low molecular stability, close to that of the GEOM test set (2.8\%). 
A detailed discussion of this value is provided in Appendix~\ref{sec:observations_on_eval_metrics}.

Neighborhood guidance encourages a generation order in which the partial molecule remains connected, a prerequisite for molecular validity. 
To verify this, we examined the final molecular graphs by iterating over atoms in generation order and checking whether each new atom $n$ was connected to the subgraph induced by atoms $1,\dots,n-1$. 
This condition held for every generated molecule, confirming that the generation process follows the intended neighborhood-guided behavior.

\begin{table}[htb!]
    \centering
    \caption{Prefix completion performance of NEAT, QUETZAL, and FlowMol3 trained on GEOM. Metrics are reported as percentages. For each of the 100 prefixes, we generate 1,000 molecules and compute average metric values. We then report the mean $\pm$ one standard deviation across the 100 prefix-level averages.}
    \resizebox{\linewidth}{!}{
    \begin{tabular}{lccccccc}
    \toprule
     Model & 
     \makecell{atom\\stable} &
     \makecell{mol\\stable} &
     \makecell{lookup\\valid}&
     \makecell{lookup\\unique}&
     \makecell{xyz2mol\\valid} &
     \makecell{xyz2mol\\unique} &
     \makecell{xyz2mol\\novel} \\
     \midrule
     FlowMol3 (inpainted) & 76.3 $\pm$ 6.5 & 0.4 $\pm$ 1.3 & ~~44.5 $\pm$ 16.3 & ~~36.3 $\pm$ 20.1 & 30.2 $\pm$ 7.8 & 29.4 $\pm$ 7.3 & 29.4 $\pm$ 7.3 \\
     QUETZAL & 69.0 $\pm$ 1.8 & 0.1 $\pm$ 0.1 & 27.8 $\pm$ 6.8 & 13.6 $\pm$ 4.1 & 28.1 $\pm$ 4.7 & 17.7 $\pm$ 2.8 & 17.7 $\pm$ 2.8 \\
     \rowcolor{gray!15} NEAT (ours) & \textbf{77.0} $\pm$ 1.5 & \textbf{0.5} $\pm$ 0.3 & \textbf{97.8} $\pm$ 1.8 & \textbf{82.2} $\pm$ 3.0 & \textbf{92.7} $\pm$ 2.3 & \textbf{77.9} $\pm$ 3.0 & \textbf{77.9} $\pm$ 3.0 \\
    \bottomrule
    \end{tabular}
    }
    \label{tab:results_prefix}
\end{table}

\paragraph{Prefix completion}
We demonstrate the utility of our approach by comparing NEAT to QUETZAL on the task of designing 3D molecules around a given scaffold, referred to as \textit{prefix completion}. 
This task has practical relevance in drug discovery, for example for lead optimization.
Both models are transformer-based autoregressive 3D molecule generators, but differ in one key aspect. 
QUETZAL uses a causal transformer with sequential encodings and is trained to generate molecules based on the canonical ordering of atoms in XYZ files, whereas NEAT uses a set transformer with bidirectional attention and is trained using neighborhood guidance.
For QUETZAL to generate a valid molecule, the prefix must align with the first $n$ atoms of the fully generated molecule after canonicalization. 
This requirement cannot be satisfied without prior knowledge of the final molecule.

We tasked both models (trained on GEOM and run with default settings) with generating 1,000 molecules from 100 incomplete prefixes extracted from GEOM, resulting in a total of 100,000 completions. 
The results are summarized in Table~\ref{tab:results_prefix}.
The prefixes represent diverse ring systems with varying substitution patterns.
2D plots of all prefixes are provided in Appendix~\ref{subsec:prefixes}.
For each prefix, we generated 3D molecular coordinates using RDKit's implementation of the Universal Force Field (UFF), initially including hydrogen atoms at the substitution positions. These hydrogen atoms were subsequently removed to enable the model to continue molecular generation at the designated sites, and each prefix was randomly rotated for every generated sample.

Unlike QUETZAL, NEAT successfully generates valid molecules from arbitrary prefixes by learning the structural features of valid molecules independent of atom ordering conventions.
Compared with unconditional generation (Table~\ref{tab:results_unconditional}), prefix completion with NEAT (Table~\ref{tab:results_prefix}) largely preserves stability and validity, while reducing diversity.
The reduced diversity is expected, since fixing a prefix constrains the space of possible molecules.
The moderate decrease in atom and molecule stability can be explained by the fact that most benchmarked prefixes contain aromatic rings (Appendix Figures~\ref{fig:full_prefixes_1}-\ref{fig:full_prefixes_3}). These stability metrics fail for aromatic compounds because the underlying pipeline does not account for the length of aromatic bonds (see Appendix~\ref{sec:observations_on_eval_metrics} for details).
The small drop in the validity metrics may be due to the inclusion of rare prefixes, which are harder for the model to work with because it has seen fewer examples of these during training.

We also compared NEAT's prefix completion capabilities to the ones of non-autoregressive models. 
To the best of our knowledge, no published diffusion- or flow-based baseline evaluates prefix-conditioned inpainting for unconditional 3D molecular generation on GEOM. 
We therefore implemented this setting in our strongest baseline, FlowMol3, by modifying only its inference procedure, without retraining. 
Details are provided in Appendix~\ref{sec:inpainting}.
Our results show that inference-time inpainting with FlowMol3 is feasible, but substantially degrades validity and related quality metrics. 
A scaffold- or prefix-conditioned transport-based model trained specifically for this task could potentially reduce this gap, but would require substantial changes to the training objective, conditioning mechanism, and data construction. 
By comparison, NEAT supports prefix completion directly, without retraining or model modifications, highlighting its practical advantage.

\begin{figure}
    \centering
    \includegraphics[width=1.0\linewidth]{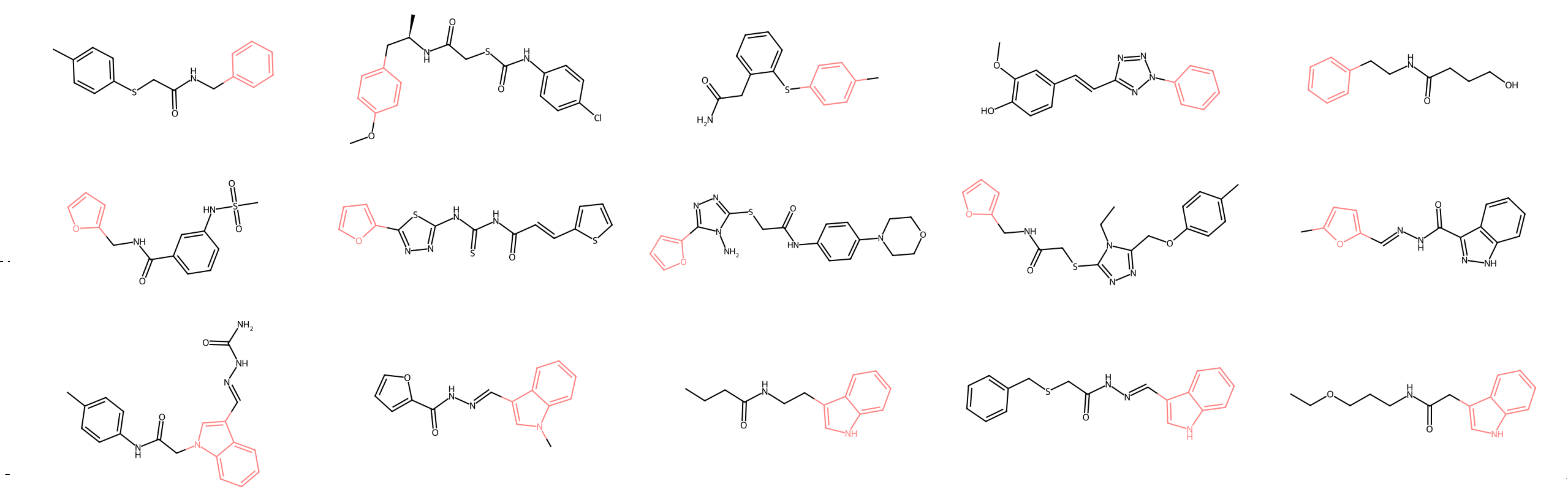}
    \caption{Randomly selected examples of molecules generated from three different prefixes: para-substituted benzene, 2,5-substituted furan, and 1,3-substituted indole.}
    \label{fig:prefixes_examples}
\end{figure}

\paragraph{Runtime}

\begin{figure}[t!]
    \centering
    \includegraphics[width=1.0\linewidth]{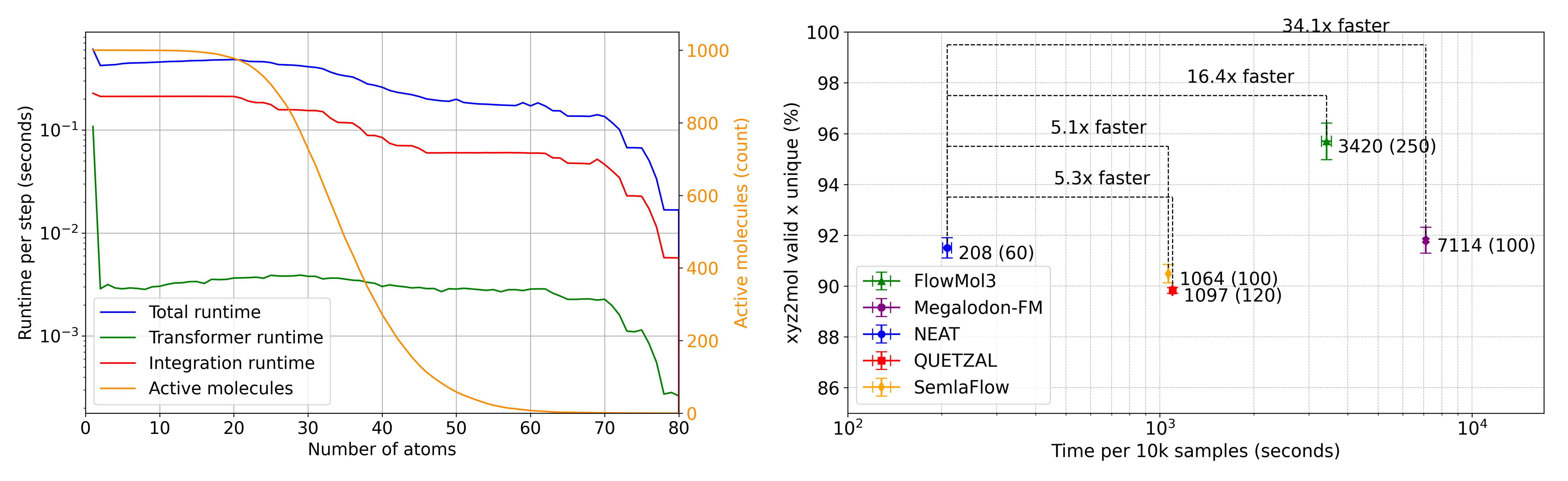}
    \caption{\textbf{Left}: Runtime per step when generating 1000 molecules, averaged over 10 runs. \textbf{Right}: Comparison of xyz2mol validity x uniqueness vs. runtime between NEAT and models trained on GEOM. The number attached to each data point corresponds to the runtime in seconds followed by the number of integration steps (chosen following the recommendations of the authors) in parenthesis. Values are averaged over three runs. Error bars correspond to one standard deviation. Batch size: largest to fit on GPU (FlowMol3: 500; SemlaFlow: 10,000; Megalodon-FM: 50; QUETZAL: 2,000; NEAT: 10,000).}
    \label{fig:runtime}
\end{figure}

We analyze NEAT's computational complexity starting with the worst-case scenario, where transformer calls dominate runtime. 
Its worst-case complexity is $\mathcal{O}(BN^3)$, with $N$ as the number of atoms per molecule in a batch of $B$ molecules. 
Diffusion and flow models have complexity $\mathcal{O}(BTN^2)$, where $T$ is the number of integration steps. 
For drug-like molecules, $N$ is typically smaller than $T$, giving NEAT an advantage even in the worst-case scenario.
In practice, NEAT’s complexity is lower than that of the worst-case scenario. 
During training, NEAT makes only one transformer call ($\mathcal{O}(BN^2)$) and one MLP call ($\mathcal{O}(BN)$) per forward pass. 
With batch size $B$ constant, the transformer accounts for approximately 90\% of the runtime. 
During inference, the dominant workload shifts to the ODE integration ($\mathcal{O}(BT)$). 
Generating $N$ atoms scales as $\mathcal{O}(BTN)$. 
Another positive effect is that $B$ decreases as molecules complete, while $N$ grows linearly. 
Initially, transformer runtime increases but declines after generating ~25 atoms. 
We illustrate these observations and report the total runtime in Figure~\ref{fig:runtime}. 

NEAT operates 34 times faster than Megalodon-FM, 16 times faster than FlowMol3, and 5 times faster than SemlaFlow and QUETZAL.
When the baselines are run with fewer integration steps than recommended by their authors to match NEAT’s runtime, their generation quality falls below that of NEAT (See Appendix~\ref{subsec:quality_runtime_tradeoff}), further highlighting NEAT’s favorable quality--runtime trade-off.
Because batched inference depends on molecule size, we report size statistics for completeness. 
The average number of heavy atoms was similar across models: $22\pm5$ for FlowMol3, $21\pm6$ for Megalodon, $21\pm5$ for NEAT, $24\pm5$ for QUETZAL, and $23\pm5$ for SemlaFlow. 
The largest molecules, including hydrogen atoms, contained 154 atoms for FlowMol3, 54 for Megalodon, 94 for NEAT, 94 for QUETZAL, and 72 for SemlaFlow.

\paragraph{Limitations}
NEAT inherits several standard limitations of autoregressive architectures. 
Specifically, error propagation may degrade sampling quality for larger molecules, and pooling node embeddings could cause oversquashing, leading to information loss. 
To address this, we analyzed the correlation between molecular size and validity (Appendix~\ref{subsec:molsize_impact}) and found no evidence of size-dependent degradation. 
However, since NEAT does not extrapolate beyond the maximum atom count seen during training (Appendix~\ref{subsec:mol_props}), this conclusion is restricted to drug-like molecules. 
Additionally, NEAT's neighborhood-guidance mechanism relies on edge information derived from the molecular graph. 
This dependency may limit its applicability in domains where bond assignment is ambiguous, such as inorganic crystal structures or metal-organic compounds. 
While spatial information, such as radius-based graphs, could serve as an alternative source of neighborhood information, evaluating the effectiveness of NEAT in such generalized settings is beyond the scope of this study and remains an avenue for future work.
Finally, the transformer architecture employed in this study is a standard transformer and does not enforce SE(3)-equivariance. 
Despite this, our results demonstrate that combining this simple architecture with random rotation augmentation during training can achieve state-of-the-art generation performance. 
To quantify the impact of not enforcing SE(3)-equivariance, we conducted an invariance-sensitivity test (Appendix~\ref{subsec:se3_impact}) and observed only minor variance in predictions. 
While we conclude that the lack of architectural SE(3)-equivariance does not significantly impair model performance, orientation dependence in predictions cannot be entirely ruled out. 
Integrating NEAT with an SE(3)-equivariant transformer represents an interesting direction for future research.
\section{Conclusion}

We introduce a new method for training an autoregressive set transformer with bidirectional attention for 3D molecular generation using neighborhood guidance. 
Instead of sequential positional encodings, we rely on a subgraph sampling schema for model supervision. 
Treating the input as a permutation-invariant set removes the need for sequence canonicalization and makes the model well-suited for tasks like molecular prefix completion. 
On QM9 and GEOM, our approach achieves state-of-the-art performance, while offering permutation-invariance w.r.t. atoms, prefix completion, and fast sampling. 
We believe this provides a strong foundation for future work in conditional 3D molecular generation.
Source code and model weights are available at: \url{https://github.com/molinfo-vienna/NEAT}.

\begin{ack}
The authors thank Leo Gaskin for testing the code and Steffen Hirte for proofreading the manuscript. The financial support received for the Christian Doppler Laboratory for Molecular Informatics in the Biosciences by the Austrian Federal Ministry of Labour and Economy, the National Foundation for Research, Technology and Development, the Christian Doppler Research Association, Boehringer-Ingelheim RCV GmbH \& Co KG and BASF SE is gratefully acknowledged. There are no conflicts of interest to declare.
\end{ack}

\section*{Author contributions}
Conceptualization: DR, RAJ.
Methodology: DR, RAJ.
Software: DR, RAJ.
Data curation: DR, RAJ.
Formal analysis: DR, RAJ.
Investigation: DR, RAJ.
Visualization: DR, RAJ.
Writing (original draft): DR, RAJ.
Writing (review and editing): DR, RAJ, JK, TL.
Supervision: JK, TL.
Funding acquisition: JK, TL.


\bibliographystyle{abbrvnat}
\bibliography{references}

@inproceedings{Lee2019SetTransformer,
    title={Set Transformer: A Framework for Attention-based Permutation-Invariant Neural Networks},
    author={Lee, Juho and Lee, Yoonho and Kim, Jungtaek and Kosiorek, Adam and Choi, Seungjin and Teh, Yee Whye},
    city={Long Beach, California, United States of America},
    booktitle={Proceedings of the 36th International Conference on Machine Learning},
    publisher={PMLR},
    volume={97},
    pages={3744-3753},
    year={2019},
}

@article{schneuing2024structure,
    title={Structure-Based Drug Design with Equivariant Diffusion Models},
    author={Schneuing, Arne and Harris, Charles and Du, Yuanqi and Didi, Kieran and Jamasb, Arian and Igashov, Ilia and Du, Weitao and Gomes, Carla and Blundell, Tom L and Lio, Pietro and others},
    journal={Nat. Comput. Sci.},
    publisher={Nature Publishing Group US New York},
    volume={4},
    issue={12},
    pages={899-909},
    year={2024},
    doi={10.1038/s43588-024-00737-x},
}

@inproceedings{peng2022pocket2mol,
    title={{P}ocket2{M}ol: Efficient Molecular Sampling Based on 3{D} Protein Pockets},
    author={Peng, Xingang and Luo, Shitong and Guan, Jiaqi and Xie, Qi and Peng, Jian and Ma, Jianzhu},
    city={Baltimore, Maryland, United States of America},
    booktitle={Proceedings of the 39th International Conference on Machine Learning},
    publisher={PMLR},
    volume={162},
    pages={17644-17655},
    year={2022},
}

@inproceedings{zaheer2017deep,
    title={Deep Sets},
    author={Zaheer, Manzil and Kottur, Satwik and Ravanbakhsh, Siamak and Poczos, Barnabas and Salakhutdinov, Russ R and Smola, Alexander J},
    city={Long Beach, California, United States of America},
    booktitle={Advances in Neural Information Processing Systems},
    publisher={Curran Associates, Inc.},
    volume={30},
    pages={},
    year={2017},
}

@misc{Falcon_PyTorch_Lightning_2025,
    title={PyTorch Lightning},
    author={Falcon, William and {the PyTorch Lightning team}},
    license={Apache-2.0},
    version={2.5.5},
    url={https://github.com/Lightning-AI/lightning},
    year={2025},
}

@misc{rdkit_2025,
    title={{RDKit}: Open-Source Cheminformatics},
    author={Greg Landrum},
    license={BSD-3-Clause},
    version={2025.9.1},
    url={https://www.rdkit.org},
    year={2025},
}

@misc{fey2019pyg,
    title={Fast Graph Representation Learning with PyTorch Geometric}, 
    author={Matthias Fey and Jan Eric Lenssen},
    year={2019},
    eprint={1903.02428},
    archivePrefix={arXiv},
    primaryClass={cs.LG},
    url={https://arxiv.org/abs/1903.02428}, 
}

@inproceedings{loshchilov2019decoupled,
    title={Decoupled Weight Decay Regularization},
    author={Loshchilov, Ilya and Hutter, Frank},
    city={New Orleans, Louisiana, United States of America},
    booktitle={Proceedings of the 7th International Conference on Learning Representations},
    publisher={Curran Associates, Inc.},
    pages={},
    year={2019},
}

@inproceedings{tancik2020fourier,
    title={Fourier Features Let Networks Learn High Frequency Functions in Low Dimensional Domains},
    author={Tancik, Matthew and Srinivasan, Pratul P and Mildenhall, Ben and Fridovich-Keil, Sara and Raghavan, Nithin and Singhal, Utkarsh and Ramamoorthi, Ravi and Barron, Jonathan T and Ng, Ren},
    city={Vancouver, Canada},
    booktitle={Advances in Neural Information Processing Systems},
    publisher={Curran Associates, Inc.},
    volume={33},
    pages={7537-7547},
    year={2020},
}

@article{su2024roformer,
    title={Ro{F}ormer: Enhanced Transformer with Rotary Position Embedding},
    author={Su, Jianlin and Ahmed, Murtadha and Lu, Yu and Pan, Shengfeng and Bo, Wen and Liu, Yunfeng},
    journal={Neurocomputing},
    publisher={Elsevier},
    volume={568},
    number={127063},
    year={2024},
    doi={10.1016/j.neucom.2023.127063},
}

@inproceedings{Lipman2023FlowMatching,
    title={Flow Matching for Generative Modeling},
    author={Lipman, Yaron and Chen, Ricky TQ and Ben-Hamu, Heli and Nickel, Maximilian and Le, Matthew},
    city={Kigali, Rwanda},
    booktitle={Proceedings of the 11th International Conference on Learning Representations},
    publisher={Curran Associates, Inc.},
    pages={},
    year={2023},
}

@inproceedings{Albergo2023NormalizingFlows,
    title={Building Normalizing Flows with Stochastic Interpolants},
    author={Michael S Albergo and Eric Vanden-Eijnden},
    city={Kigali, Rwanda},
    booktitle={Proceedings of the 11th International Conference on Learning Representations},
    publisher={Curran Associates, Inc.},
    pages={},
    year={2023},
}

@inproceedings{wang2025learningorder,
    title={Learning-Order Autoregressive Models with Application to Molecular Graph Generation},
    author={Zhe Wang and Jiaxin Shi and Nicolas Heess and Arthur Gretton and Michalis Titsias},
    city={Vancouver, Canada},
    booktitle={Proceedings of the 42nd International Conference on Machine Learning},
    publisher={PMLR},
    volume={},
    pages={},
    year={2025},
}

@inproceedings{you2018graphrnn,
    title={Graph{RNN}: Generating Realistic Graphs with Deep Auto-Regressive Models},
    author={You, Jiaxuan and Ying, Rex and Ren, Xiang and Hamilton, William and Leskovec, Jure},
    city={Stockholm, Sweden},
    booktitle={Proceedings of the 35th International Conference on Machine Learning},
    publisher={PMLR},
    volume={80},
    pages={5708-5717},
    year={2018},
}

@article{Dunn2025FlowMol,
    title={Flow{M}ol3: Flow Matching for 3{D} De Novo Small-Molecule Generation},
    author={Dunn, Ian and Koes, David R.},
    journal={Digit. Discov.},
    year={2026},
    publisher={Royal Society of Chemistry},
    doi={10.1039/d5dd00363f},
}

@misc{wang2025simplefold,
    title={Simple{F}old: Folding Proteins is Simpler than You Think}, 
    author={Yuyang Wang and Jiarui Lu and Navdeep Jaitly and Josh Susskind and Miguel Angel Bautista},
    year={2025},
    eprint={2509.18480},
    archivePrefix={arXiv},
    primaryClass={cs.LG},
    url={https://arxiv.org/abs/2509.18480}, 
}

@misc{tong2024improving,
    title={Improving and Generalizing Flow-Based Generative Models with Minibatch Optimal Transport}, 
    author={Alexander Tong and Kilian Fatras and Nikolay Malkin and Guillaume Huguet and Yanlei Zhang and Jarrid Rector-Brooks and Guy Wolf and Yoshua Bengio},
    year={2024},
    eprint={2302.00482},
    archivePrefix={arXiv},
    primaryClass={cs.LG},
    url={https://arxiv.org/abs/2302.00482}, 
}

@article{radford2019language,
    title={Language Models are Unsupervised Multitask Learners},
    author={Radford, Alec and Wu, Jeffrey and Child, Rewon and Luan, David and Amodei, Dario and Sutskever, Ilya and others},
    journal={OpenAI blog},
    volume={1},
    number={8},
    year={2019},
}

@misc{Karpathy2025nanoGPT,
    title={Nano{GPT}},
    author={Andrej Karpathy},
    year={2025},
    license={MIT},
    url={https://github.com/karpathy/nanoGPT},
    note={Accessed 2025-11-15},
}

@inproceedings{Li2024AutoregressiveImageGenerationWithoutVQ,
    title={Autoregressive Image Generation without Vector Quantization},
    author={Li, Tianhong and Tian, Yonglong and Li, He and Deng, Mingyang and He, Kaiming},
    city={Vancouver, Canada},
    booktitle={Advances in Neural Information Processing Systems},
    publisher={Curran Associates, Inc.},
    volume={37},
    pages={56424-56445},
    year={2024},
    doi={10.52202/079017-1797},
}

@inproceedings{Kusner2017GVAE,
    title={Grammar Variational Autoencoder},
    author={Matt J Kusner and Brooks Paige and José Miguel Hernández-Lobato},
    city={Sydney, Australia},
    booktitle={Proceedings of the 34th International Conference on Machine Learning},
    publisher={PMLR},
    volume={70},
    pages={1945-1954},
    year={2017},
}

@article{GomezBombarelli2018CVAE,
    title={Automatic Chemical Design Using a Data-Driven Continuous Representation of Molecules},
    author={Rafael Gómez-Bombarelli and Jennifer N. Wei and David Duvenaud and José Miguel Hernández-Lobato and Benjamín Sánchez-Lengeling and Dennis Sheberla and Jorge Aguilera-Iparraguirre and Timothy D. Hirzel and Ryan P. Adams and Alán Aspuru-Guzik},
    journal={ACS Central Science},
    publisher={ACS Publications},
    volume={4},
    issue={2},
    pages={268-276},
    year={2018},
    doi={10.1021/acscentsci.7b00572},
}

@inproceedings{Simonovsky2018GraphVAE,
    title={Graph{VAE}: Towards Generation of Small Graphs Using Variational Autoencoders},
    author={Simonovsky, Martin and Komodakis, Nikos},
    city={Rhodes, Greece},
    booktitle={Proceedings of the 27th International Conference on Artificial Neural Networks},
    publisher={Springer},
    volume={},
    pages={412-422},
    year={2018},
    doi={10.1007/978-3-030-01418-6_41},
}

@misc{DeCao2022MolGAN,
    title={Mol{GAN}: An Implicit Generative Model for Small Molecular Graphs}, 
    author={Nicola De Cao and Thomas Kipf},
    year={2022},
    eprint={1805.11973},
    archivePrefix={arXiv},
    primaryClass={stat.ML},
    url={https://arxiv.org/abs/1805.11973}, 
}

@inproceedings{Hoogeboom2022EDM,
    title={Equivariant Diffusion for Molecule Generation in 3{D}},
    author={Emiel Hoogeboom and Victor Garcia Satorras and Clément Vignac and Max Welling},
    city={Baltimore, Maryland, United States of America},
    booktitle={Proceedings of the 39th International Conference on Machine Learning},
    publisher={PMLR},
    volume={162},
    pages={8867-8887},
    year={2022},
}

@inproceedings{Gebauer2019GSchnNet,
    title={Symmetry-Adapted Generation of 3{D} Point Sets for the Targeted Discovery of Molecules},
    author={Niklas W A Gebauer and Michael Gastegger and Kristof T Schütt},
    city={Vancouver, Canada},
    booktitle={Advances in Neural Information Processing Systems},
    publisher={Curran Associates, Inc.},
    volume={32},
    pages={7566-7578},
    year={2019},
}

@inproceedings{Luo2022GSphereNet,
    title={An Autoregressive Flow Model for 3{D} Molecular Geometry Generation from Scratch},
    author={Youzhi Luo and Shuiwang Ji},
    city={Virtual},
    booktitle={Proceedings of the 10th International Conference on Learning Representations},
    publisher={Curran Associates, Inc.},
    pages={},
    year={2022},
}

@inproceedings{Daigavane2024Symphony,
    title={Symphony: Symmetry-Equivariant Point-Centered Spherical Harmonics for 3{D} Molecule Generation},
    author={Ameya Daigavane and Song Eun Kim and Mario Geiger and Tess Smidt},
    city={Vienna, Austria},
    booktitle={Proceedings of the 12th International Conference on Learning Representations},
    publisher={Curran Associates, Inc.},
    pages={33975-34002},
    year={2024},
}

@inproceedings{Zhang2025SymDiff,
    title={Sym{D}iff: Equivariant diffusion via stochastic symmetrisation},
    author={Zhang, Leo and Ashouritaklimi, Kianoosh and Teh, Yee Whye and Cornish, Rob},
    city={Singapore},
    booktitle={Proceedings of the 13th International Conference on Learning Representations},
    publisher={Curran Associates, Inc.},
    pages={47651-47673},
    year={2025}
}

@inproceedings{Hua2023MuDiff,
    title={{MUD}iff: Unified Diffusion for Complete Molecule Generation},
    author={Chenqing Hua and Sitao Luan and Minkai Xu and Rex Ying and Jie Fu and Stefano Ermon and Doina Precup},
    city={Virtual},
    booktitle={Proceedings of the 2nd Learning on Graphs Conference},
    publisher={PMLR},
    volume={231},
    pages={},
    year={2023},
}

@inproceedings{Peng2023MolDiff,
    title={Mol{D}iff: Addressing the Atom-Bond Inconsistency Problem in 3{D} Molecule Diffusion Generation},
    author={Peng, Xingang and Guan, Jiaqi and Liu, Qiang and Ma, Jianzhu},
    city={Honolulu, Hawaii, United States of America},
    booktitle={Proceedings of the 40th International Conference on Machine Learning},
    publisher={PMLR},
    volume={202},
    pages={27611-27629},
    year={2023},
}

@inproceedings{Vignac2023MiDi,
    title={Mi{D}i: Mixed Graph and 3{D} Denoising Diffusion for Molecule Generation},
    author={Vignac, Clement and Osman, Nagham and Toni, Laura and Frossard, Pascal},
    city={Turin, Italy},
    booktitle={Proceedings of the Joint European Conference on Machine Learning and Knowledge Discovery in Databases},
    publisher={Springer},
    volume={},
    pages={560-576},
    year={2023},
    doi={10.1007/978-3-031-43415-0_33},
}

@inproceedings{Xu2023GeoLDM,
    title={Geometric Latent Diffusion Models for 3{D} Molecule Generation},
    author={Xu, Minkai and Powers, Alexander S and Dror, Ron O. and Ermon, Stefano and Leskovec, Jure},
    city={Hnolulu, Hawaii, United States of America},
    booktitle={Proceedings of the 40th International Conference on Machine Learning},
    publisher={PMLR},
    volume={202},
    pages={38592-38610},
    year={2023},
}

@misc{Cheng2025Quetzal,
    title={Scalable Autoregressive 3{D} Molecule Generation}, 
    author={Austin H. Cheng and Chong Sun and Alán Aspuru-Guzik},
    year={2025},
    eprint={2505.13791},
    archivePrefix={arXiv},
    primaryClass={cs.LG},
    url={https://arxiv.org/abs/2505.13791}, 
}

@article{Ramakrishnan2014QM9,
    title={Quantum Chemistry Structures and Properties of 134 kilo Molecules},
    author={Raghunathan Ramakrishnan and Pavlo O. Dral and Matthias Rupp and O. Anatole Von Lilienfeld},
    journal={Scientific Data},
    publisher={Nature Publishing Groups},
    volume={1},
    number={140022},
    year={2014},
    doi={10.1038/sdata.2014.22},
}

@article{wu2018moleculenet,
  title={{MoleculeNet}: a benchmark for molecular machine learning},
  author={Wu, Zhenqin and Ramsundar, Bharath and Feinberg, Evan N and Gomes, Joseph and Geniesse, Caleb and Pappu, Aneesh S and Leswing, Karl and Pande, Vijay},
  journal={Chem. Sci.},
  volume={9},
  number={2},
  pages={513-530},
  year={2018},
  publisher={Royal Society of Chemistry}
}

@article{Kim2015xyz2mol,
   title={Universal Structure Conversion Method for Organic Molecules: From Atomic Connectivity to Three-Dimensional Geometry},
   author={Yeonjoon Kim and Woo Youn Kim},
   journal={Bull. Korean Chem. Soc.},
   publisher={Korean Chemical Society},
   volume={36},
   issue={7},
   pages={1769-1777},
   year={2015},
   doi={10.1002/bkcs.10334},
}

@inproceedings{Ho2020DDPM,
    title={Denoising Diffusion Probabilistic Models},
    author={Jonathan Ho and Ajay Jain and Pieter Abbeel},
    city={Vancouver, Canada},
    booktitle={Advances in Neural Information Processing Systems},
    publisher={Curran Associates, Inc.},
    volume={9},
    pages={6840-6851},
    year={2020},
}

@inproceedings{Song2023EquiFM,
    title={Equivariant Flow Matching with Hybrid Probability Transport for 3{D} Molecule Generation},
    author={Yuxuan Song and Jingjing Gong and Minkai Xu and Ziyao Cao and Yanyan Lan and Stefano Ermon and Hao Zhou and Wei-Ying Ma},
    city={New Orleans, Louisiana, United States of America},
    booktitle={Advances in Neural Information Processing Systems},
    publisher={Curran Associates, Inc.},
    volume={36},
    pages={549-568},
    year={2023},
}

@inproceedings{Song2024GeoBFN,
    title={Unified Generative Modeling of 3{D} Molecules via Bayesian Flow Networks},
    author={Yuxuan Song and Jingjing Gong and Hao Zhou and Mingyue Zheng and Jingjing Liu and Wei-Ying Ma},
    city={Vienna, Austria},
    booktitle={Proceedings of the 12th International Conference on Learning Representations},
    publisher={Curran Associates, Inc.},
    pages={47851-47869},
    year={2024},
}

@inproceedings{Li2025Geo2Seq,
    title={Geometry Informed Tokenization of Molecules for Language Model Generation},
    author={Li, Xiner and Wang, Limei and Luo, Youzhi and Edwards, Carl and Gui, Shurui and Lin, Yuchao and Ji, Heng and Ji, Shuiwang},
    city={Vancouver, Canada},
    booktitle={Proceedings of the 42nd International Conference on Machine Learning},
    publisher={PMLR},
    volume={267},
    pages={36096-36128},
    year={2025},
}

@inproceedings{Feng2025UniGEM,
    title={Uni{GEM}: A Unified Approach to Generation and Property Prediction for Molecules},
    author={Feng, Shikun and Ni, Yuyan and yan, Lu and Ma, Zhi-Ming and Ma, Wei-Ying and Lan, Yanyan},
    city={Singapore},
    booktitle={Proceedings of the 13th International Conference on Learning Representations},
    publisher={Curran Associates, Inc.},
    pages={12824-12849},
    year={2025},
}

@inproceedings{Li2025InertialAR,
      title={Inertial{AR}: Autoregressive 3{D} Molecule Generation with Inertial Frames}, 
      author={Haorui Li and Weitao Du and Yuqiang Li and Hongyu Guo and Shengchao Liu},
      city={Seoul, South Korea},
      booktitle={Proceedings of the 43rd International Conference on Machine Learning},
      publisher={PMLR},
      volume={},
      pages={},
      year={2026},
}

@inproceedings{Fu2025Frag2Seq,
    title={Fragment and Geometry Aware Tokenization of Molecules for Structure-Based Drug Design Using Language Models},
    author={Fu, Cong and Li, Xiner and Olson, Blake and Ji, Heng and Ji, Shuiwang},
    city={Singapore},
    booktitle={Proceedings of the 13th International Conference on Learning Representations},
    publisher={Curran Associates, Inc.},
    pages={64203-64227},
    year={2025},
}

@misc{flamshepherd2023llm2mol,
    title={Language Models Can Generate Molecules, Materials, and Protein Binding Sites Directly in Three Dimensions as {XYZ}, {CIF}, and {PDB} Files}, 
    author={Daniel Flam-Shepherd and Alán Aspuru-Guzik},
    year={2023},
    eprint={2305.05708},
    archivePrefix={arXiv},
    primaryClass={cs.LG},
    url={https://arxiv.org/abs/2305.05708}, 
}

@misc{Lu2025Uni-3DAR,
    title={Unified Cross-Scale 3{D} Generation and Understanding via Autoregressive Modeling}, 
    author={Shuqi Lu and Haowei Lin and Lin Yao and Zhifeng Gao and Xiaohong Ji and Yitao Liang and Weinan E and Linfeng Zhang and Guolin Ke},
    year={2025},
    eprint={2503.16278},
    archivePrefix={arXiv},
    primaryClass={cs.LG},
    url={https://arxiv.org/abs/2503.16278}, 
}

@inproceedings{simm2021symmetryaware,
    title={Symmetry-Aware Actor-Critic for {3D} Molecular Design},
    author={Gregor N. C. Simm and Robert Pinsler and G{\'a}bor Cs{\'a}nyi and Jos{\'e} Miguel Hern{\'a}ndez-Lobato},
    city={Virtual},
    booktitle={Proceedings of the 9th International Conference on Learning Representations},
    publisher={Curran Associates, Inc.},
    pages={},
    year={2021},
}

@inproceedings{Song2021ScoreMatching,
    title={Score-Based Generative Modeling through Stochastic Differential Equations},
    author={Yang Song and Jascha Sohl-Dickstein and Diederik P Kingma and Abhishek Kumar and Stefano Ermon and Ben Poole},
    city={Virtual},
    booktitle={Proceedings of the 9th International Conference on Learning Representations},
    publisher={Curran Associates, Inc.},
    pages={},
    year={2021},
}

@misc{raw_qm9_data,
    title={"Quantum chemistry structures and properties of 134 kilo molecules" Figshare repository},
    author={Raghunathan Ramakrishnan and Pavlo O. Dral and Matthias Rupp and O. Anatole Von Lilienfeld},
    year={2019},
    license={CC0},
    url={https://springernature.figshare.com/collections/Quantum_chemistry_structures_and_properties_of_134_kilo_molecules/978904},
    note={Accessed 2025-12-16},
}

@misc{raw_geom_data,
    title={},
    author={David Ryan Koes},
    year={2024},
    license={CC0 1.0 Universal},
    url={https://bits.csb.pitt.edu/files/geom_raw/},
    note={Accessed 2026-01-08},
}

@misc{XYZ_by_symphony,
    title={Sampled 3{D} Molecular Structures from Generative Models (Symphony)},
    author={Ameya Daigavane and Song Eun Kim and Mario Geiger and Tess Smidt},
    year={2023},
    license={CC BY 4.0},
    url={https://figshare.com/s/a17ccface17f0c22f15a?file=42504807},
    note={Accessed 2025-12-15},
}

@article{Axelrod2022,
   title={{GEOM}, Energy-Annotated Molecular Conformations for Property Prediction and Molecular Generation},
   author={Simon Axelrod and Rafael Gómez-Bombarelli},
   journal={Sci. Data},
   publisher={Nature Research},
   volume={9},
   number={185},
   year={2022},
   doi={10.1038/s41597-022-01288-4},
}

@article{Morehead2024,
   title={Geometry-Complete Diffusion for 3{D} Molecule Generation and Optimization},
   author={Alex Morehead and Jianlin Cheng},
   journal={Commun. Chem.},
   publisher={Nature Research},
   volume={7},
   number={150},
   year={2024},
   doi={10.1038/s42004-024-01233-z},
}

@inproceedings{ma2024sit,
  title={Sit: Exploring Flow and Diffusion-Based Generative Models with Scalable Interpolant Transformers},
  author={Ma, Nanye and Goldstein, Mark and Albergo, Michael S and Boffi, Nicholas M and Vanden-Eijnden, Eric and Xie, Saining},
  city={Milano, Italy},
  booktitle={Proceedings of the 18th European Conference on Computer Vision},
  publisher={Springer},
  volume={},
  pages={23-40},
  year={2024},
}

@inproceedings{Schneuing2025,
   title={Multi-Domain Distribution Learning for de novo Drug design},
   author={Arne Schneuing and Ilia Igashov and Adrian W Dobbelstein and Thomas Castiglione and Michael Bronstein and Bruno Correia},
   city={Singapore},
   booktitle={Proceedings of the 13th International Conference on Learning Representations},
   publisher={Curran Associates, Inc.},
   pages={84997-85033},
   year={2025},
}

@inproceedings{Irwin2025,
   title={Semla{F}low - Efficient 3{D} Molecular Generation with Latent Attention and Equivariant Flow Matching},
   author={Ross Irwin and Alessandro Tibo and Jon Paul Janet and Simon Olsson},
   city={Mai Khao, Thailand},
   booktitle={Proceedings of the 28th International Conference on Artificial Intelligence and Statistics},
   publisher={PMLR},
   volume={258},
   pages={3772-3780},
   year={2025},
}

@article{Reidenbach2025,
   title={Applications of Modular Co-Design for De Novo 3{D} Molecule Generation},
   author={Danny Reidenbach and Filipp Nikitin and Olexandr Isayev and Saee Paliwal},
   journal={Digit. Discov.},
   publisher={Royal Society of Chemistry},
   volume={5},
   issue={2},
   pages={754-768},
   year={2026},
}

@inproceedings{campbell2023trans,
  title={Trans-Dimensional Generative Modeling via Jump Diffusion Models},
  author={Campbell, Andrew and Harvey, William and Weilbach, Christian and De Bortoli, Valentin and Rainforth, Thomas and Doucet, Arnaud},
  city={New Orleans, Louisiana, United States of America},
  booktitle={Advances in Neural Information Processing Systems},
  publisher={Curran Associates, Inc.},
  volume={36},
  pages={42217-42257},
  year={2023},
}

@inproceedings{Simm2020,
   title={Reinforcement Learning for Molecular Design Guided by Quantum Mechanics},
   author={Gregor N C Simm and Robert Pinsler and José Miguel Hernández-Lobato},
   city={Virtual},
   booktitle={Proceedings of the 37th International Conference on Machine Learning},
   publisher={PMLR},
   volume={},
   pages={},
   year={2020},
}

@inproceedings{Ansel_PyTorch_2_Faster_2024,
    title={Py{T}orch 2: Faster Machine Learning Through Dynamic Python Bytecode Transformation and Graph Compilation},
    author={Ansel, Jason and Yang, Edward and He, Horace and Gimelshein, Natalia and Jain, Animesh and Voznesensky, Michael and Bao, Bin and Bell, Peter and Berard, David and Burovski, Evgeni and Chauhan, Geeta and Chourdia, Anjali and Constable, Will and Desmaison, Alban and DeVito, Zachary and Ellison, Elias and Feng, Will and Gong, Jiong and Gschwind, Michael and Hirsh, Brian and Huang, Sherlock and Kalambarkar, Kshiteej and Kirsch, Laurent and Lazos, Michael and Lezcano, Mario and Liang, Yanbo and Liang, Jason and Lu, Yinghai and Luk, CK and Maher, Bert and Pan, Yunjie and Puhrsch, Christian and Reso, Matthias and Saroufim, Mark and Siraichi, Marcos Yukio and Suk, Helen and Suo, Michael and Tillet, Phil and Wang, Eikan and Wang, Xiaodong and Wen, William and Zhang, Shunting and Zhao, Xu and Zhou, Keren and Zou, Richard and Mathews, Ajit and Chanan, Gregory and Wu, Peng and Chintala, Soumith},
    city={La Jolla, California, United States of America},
    booktitle={Proceedings of the 29th ACM International Conference on Architectural Support for Programming Languages and Operating Systems},
    publisher={ACM},
    volume={2},
    pages={},
    year={2024},
    doi={10.1145/3620665.3640366},
}

@article{Hunter2007,
    title={Matplotlib: A 2{D} Graphics Environment},
    author={Hunter, J. D.},
    journal={Comput. Sci. Eng.},
    publisher={IEEE COMPUTER SOC},
    volume={9},
    number={3},
    pages={90-95},
    year={2007},
}

@inproceedings{Yu2025RandomAugmentations,
    author={Yu, Qihang and He, Ju and Deng, Xueqing and Shen, Xiaohui and Chen, Liang-Chieh},
    title={Randomized Autoregressive Visual Generation}, 
    city={Honolulu, Hawaii, United States of America},
    booktitle={IEEE/CVF International Conference on Computer Vision (ICCV)},
    publisher={IEEE Computer Society Conference Publishing Services (CPS)},
    volume={},
    issue={},
    pages={18431-18441},
    year={2025},
}


\clearpage
\appendix
\section{Broader societal impact}\label{sec:impact}

This paper aims to advance the field of machine learning in general, and of \textit{de novo} molecular design in particular. While we recognize the potential misuse of our approach to generate harmful compounds, we emphasize that there are numerous beneficial applications, such as the development of new pharmaceuticals and safer agrochemicals. Importantly, the generation of molecules is only one step in a complex pipeline, requiring extensive validation, synthesis, and testing before any practical use, which mitigates immediate risks. Furthermore, the ethical deployment of such models should be guided by responsible practices, including rigorous oversight and adherence to safety standards. We do not foresee ethical concerns beyond those generally associated with generative models, but we stress the importance of responsible use to maximize societal benefit while minimizing risks.

\section{Flow matching}\label{sec:flow_matching}

Flow matching \citep{Lipman2023FlowMatching, Albergo2023NormalizingFlows} offers a conceptually simple alternative to diffusion models and often achieves faster inference due to the need for fewer integration steps during the inference process. It is a simulation-free training framework for generative modeling, where a time-dependent velocity field $\psi_t$ is learned to couple a tractable source distribution $p_0$ with a target distribution $p_1$.  
Typically, $p_0$ is chosen as a standard normal distribution, $p_0 = \mathcal{N}(0, I)$, and the transport is defined over a time interval $t \in [0,1]$. The transformation of $p_0$ into intermediate distributions $p_t$ is achieved \textit{via} the push-forward operation $\#$, resulting in a trajectory of probability distributions $p_t(\mathbf{x}_t)$ at time $t$. 
The flow $\psi_t$ is modeled using a parameterized function $\mathbf{v}_\theta(\mathbf{x}_t, t)$, which is trained to approximate the velocity field. Sampling from the target distribution $p_1$ is performed by drawing samples $\mathbf{x}_0 \sim p_0$ and integrating the ordinary differential equation (ODE) $\mathrm{d}\mathbf{x}_t = \mathbf{v}_\theta(\mathbf{x}_t, t)$.

\citet{Lipman2023FlowMatching} introduced simulation-free training by constructing conditional probability paths. 
A widely used choice for these paths is the linear interpolant \citep{Albergo2023NormalizingFlows}, defined as $\mathbf{x}_t = t\mathbf{x}_1 + (1-t)\mathbf{x}_0$, with the conditional target velocity $\mathbf{v}_t = \mathbf{x}_1 - \mathbf{x}_0$. 
The training objective, known as conditional flow matching (CFM), minimizes the $l_2$ regression loss:
$\mathcal{L}_\text{CFM} = \mathbb{E}[\lVert \mathbf{v}_\theta(\mathbf{x}_t, t) - \mathbf{v}_t \rVert]$.
Flow-based molecular generation models include for instance EquiFM \citep{Song2023EquiFM}, SemlaFlow \citep{Irwin2025}, Megalodon-FM \citep{Reidenbach2025}, and FlowMol3 \citep{Dunn2025FlowMol}. 
Independent coupling of $p_0$ and $p_1$ can result in high transport costs and intersecting conditional paths. 
These issues can be alleviated by employing optimal transport couplings between the source and target distributions, thereby improving convergence and enhancing flow performance \citep{tong2024improving}. 
In molecular generation, this has for example been approached through rigid alignment of noisy and true coordinates, combined with solving the assignment problem to minimize transport costs \citep{Dunn2025FlowMol}.

\section{Implementation details}\label{sec:implementation_details}

\subsection{Data processing}\label{subsec:data_processing}
The QM9 dataset was retrieved from the figshare repository of \citet{raw_qm9_data}, licensed under a Creative Commons Attribution-NonCommercial-ShareAlike 4.0 (CC BY-NC-SA 4.0) international license. The repository comprises 133,885 molecules and additionally provides a list of 3,054 molecule IDs for which OpenBabel failed to infer correct bonds from the XYZ geometry files. Following prior work \citep{Hoogeboom2022EDM, Cheng2025Quetzal}, we excluded these molecules. Because the dataset is distributed as XYZ files without explicit bond annotations, and our neighborhood-guidance strategy requires bond information, we reconstructed bonds for each molecule using RDKit (MolFromXYZBlock together with rdDetermineBonds). Bond assignment failed for 6810 molecules, yielding a processed dataset of 124,021 small molecules. We randomly split these into 100,000/11,619/12,402 molecules for the training/validation/test sets, respectively.
While some prior works on 3D molecular generation obtained QM9 \textit{via} the MoleculeNet project \citep{wu2018moleculenet} rather than the original repository \citep{raw_qm9_data}, we decided against using the MoleculeNet files. Although they include bond information, the procedure used to determine bonds is not documented, and a non-negligible fraction of molecules in that distribution are charged. This conflicts with the original QM9 specification \citep{Ramakrishnan2014QM9}, which describes the dataset as comprising \textit{neutral} molecules with up to nine heavy atoms (C, N, O, F), excluding hydrogen atoms.

The GEOM-Drugs dataset was obtained from the FlowMol project’s data repository \citep{Dunn2025FlowMol} as distributed by \citet{raw_geom_data}. This dataset is licensed under a Creative Commons Attribution 4.0 (CC BY 4.0) international license. The files provide a predefined train/validation/test split totaling 304,313 molecules, which we adopted unchanged. We removed 35 molecules that failed RDKit sanitization. For molecules comprising multiple disconnected fragments, we retained only the largest fragment by atom count. Following \citet{Vignac2023MiDi}, we used up to five low-energy conformations per molecule.
After preprocessing, the dataset contained 243,451/30,430/30,432 molecules and 1,172,436/146,394/146,788 conformations in the train/validation/test splits, respectively.

\subsection{System and software specification}\label{subsec:system_software_specification}
Training was performed on an AlmaLinux 9.7 system with an NVIDIA RTX Pro 6000 Blackwell GPU with 96 GB of GDDR7 VRAM and an AMD EPYC 9555 64-Core Processor. Inference and data analysis was performed on a Rocky Linux 9.7 system with an NVIDIA GeForce RTX 4090 GPU with 24 GB of GDDR6X VRAM and an AMD Ryzen 9 7950X 16-Core Processor.
We use CUDA 13.0 for GPU acceleration.
Our model is implemented in Python (v3.11.13) with PyTorch (v2.9.0) \citep{Ansel_PyTorch_2_Faster_2024}, PyTorch Geometric (v2.7.0) \citep{fey2019pyg}, and Pytorch Lightning (v2.5.5) \citep{Falcon_PyTorch_Lightning_2025}.
Model training was monitored with Tensorboard (v2.20.0). 
For chemical data processing we employed RDKit (v2025.9.1) \citep{rdkit_2025}. 
For visualization, we used Py3Dmol (v2.5.3) and matplotlib (v3.10.7) \citep{Hunter2007}.

\subsection{Model implementation}\label{subsec:model_implementation}
The basis of our implementation is the NanoGPT implementation of \citet{Karpathy2025nanoGPT}, which provides a concise implementation of the GPT-2 \citep{radford2019language} model. 
For positional encoding of the Cartesian coordinates, we employ Fourier features \citep{tancik2020fourier} as implemented in \citet{wang2025simplefold}.
We experimented with qk-normalization and rotary position embeddings \citep{su2024roformer}, but did not obtain improved results. 
We found additive pooling to be the most effective pooling strategy, but also tried mean and attention pooling.
We also tried local attention, defined by a radius graph, instead of full attention over all atom nodes, which led to performance degradation. 
Inclusion of charged states into the atom type vocabulary did not lead to model improvement.
Since our model architecture is not $SE(3)$-equivariant, we augment data samples by random rotation, which stabilizes the training. 
The flow head is the same adaptive layer-norm MLP as used in \citet{Li2024AutoregressiveImageGenerationWithoutVQ}, which was also used in the QUETZAL model \citep{Cheng2025Quetzal} for implementation of the diffusion head. 

For training the flow matching head, we sample random time steps analog to the SimpleFold model \citep{wang2025simplefold} from the following log-normal distribution:

\begin{equation}
    p(t) = 0.02 \mathcal{U}(0, 1) + 0.98\text{LN}(0.8, 1.7) 
\end{equation}
with
\begin{equation}
    \text{LN}(t;m, s) = \frac{1}{t(1-t)s\sqrt{2\pi}}\exp \frac{-(\text{logit}(t) -m)^2}{2s^2}
\end{equation}
where we set $m=0.8$ and $s=1.7$. With these settings, time steps are sampled more densely around $t=1$. We found that this sampling schema slightly improved training above time step sampling from a uniform distribution. 
For each conditional path, we sampled four random time steps, following \citet{Li2024AutoregressiveImageGenerationWithoutVQ}.
Using endpoint parametrization for construction of the conditional paths, like done in \citet{Dunn2025FlowMol}, led to performance degradation.

\subsection{Source-target split}\label{subsec:source_target_split}

\begin{figure}[htb]
    \centering
    \includegraphics[width=.95\linewidth]{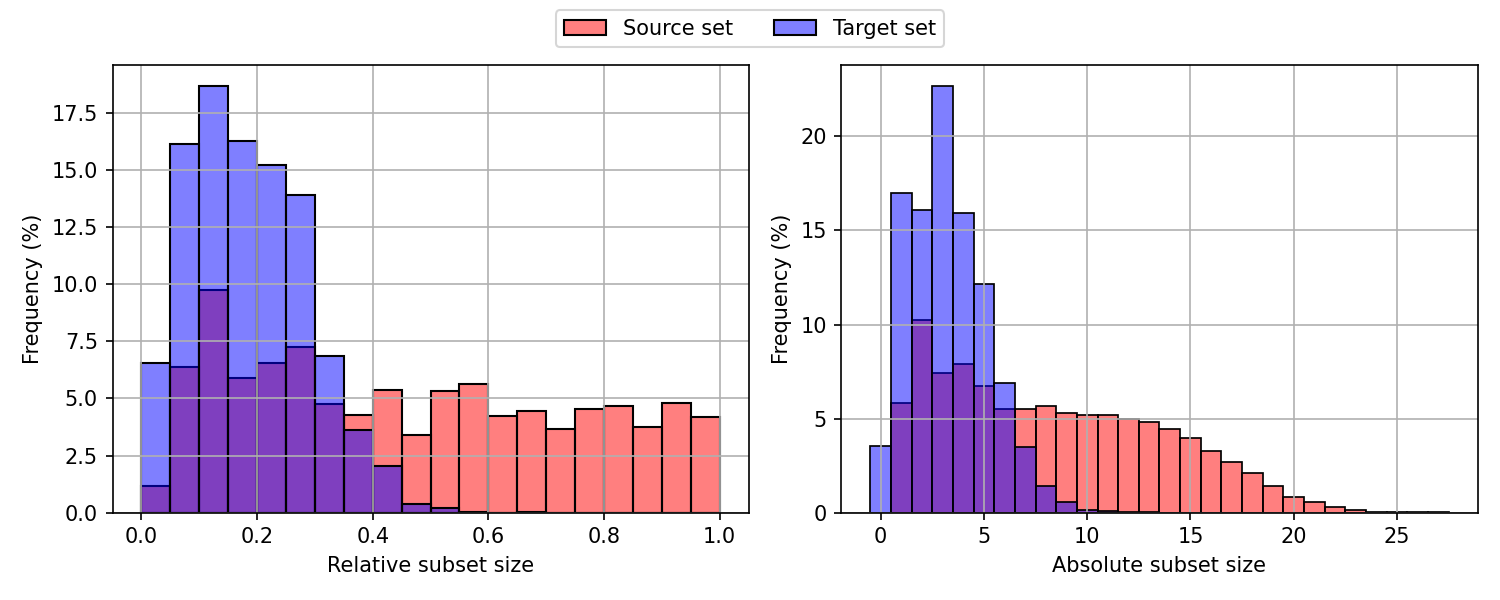}
    \caption{Relative and absolute size distribution of the source and target set w.r.t. the original graphs in the QM9 dataset. Sampling was performed with $\beta=1.5$ and $\gamma=0.45$.}
    \label{fig:subset_sampling_qm9}
\end{figure}

\begin{figure}[htb]
    \centering
    \includegraphics[width=.95\linewidth]{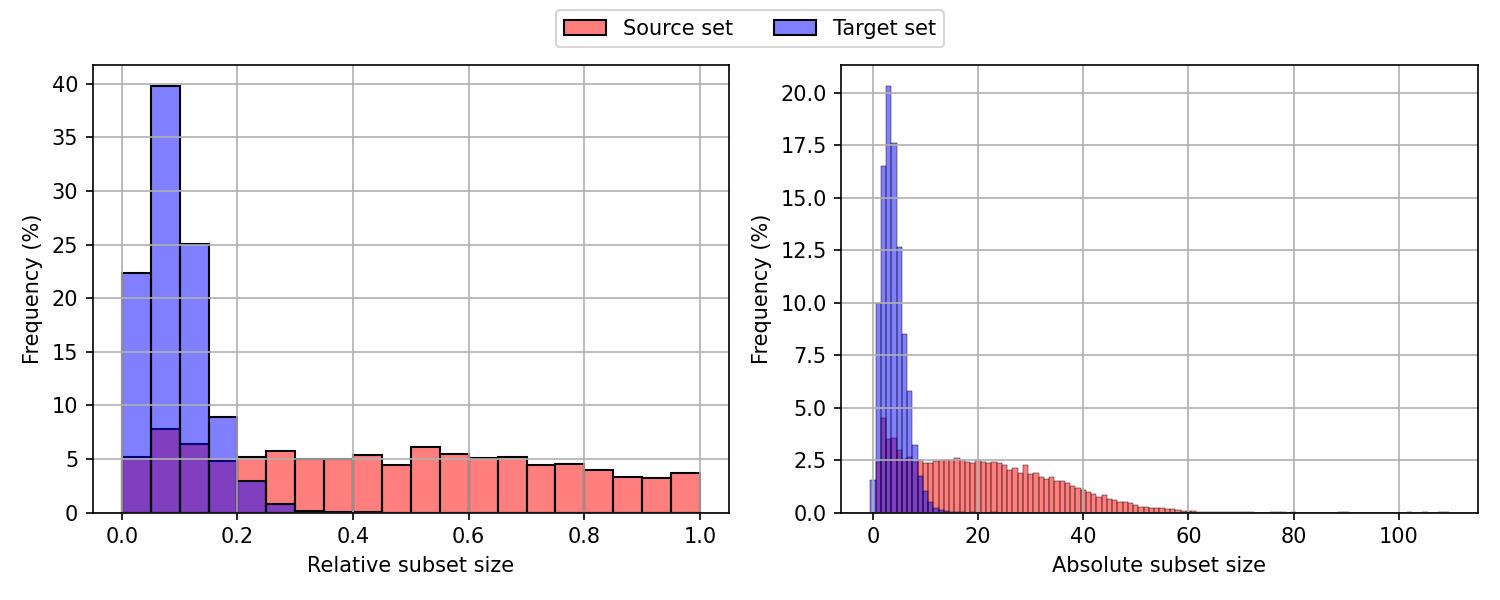}
    \caption{Relative and absolute size distribution of the source and target set w.r.t. the original graphs in the GEOM dataset. Sampling was performed with $\beta=1.5$ and $\gamma=0.45$.}
    \label{fig:subset_sampling_geom}
\end{figure}

The splitting procedure in Algorithm~\ref{alg:source_target} is dependent on the hyperparameters $\beta$ and $\gamma$ (see Section~\ref{sec:Methods}). 
Figure~\ref{fig:subset_sampling_qm9} and \ref{fig:subset_sampling_geom} report the absolute and relative (w.r.t. the original graphs) source and target set sizes for a batch of 25,000 randomly sampled training graphs. 
Across all runs, we used $\beta = 1.5$ and $\gamma = 0.45$. 
These settings produced an approximately uniform source set distribution with a slight bias toward smaller sets (corresponding to relative size). 
On QM9, the resulting target sets contain on average $\sim 18\%$ of the nodes of the original graphs. 
We also evaluated source-set distributions emphasizing mid-sized to larger sets, but near-uniform sampling performed best.

\subsection{Inference details}\label{subsec:inference_details}

\paragraph{Atom type sampling strategies}
The next atom type $a_{n+1}$ is sampled from the predicted distribution over possible atom types using a hybrid sampling strategy between greedy and multinomial sampling. 
Greedy sampling selects the atom type with the highest predicted probability, multinomial sampling draws samples from the predicted atom type distribution, and hybrid sampling greedily select the atom type with the highest probability if it is a hydrogen or the stop token, and otherwise samples from the multinomial distribution. 
An evaluation of the impact of this hybrid sampling strategy versus greedy and multinomial sampling can be found in the "Scaling and ablation" section (Appendix~\ref{subsec:scaling_and_ablations}).
We also investigated the impact of temperature scaling on the sampling of the next atom type. These results can be found in a separate section (Appendix~\ref{subsec:temperature_impact}). 

\paragraph{Integration scheme for continuous coordinates}
We found Euler integration with $N=60$ equidistant time steps to be best for integration of the flow ODE for the model trained on QM9 data. 
We also tried a Runge-Kutta-4 and an adaptive time step Dormand-Prince integrator, but did not observe any performance improvement. 
For the model trained on GEOM, the best results could be obtained by using the Euler-Maruyama integrator, which is described in the following.
Sampling from a flow-based generative model can be performed by integrating the ODE $\,\mathrm{d}\mathbf{x}_t = \mathbf{v}_\theta(\mathbf{x}_t, t)\,\mathrm{d}t$. Beyond deterministic sampling, the velocity field can be used to simulate the following reverse-time stochastic differential equation (SDE) \citep{ma2024sit}, thereby yielding a stochastic generative model:

\begin{equation}
    \mathrm{d}\mathbf{x}_t = \mathbf{v}_\theta(\mathbf{x}_t, t)\mathrm{d}t + \frac{1}{2} w(t) \mathbf{s}_\theta (\mathbf{x}_t, t, c)\mathrm{d}t + \sqrt{\tau \cdot w(t)} \mathrm{d}\mathbf{\bar{W}}_t
\end{equation}

where $\mathbf{\bar{W}}_t$ is a reverse-time Wiener process, $w_t$ is the diffusion coefficient, and $\mathbf{s}_\theta(\mathbf{x}_t, t)=\nabla \log p_t(\mathbf{x}_t)$ is the Stein score function. 
We employ the same Euler-Maruyama integrator as described in \citet{wang2025simplefold}. 
The score $\mathbf{s}_\theta(\mathbf{x}_t,t) = \frac{t \mathbf{v}_\theta(\mathbf{x}_t, t) -  \mathbf{x}_t}{1-t}$ can be derived from the velocity  and the diffusion coefficient is calculated as $w(t)=\frac{2(1-t)}{t+\eta}$, where $\eta=10^{-3}$ for numerical stability. In all of our experiments we use $\tau=0.3$, $N=60$ integration steps, and equidistant time step spacing.

\subsection{Hyperparameters}\label{subsec:hyperparameters}
Table~\ref{tab:hyperparameters_qm9} and \ref{tab:hyperparameters_geom} summarize the hyperparameters of our best-performing models on QM9 and GEOM. 
The models both have 165M parameters. 
The loss function was minimized with the AdamW \citep{loshchilov2019decoupled} optimizer with parameters $\beta_1=0.9$ and $\beta_2=0.95$. 
We further used gradient clipping to $1.0$ of the gradient norm.
We implemented a cosine annealing learning rate schedule with linear warm-up epochs and set the minimum learning rate to 10\% of the initial learning rate. 
Model training took 60 and 106 hours on QM9 and GEOM, respectively, given the hardware specifications listed above.

\begin{table}[htb!]
    \caption{Hyperparameters of the best performing NEAT model trained on the QM9 dataset.}
    \centering
    \resizebox{0.8\linewidth}{!}{
    \begin{tabular}{ll}
    \toprule
    \textbf{Parameter} & \textbf{Value} \\
    \midrule
        batch size & 2048 \\
        $\beta$ & 1.5 \\
        bias & true \\
        dropout & 0.1 \\
        epochs & 10,000 \\
        flow matching head hidden dimension  & 1,536 \\
        flow matching head layers & 6 \\
        flow matching noise standard deviation $\sigma$ & 1.4 \\
        $\gamma$ & 0.45 \\
        integrator & Euler \\
        learning rate, weight decay & $1.0 \times 10^{-4}$, $10^{-6}$ \\
        learning rate decay number of epochs (cosine annealing scheduling) & 10,000 \\
        learning rate minimum (cosine annealing scheduling) & $10\%$ \\
        learning rate warm-up number of epochs (cosine annealing scheduling) & 50 \\
        source set noise fraction $\delta$ & 0.2 \\
        source set noise standard deviation & 0.2 \\
        time step resampling & 4 \\
        time step sampling & logit normal \\
        transformer heads & 12 \\
        transformer hidden dimension & 768 \\
        transformer layers & 12 \\
        vocabulary size & 6 \\
    \bottomrule
    \end{tabular}
    }
    \label{tab:hyperparameters_qm9}
\end{table}

\begin{table}[htb!]
    \caption{Hyperparameters of the best performing NEAT model trained on the GEOM dataset.}
    \centering
    \resizebox{0.8\linewidth}{!}{
    \begin{tabular}{ll}
    \toprule
    \textbf{Parameter} & \textbf{Value} \\
    \midrule
        batch size & 512 \\
        $\beta$ & 1.5 \\
        bias & true \\
        dropout & 0.1 \\
        epochs & 1,000 \\
        flow matching head hidden dimension  & 1,536 \\
        flow matching head layers & 6 \\
        flow matching noise standard deviation $\sigma$ & 2.5 \\
        $\gamma$ & 0.45 \\
        integrator & Euler-Maruyama \\
        learning rate, weight decay & $2.5 \times 10^{-5}$, $10^{-6}$ \\
        learning rate decay number of epochs (cosine annealing scheduling) & 1,000 \\
        learning rate minimum (cosine annealing scheduling) & $10\%$ \\
        learning rate warm-up number of epochs (cosine annealing scheduling) & 10 \\
        source set noise fraction $\delta$ & 0.5 \\
        source set noise standard deviation & 0.5 \\
        time step resampling & 4 \\
        time step sampling & logit normal \\
        transformer heads & 12 \\
        transformer hidden dimension & 768 \\
        transformer layers & 12 \\
        vocabulary size & 17 \\
    \bottomrule
    \end{tabular}
    }
    \label{tab:hyperparameters_geom}
\end{table}

\begin{table}[htb!]
    \caption{Model scaling results on the QM9 dataset. Best in \textbf{bold}, second-best \underline{underlined}. The configuration selected for the final model is highlighted in \textit{italics}.}
    \centering
    \resizebox{\linewidth}{!}{
        \begin{tabular}{cccccccccccc}
        \toprule
         \makecell{transformer\\heads} &
         \makecell{transformer\\layers} &
         \makecell{transformer\\dimension} &
         \makecell{model parameters\\$\times 10^6$} &
         \makecell{atom\\stable} &
         \makecell{mol\\stable} &
         \makecell{lookup\\valid}&
         \makecell{lookup\\unique}&
         \makecell{xyz2mol\\valid} &
         \makecell{xyz2mol\\unique} & 
         \makecell{xyz2mol\\novel} 
         \\
         \midrule
        6 & 6 & 384 & 89.46 & 96.8 & 74.8 & 89.6 & 84.3 & 91.5 & 86.9 & 52.5 \\ 
        8 & 8 & 512 & 104.32 & 97.4 & 79.9 & 92.4 & 86.7 & 93.7 & 88.9 & 53.2 \\ 
        10 & 10 & 640 & 128.62 & \underline{97.6} & \underline{81.6} & \underline{93.2} & 88.0 & 94.2 & 90.0 & 53.3 \\ 
        \textit{12} & \textit{12} & \textit{768} & \textit{164.72} & \textbf{97.8} & \textbf{83.1} & \textbf{93.3} & \underline{88.8} & \textbf{95.2} & \textbf{91.6} & \underline{53.5} \\ 
        14 & 14 & 896 & 214.97 & \underline{97.6} & 81.1 & 93.0 & \textbf{89.1} & \underline{94.3} & \underline{91.2} & \textbf{54.5} \\ 
         \bottomrule
        \end{tabular}
    }
    \label{tab:scaling_qm9}
\end{table}

\begin{table}[htb!]
    \caption{Hyperparameter screening results on the GEOM dataset. Best in \textbf{bold}, second-best \underline{underlined}. The configuration selected for the final model is highlighted in \textit{italics}, chosen to balance optimal performance metrics with model size.}
    \centering
    \resizebox{\linewidth}{!}{
        \begin{tabular}{cccccccccc}
        \toprule
         \makecell{transformer\\heads} &
         \makecell{transformer\\layers} &
         \makecell{transformer\\dimension} &
         \makecell{FM head\\dimension} &
         \makecell{model parameters\\$\times 10^6$} &
         \makecell{atom\\stable} &
         \makecell{lookup\\valid}&
         \makecell{lookup\\unique}&
         \makecell{xyz2mol\\valid} &
         \makecell{xyz2mol\\unique} \\
         \midrule
         6  & 6  & 384  & 1536 & 89.46  & 79.4 & \underline{99.7} & \textbf{98.3} & 86.2 & 85.2 \\ 
         8  & 8  & 512  & 1536 & 104.32 & 80.4 & \textbf{99.8} & \underline{97.6} & 91.7 & 89.9 \\
         10 & 10 & 640  & 1536 & 128.62 & \textbf{81.5} & \textbf{99.8} & 96.7 & 94.0 & 91.4 \\
         \textit{12} & \textit{12} & \textit{768}  & \textit{1536} & \textit{164.72} & \textbf{81.5} & \underline{99.7} & 96.9 & 93.9 & \underline{91.6} \\
         14 & 14 & 896  & 1536 & 214.97 & \underline{80.9} & \underline{99.7} & 96.7 & \underline{94.4} & \textbf{91.8} \\
         16 & 16 & 1024 & 1536 & 281.72 & \underline{80.9} & \textbf{99.8} & 96.1 & \textbf{94.7} & 91.5 \\
         \bottomrule
        \end{tabular}
    }
    \label{tab:scaling_geom}
\end{table}

\begin{table}[!ht]
    \caption{Impact of centering, noising, and sampling strategies on the GEOM model. Best in \textbf{bold}, second-best \underline{underlined}. The configuration selected for the final model is highlighted in \textit{italics}.}
    \centering
    \resizebox{\linewidth}{!}{
    \begin{tabular}{cccccccccc}
        \toprule
        centering & noising & sampling & \makecell{atom\\stable} & \makecell{mol\\stable} & \makecell{lookup\\valid} & \makecell{lookup\\unique} & \makecell{xyz2mol\\valid} & \makecell{xyz2mol\\unique} & \makecell{xyz2mol\\novel} \\ 
        \midrule
        no & no & greedy & 81.1 & 0.5 & 97.9 & 97.4 & 74.2 & 73.7 & 73.7 \\ 
        no & no & multinomial & 78.1 & \underline{1.8} & 98.6 & 91.8 & 50.8 & 47.1 & 47.1 \\ 
        no & no & hybrid & 79.0 & 2.5 & 99.5 & 83.6 & 53.7 & 45.1 & 45.1 \\ 
        yes & no & greedy & \underline{82.3} & 1.6 & 99.2 & 91.7 & 87.8 & 80.9 & 80.9 \\ 
        yes & no & multinomial & 81.6 & 1.6 & 99.2 & \underline{97.7} & 85.4 & 84.2 & 84.2 \\ 
        yes & no & hybrid & \textbf{82.7} & \textbf{2.5} & \textbf{99.7} & 95.5 & 88.7 & 85.2 & 85.2 \\ 
        yes & yes & greedy & 79.9 & 0.1 & \underline{99.6} & 69.1 & \textbf{95.4} & 65.1 & 65.1 \\ 
        yes & yes & multinomial & 80.4 & 0.4 & 99.5 & \textbf{98.8} & 90.9 & \underline{90.5} & \underline{90.5} \\ 
        \textit{yes} & \textit{yes} & \textit{hybrid} & 81.5 & 0.6 & \textbf{99.7} & 96.9 & \underline{93.9} & \textbf{91.6} & \textbf{91.6} \\ 
        \bottomrule
    \end{tabular}
    }
    \label{tab:impact_of_strategies}
\end{table}

\begin{table}[!ht]
    \caption{Impact of additive vs. attention pooling on the QM9 model.}
    \centering
    \begin{tabular}{cccccccc}
        \toprule
        \makecell{pooling\\strategy} & \makecell{atom\\stable} & \makecell{mol\\stable} & \makecell{lookup\\valid} & \makecell{lookup\\unique} & \makecell{xyz2mol\\valid} & \makecell{xyz2mol\\unique} & \makecell{xyz2mol\\novel} \\ 
        \midrule
        additive & 97.8 & 83.1 & 93.3 & 88.8 & 95.2 & 91.6 & 53.5 \\ 
        attention & 95.0 & 63.2 & 83.6 & 81.4 & 89.5 & 87.4 & 56.0 \\ 
        \bottomrule
    \end{tabular}
    \label{tab:impact_of_attention_pooling}
\end{table}

\subsection{Scaling and ablations}\label{subsec:scaling_and_ablations}

We studied performance across various transformer sizes (Table~\ref{tab:scaling_qm9} and \ref{tab:scaling_geom}). 
While increasing the number of transformer layers from 6 to 12 improves performance on the QM9 dataset, gains are lost beyond 12 layers.
We further ablate key design choices like re-centering to the positional mean of the source set atoms, noising of the source set atoms, and different sampling strategies of the next atom type. 
Overall trends on the best GEOM model are summarized in Table~\ref{tab:impact_of_strategies}.
Centering consistently improves results across most metrics, particularly for multinomial and hybrid sampling strategies, regardless of whether noise is used. 
When greedy sampling is employed, centering still improves most metrics, with the exception of uniqueness. 
Overall, centering brings the strongest and most consistent improvements across all three modifications.
We observe a tradeoff between validity and uniqueness across different sampling strategies. 
Greedy sampling maximizes validity but sacrifices uniqueness. 
Multinomial sampling introduces stochasticity into the sampling process, which maximizes uniqueness at the cost of validity. 
Hybrid sampling strikes a good balance between greedy and multinomial strategies.
Noising has mixed effects: It does not greatly impact the lookup metrics (atom stability, molecule stability, lookup validity, and lookup uniqueness). 
However, it significantly improves xyz2mol metrics, especially when combined with multinomial or hybrid sampling strategies. 
The combination of noising and hybrid sampling appears to work synergistically, bringing gains that are less pronounced when either is used alone.
In summary, centering is the most impactful improvement, while noising and sampling work together to bring additional gains, particularly in xyz2mol metrics. 
A comparison of additive pooling and attention pooling on the best QM9 model is given in Table~\ref{tab:impact_of_attention_pooling}.

\section{Discussion of evaluation metrics}\label{sec:observations_on_eval_metrics}

\citet{Hoogeboom2022EDM} introduced a pipeline for generating molecular structures from atomic point clouds by using a look-up table of interatomic distances for single, double, and triple bonds. 
A bond is established if the interatomic distance is smaller than the corresponding table value plus a defined tolerance margin, starting with the shortest bond type and progressing sequentially to larger bond types. 
If no bond type matches the distance criteria, the atoms remain unbonded. 
Atom stability within this framework is defined by adherence to the octet or duet rule based on the assigned bonds. 
Molecular stability is achieved when all constituent atoms are stable, and molecular validity is determined by whether the molecule can be converted into a valid SMILES string after sanitization using RDKit.

A notable limitation of this pipeline is that it processes interatomic distances in a sequential manner, which does not guarantee convergence to an optimal solution.
Moreover, this pipeline does not consider aromatic bonds, which have lengths between single and double bonds. 
Consequently, aromatic molecules are often falsely categorized as unstable. 
For example, evaluating benzene with this pipeline results in 0\% atom stability. This occurs because all bonds in benzene are incorrectly assigned as "double" bonds, leading to a chemically invalid configuration where each carbon atom has five bonds (two double bonds to neighboring carbon atoms and one single bond to a hydrogen atom). 
Given the high prevalence of aromatic compounds in the GEOM-Drugs dataset, this issue leads to a surprisingly low molecular stability on the GEOM-Drugs test set and on the sets of molecules generated from models trained on GEOM-Drugs. 
Similar concerns have been raised by \citet{Daigavane2024Symphony} and \citet{Irwin2025}.

Another peculiarity of the pipeline is that the code used in \citet{Hoogeboom2022EDM} and related works \citep{Vignac2023MiDi, Cheng2025Quetzal} reduces all assigned bonds to single bonds when evaluating the GEOM-Drugs dataset. 
This simplification allows many SMILES strings to pass the RDKit sanitization test, even though they contain carbon atoms with incorrect multiplicities.
To ensure fair comparison with other models, we have adhered to this widely used pipeline without modifications, but also evaluate our models using the xyz2mol metric, which is described in the following. 

The xyz2mol metric tests whether generated 3D atomic point clouds can be converted into valid SMILES after bond assignment \textit{via} RDKit’s rdDetermineBonds. 
We would like to point out that this metric is sensitive to the RDKit version used. 
We observe decreases of up to $5\%$ with RDKit versions later than v2023.3.3, likely due to stricter bond assignment. 
We use RDKit v2025.9.1 throughout this work, which was the latest available update at project start. 
We recomputed xyz2mol metrics where model weights or pre-generated samples were available. 
Literature-reported xyz2mol results with undocumented RDKit versions are marked accordingly. 
While xyz2mol appears more robust than lookup metrics, we advocate for alternative evaluation pipelines to improve the accuracy and reliability of molecular generator assessment.

\section{Additional results}\label{sec:additional_results}

\subsection{Molecular properties of generated molecules}\label{subsec:mol_props}
\begin{figure}[htb!]
    \centering
    \includegraphics[width=1.0\linewidth]{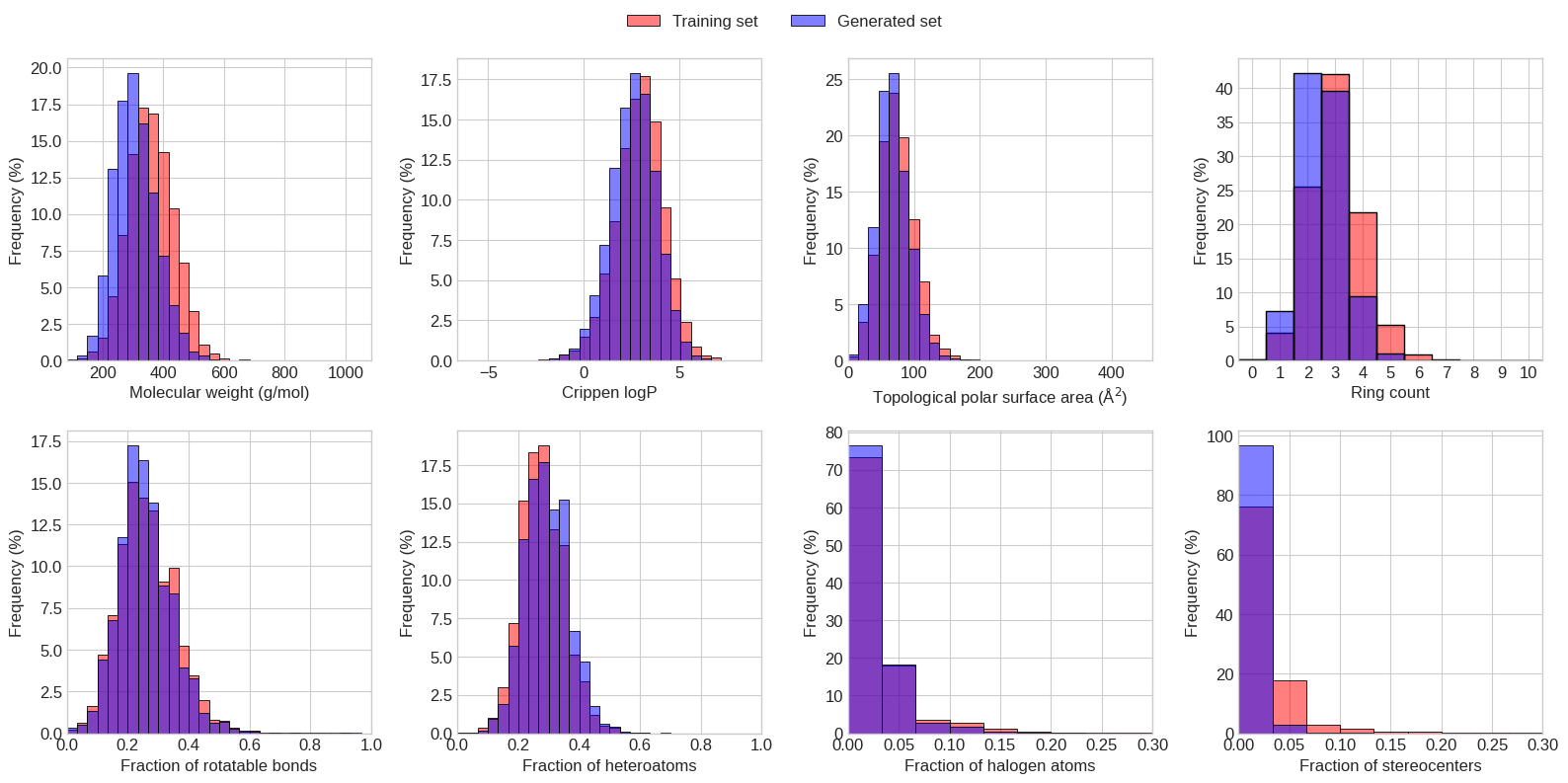}
    \caption{Comparison of generated molecules to the training set across eight molecular properties: molecular weight, Crippen logP, topological polar surface area, ring count, fraction of rotatable bonds, fraction of heteroatoms, fraction of halogen atoms, and fraction of stereocenters. Each panel shows overlaid histograms (training: red; generated: blue), reported as frequency in percent. Fractional properties are normalized by heavy atom count. Integer-valued properties use integer-centered bins.}
    \label{fig:molecular_properties_comparison}
\end{figure}

We compared the molecules generated by our GEOM model to the training set across eight molecular properties: molecular weight, Crippen logP, topological polar surface area (TPSA), ring count, fraction of rotatable bonds, fraction of heteroatoms, fraction of halogen atoms, and fraction of stereocenters (Figure~\ref{fig:molecular_properties_comparison}). 
Overall, the distributions of these properties are broadly similar between the two sets. 
The generated molecules exhibit a small bias toward simpler structures with lower molecular weight, fewer rings, and few stereocenters. 
None of the generated molecules were larger than the largest molecule found in the training set.

\subsection{Comparison to diffusion and Bayesian flow models}\label{subsec:comparison_to_diffusion}

Diffusion models \citep{Ho2020DDPM, Song2021ScoreMatching} operate by progressively corrupting the training data with random noise through a forward diffusion process, and training a neural network to reverse this corruption \textit{via} step-by-step denoising. 
During inference, new samples are generated by iteratively applying the learned reverse diffusion process to random noise.  
Prominent diffusion-based molecular generation models include EDM \citep{Hoogeboom2022EDM}, GeoLDM \citep{Xu2023GeoLDM}, GCDM \citep{Morehead2024}, SymDiff \citep{Zhang2025SymDiff}, and UniGEM \citep{Feng2025UniGEM}. 
Another alternative to 3D molecular generation are Bayesian flow networks like GeoBFN \citep{Song2024GeoBFN}.
We compare NEAT to these six models in addition to the autoregressive and flow models already discussed in the main body of this work.
Regarding QM9, we provide both lookup and xyz2mol metrics. 
For GEOM-Drugs, existing literature primarily reports stability and lookup metrics, and uniqueness and novelty are often omitted. 
Due to the unavailability of pre-trained weights for several baseline models, we report performance values directly from their original publications rather than performing independent re-evaluations. 
Consequently, due to this lack of available data, some metrics are missing in our comparison in Table~\ref{tab:results_qm9_diff}.
Given the observations discussed in Section~\ref{sec:observations_on_eval_metrics}, both stability and lookup metrics should be interpreted cautiously. 
Overall, NEAT is competitive with the reported diffusion models, outperforming them in some metrics, but not all. 
As detailed in the main text, autoregressive frameworks offer practical advantages like adaptive atom counts, faster inference, and straightforward prefix conditioning.

\begin{table}[htb!]
    \centering
    \caption{Performance comparison of NEAT with diffusion or Bayesian flow (middle block) and autoregressive (lower block) models trained on the QM9 and GEOM dataset, respectively. All metrics are reported as percentages. The first row reports results on the QM9 and GEOM test set. All other rows report results computed from 10,000 generated molecules. The \textbf{best} and \underline{second-best} entries are shown in bold and underlined, respectively. Entries for which generated molecules could be retrieved to be re-evaluated with our pipeline are marked with a single star (*). Entries for which the source code and model weights were available to generate molecules are marked with two stars (**). For these models and for NEAT, we report the mean and one standard deviation over three runs. Entries marked by a dagger ($\dagger$) were taken from \citet{Cheng2025Quetzal}. Metrics for all other entries were taken from the original publications. xyz2mol entries marked with a double dagger ($\ddagger$) are sourced from publications where the RDKit version is either v2023.3.3 or unspecified; these scores may decrease when re-evaluated with RDKit v2025.9.1.}
    \resizebox{0.9\linewidth}{!}{
        \begin{tabular}{lcccccccc}
            \toprule
            & \multicolumn{6}{c}{QM9} & \multicolumn{2}{c}{GEOM} \\
            \cmidrule(lr){2-7}\cmidrule(lr){8-9}
            &
             \makecell{atom\\stab.} &
             \makecell{mol.\\stab.} &
             \makecell{lookup\\valid} &
             \makecell{lookup\\unique} &
             \makecell{xyz2mol\\valid} &
             \makecell{xyz2mol\\unique} &
             \makecell{atom\\stab.} &
             \makecell{lookup\\valid}
             \\
             \midrule
             Test data       & 99.3 & 94.8 & 97.5 & 97.5 & 100 & 100 & 86.2 & 96.6\\
             \midrule
             EDM             & 98.7$^\dagger$ & 82.0$^\dagger$ & 91.9$^\dagger$ & 90.7$^\dagger$ & 86.7$^{\dagger, \ddagger}$ & 86.0$^{\dagger, \ddagger}$ & 81.3$^\dagger$ & 92.6$^\dagger$\\
             GeoLDM          & 98.9 & 89.4 & 93.8 & 92.7 & 91.3$^{\dagger, \ddagger}$ & 90.3$^{\dagger, \ddagger}$ & 84.4 & \underline{99.3}\\
             GCDM            & 98.7 & 86.0 & 94.9 & \underline{93.4} & n.a. & n.a. & \textbf{88.1} & 95.5\\
             GeoBFN          & \textbf{99.3} & \textbf{93.3} & \underline{96.9} & 92.4 & n.a. & n.a. & 86.2 & 91.7\\
             SymDiff         & 98.9 & 89.7 & 96.4 & \textbf{94.1} & 92.8$^{\dagger, \ddagger}$ & \underline{91.4}$^{\dagger, \ddagger}$ & 86.2 & \underline{99.3}\\
             UniGEM          & \underline{99.0} & 89.8 & 95.0 & 93.2 & n.a. & n.a. & 85.1& 98.4\\
             \midrule
             G-SchNet*       & 93.7 & 63.6 & 80.1 & 74.7 & 73.0 & 70.6 & n.a. & n.a.\\
             G-SphereNet*    & 67.8 & 14.0 & 16.4 & 3.8  & 37.7 & 7.2  & n.a.& n.a.\\
             Symphony*       & 90.8 & 43.9 & 68.1 & 66.5 & 77.0 & 75.2 & n.a.& n.a.\\
             Geo2Seq         & 98.9 & \underline{93.2} & \textbf{97.1} & 81.7 & n.a. & n.a. & 82.5 & 96.1 \\
             QUETZAL**       & 98.3 & 87.5 & 94.6 & 90.1 & \underline{93.7} & 89.1 & \underline{86.7} & 95.7 \\
                             & $\pm$ 0.1 & $\pm$ 0.1 & $\pm$ 0.1 & $\pm$ 0.2 & $\pm$ 0.3 & $\pm$ 0.2 & $\pm$ 0.1 & $\pm$ 0.1 \\
             \rowcolor{gray!15} NEAT (ours)     & 97.8 & 83.1 & 93.3 & 88.8 & \textbf{95.2} & \textbf{91.6} & 81.5 & \textbf{99.7}\\
             \rowcolor{gray!15}                & $\pm$ 0.1 & $\pm$ 0.1 & $\pm$ 0.1 & $\pm$ 0.3 & $\pm$ 0.2 & $\pm$ 0.3 & $\pm$ 0.1 & $\pm$ 0.2 \\
             \bottomrule
        \end{tabular}
    }
    \label{tab:results_qm9_diff}
\end{table}

\subsection{Failure modes}\label{subsec:failure_modes}
\begin{figure}[htb!]
    \centering
    \includegraphics[width=0.9\linewidth]{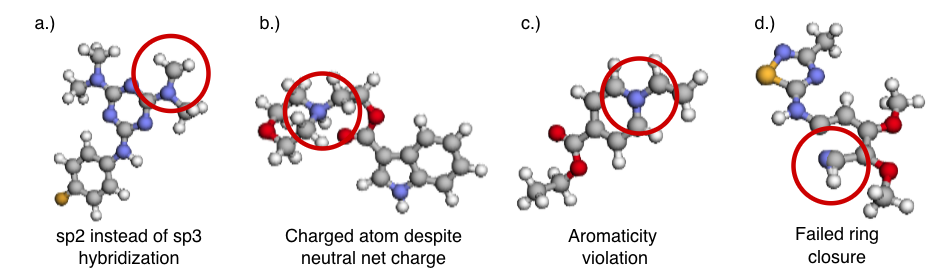}
    \caption{Qualitative illustration of possible failure modes.}
    \label{fig:failure_modes}
\end{figure}

We visually inspected generated molecules of our best GEOM model that failed xyz2mol bond assignment, examples are shown in Figure~\ref{fig:failure_modes}. 
Most issues were attributed to incorrect valency, such as issues with $sp^3$/$sp^2$ carbons, nitrogen atoms with four bonds (which fail under the assumption of a net charge of 0), or violations of aromaticity. 
In some cases, ring closure failed. 

\subsection{Impact of not using an SE(3)-equivariant model}\label{subsec:se3_impact}

NEAT uses a standard transformer, thereby benefiting from highly optimized building blocks, but it is not SE(3)-equivariant. 
We address this limitation with random rotation augmentations during training. 
We quantify the effect of rotations on prediction outcomes as follows.
We begin with a single carbon atom positioned at the origin of the coordinate system and iteratively add atoms to construct a molecule. 
At each iteration, we sample $N = 32$ rotation matrices to generate $N$ rotated versions of the partial molecule.
For each rotated version, we compute its latent representation and derive the multinomial distribution over atom types. 
The next atom type is selected as the one with the highest probability and most frequent occurrence across the $N$ rotated versions.
Given the selected atom type, we compute distributions over atomic positions for all $N$ rotated molecules. 
To achieve this, we sample an initial 3D Gaussian distribution with $K = 256$ points as the source distribution for flow matching. This distribution is rotated into the respective frames defined by the $N$ rotation matrices. 
Flow matching then produces target distributions, which are rotated back to the reference frame using the inverse operation.
The atom position for the next iteration is sampled randomly from the points in the aligned target distributions.

\begin{figure}[htb!]
    \centering
    \includegraphics[width=0.75\linewidth]{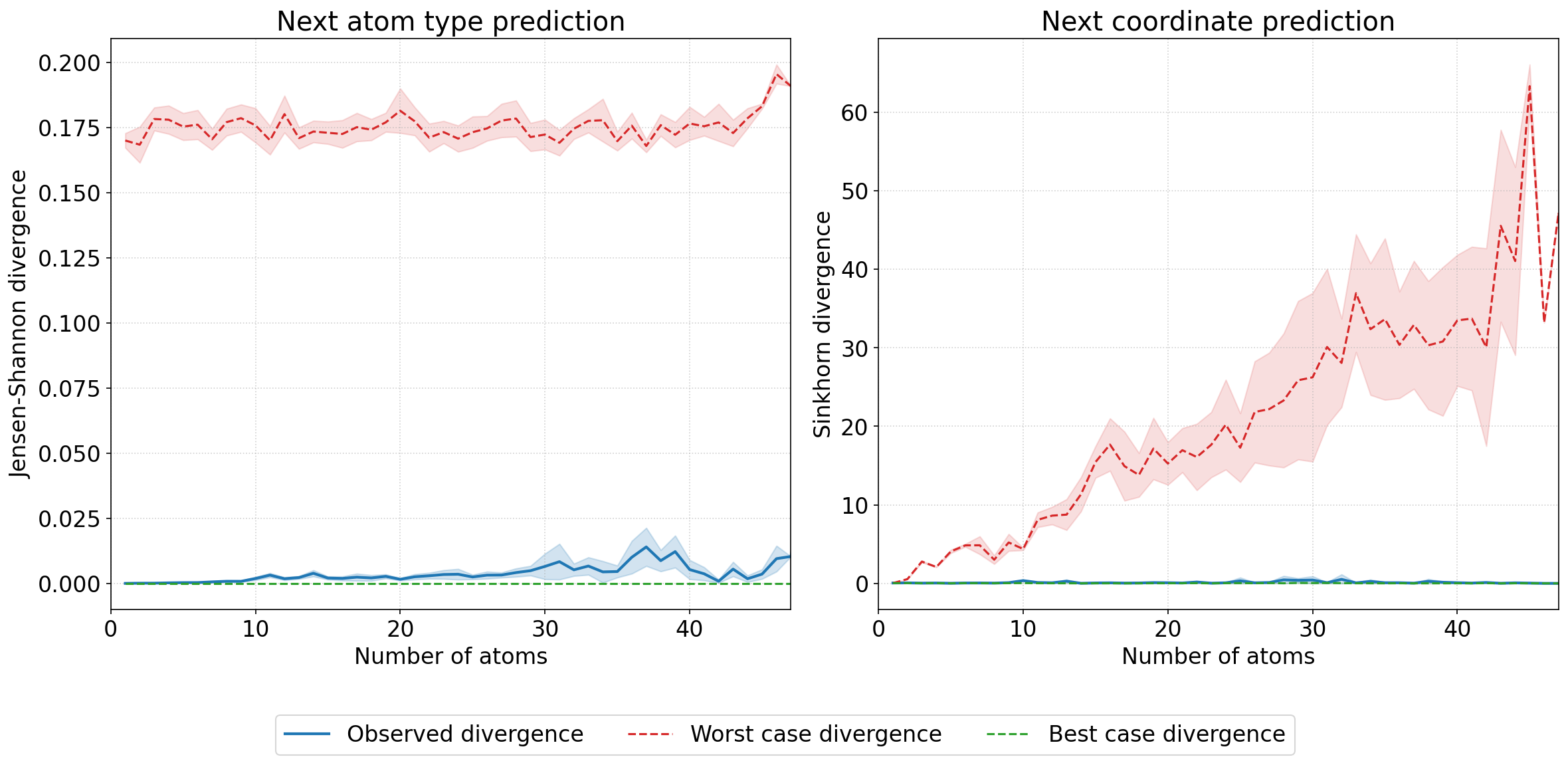}
    \caption{Mean pairwise Jensen-Shannon divergence of next atom-type distributions (left) and mean pairwise Sinkhorn divergence of next-coordinate distributions (right) across $N=32$ rotated versions of partial molecules, plotted against the number of atoms in the molecule. The shaded regions represent the 90\% confidence intervals (CI) across 10 repeated runs.}
    \label{fig:divergences}
\end{figure}

\begin{figure}[htb!]
    \centering
    \includegraphics[width=0.75\linewidth]{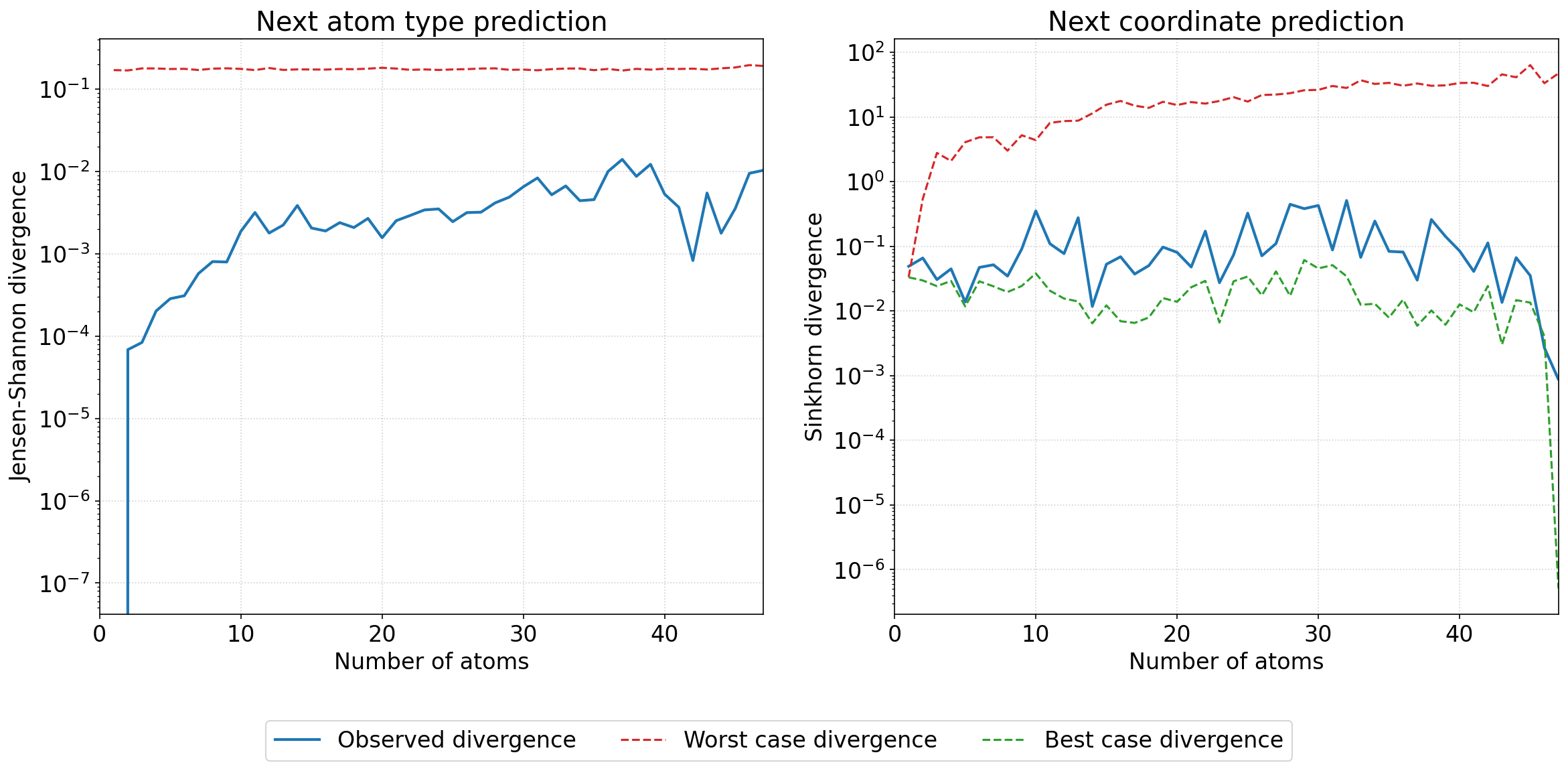}
    \caption{Log-scaled pairwise Jensen-Shannon divergence of next atom-type distributions (left) and mean pairwise Sinkhorn divergence of next-coordinate distributions (right) across $N=32$ rotated versions of partial molecules, plotted against the number of atoms in the molecule. Confidence intervals (CI) are omitted for clarity.}
    \label{fig:divergences_log}
\end{figure}

We use the following metrics for evaluation.
For atom type prediction, we calculate the pairwise Jensen-Shannon divergence between the multinomial distributions of atom types across the $N$ rotated molecules and report the mean divergence.
For atomic coordinates, we compute the pairwise Sinkhorn divergence between the $N$ re-aligned next-coordinate distributions and report the mean divergence. 
The experiment is repeated ten times, and we report the mean and 90\% confidence interval (CI) of the evaluation metrics, as shown in Figure~\ref{fig:divergences} and \ref{fig:divergences_log}.
To contextualize the scaling of these metrics, we also provide lower and upper bounds.
For atom type prediction, the upper bound is modeled with randomly sampled multinomial distributions, representing a dummy classifier.
The lower bound is a divergence of zero, corresponding to perfect SE(3) equivariance.
For atomic coordinate prediction, the upper bound is calculated as the average Sinkhorn divergence between $N$ target distributions without realignment back to the reference frame.
The lower bound is the average Sinkhorn divergence between $N$ target distributions without randomly rotating partial molecules but initialized with $N$ different 3D Gaussian distributions as source distribution for flow matching.

As shown in Figure~\ref{fig:divergences} and \ref{fig:divergences_log}, we observe only small variations in the mean pairwise Jensen-Shannon divergences of next atom-type distributions and the mean pairwise Sinkhorn divergences of next-coordinate distributions. 
The Jensen-Shannon divergence appears to slightly increase with the number of atoms, but remains relatively low compared to the worst-case upper bound. 
While the worst-case upper bound of the Sinkhorn divergence increases significantly with the number of atoms, the observed divergence remains unaffected by atom count and stays close to the lower bound.
We conclude that the lack of architectural SE(3)-equivariance does not significantly impact the performance of our model. 
Although our architecture is not explicitly SE(3)-equivariant, the incorporation of random rotations during training effectively compensates for this limitation. 

\subsection{Impact of temperature scaling}\label{subsec:temperature_impact}
\begin{figure}[htb!]
    \centering
    \includegraphics[width=0.75\linewidth]{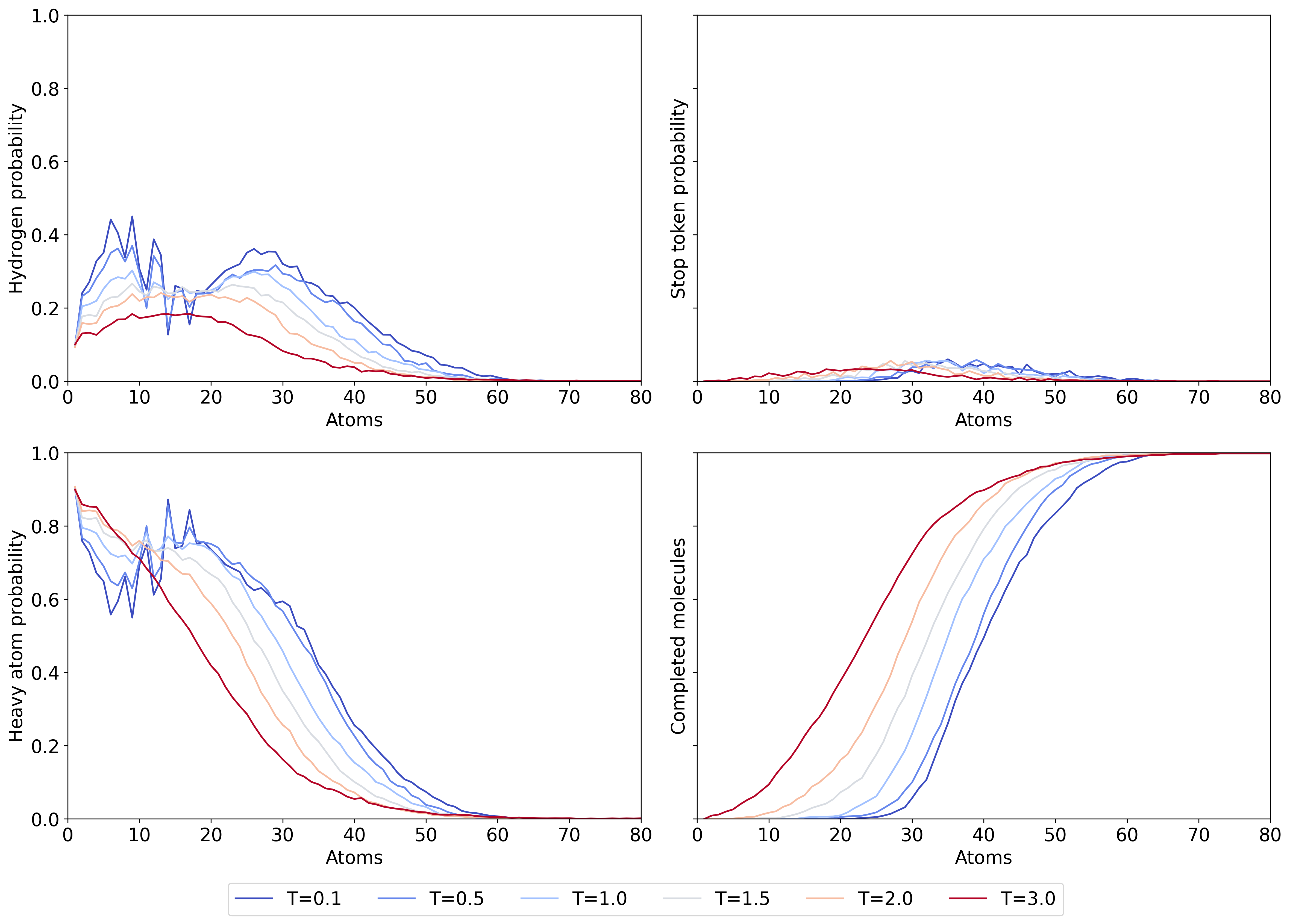}
    \caption{Mean probability of the next atom type, averaged over 1000 molecules during the generation process. The plot shows the probabilities of the atom type prediction head before temperature scaling, evaluated across different temperatures.}
    \label{fig:temperature_scaling}
\end{figure}

\begin{figure}[htb!]
    \centering
    \includegraphics[width=0.75\linewidth]{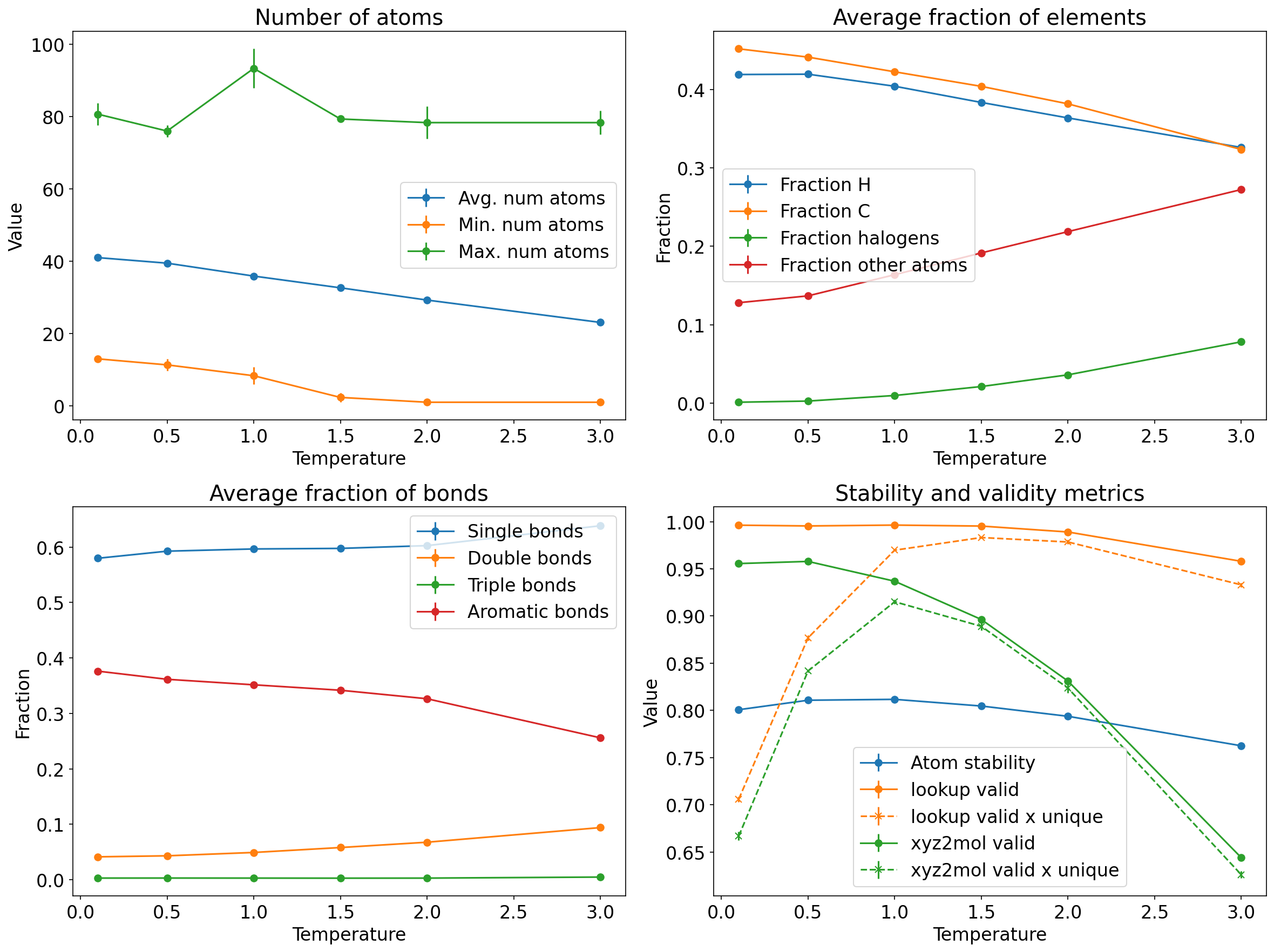}
    \caption{Correlation between evaluation metrics and various statistics of the generated data distribution w.r.t. temperature-scaled atom type sampling. For each temperature setting, we generate 10,000 molecules and report the mean and standard deviation (error bars) across three random seeds.}
    \label{fig:temperature_metrics}
\end{figure}

In language modeling, it is standard practice to use a temperature parameter to adjust the sharpness or flatness of the predicted next-token distribution, thereby influencing the generation process. 
As outlined in Section~\ref{subsec:inference_details}, our best GEOM model employs a hybrid sampling approach for predicting the next atom type, utilizing multinomial sampling for heavy atom tokens. 
To explore the impact of temperature scaling on the generated molecules, we modify this multinomial distribution by introducing a temperature parameter.
Figure~\ref{fig:temperature_scaling} demonstrates how temperature influences token probabilities during generation, revealing distinct patterns in the prediction of hydrogens and heavy atoms across low and high temperatures. 
Figure~\ref{fig:temperature_metrics} shows that higher temperatures decrease atom count, hydrogen and carbon content, and aromatic bonds, while increasing heteroatom content. 
Lower temperatures improve validity but reduce uniqueness, with $T=1$ offering the best balance of validity and uniqueness based on the xyz2mol metric. 
Additionally, molecule size decreases as temperature rises. 
These findings highlight the role of temperature in balancing molecular properties during generation.

\subsection{Impact of molecular size on validity}\label{subsec:molsize_impact}

\begin{figure}[htb!]
    \centering
    \includegraphics[width=0.75\linewidth]{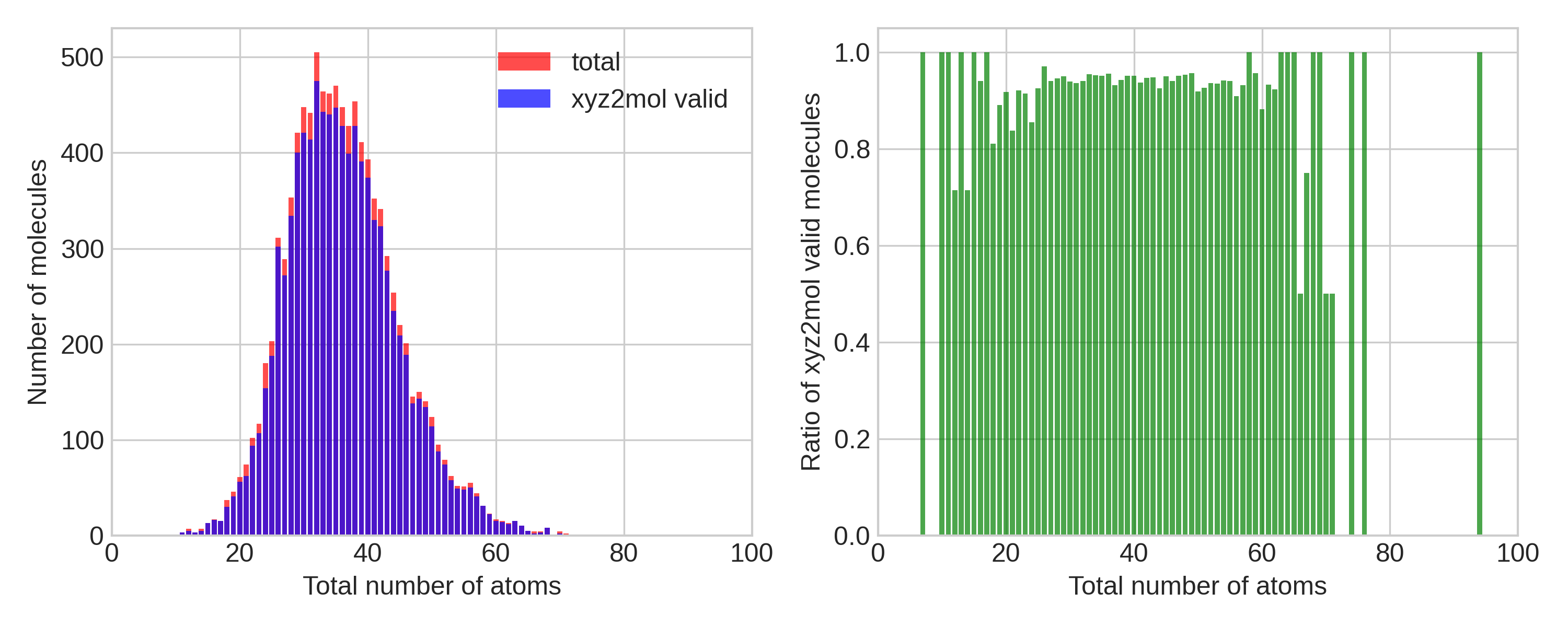}
    \caption{xyz2mol validity vs. total number of atoms. Left: distribution of number of atoms. The red bars represent the total number of molecules in each bin, the blue bars represent the absolute number of xyz2mol valid molecules. Right: ratio of xyz2mol valid molecules to the total number of molecules for each bin.}
    \label{fig:validity_molsize}
\end{figure}

\begin{figure}[htb!]
    \centering
    \includegraphics[width=0.75\linewidth]{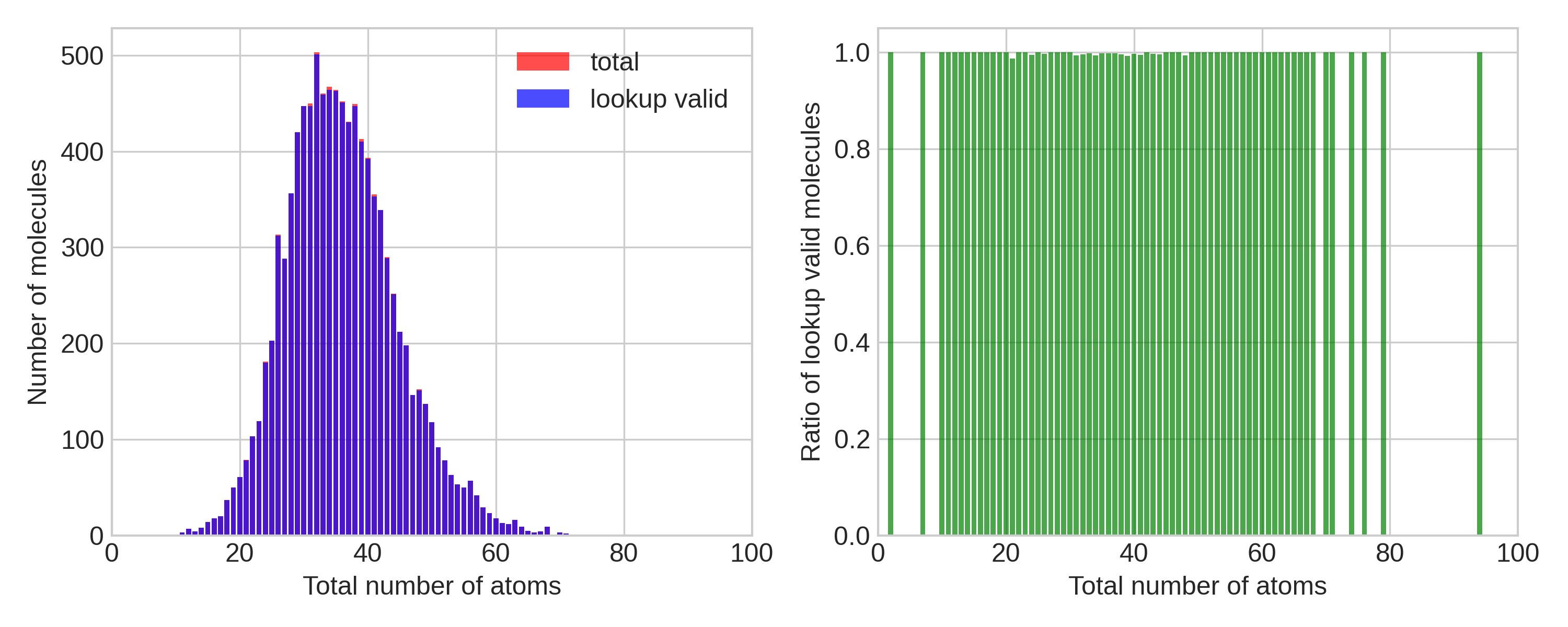}
    \caption{Lookup validity vs. total number of atoms. Left: distribution of number of atoms. The red bars represent the total number of molecules in each bin, the blue bars represent the absolute number of lookup valid molecules. Right: ratio of lookup valid molecules to the total number of molecules for each bin.}
    \label{fig:validity_molsize2}
\end{figure}

We binned molecules generated by our best GEOM-model based on their total number of atoms and computed three metrics for each bin: the total number of molecules, the absolute number of xyz2mol-valid molecules, and the ratio of xyz2mol-valid molecules to the total number of molecules. 
Results can be seen in Figure~\ref{fig:validity_molsize}. 
We find no definitive correlation between molecular size and validity, indicating that NEAT's performance does not degrade as molecular size increases within the range of the training data distribution, although we observe a slight variation in validity for molecules whose sizes deviate significantly from the mean. 
This trend could be due to two possible factors: either a sample size effect, as the tails of the distribution contain relatively few molecules, or genuine model errors. 
If the second explanation is correct, the observed variation cannot be attributed to error accumulation during autoregressive generation, as small molecules fail at similar rates to large ones. 
A more plausible explanation is that the model has encountered fewer examples of molecules at the extremes of the size distribution during training, compared to those closer to the mean, leading to a higher likelihood of mistakes for these less familiar cases.
We perform the same analysis of the ratio of valid molecules \textit{vs.} total number of atoms using the bond lookup pipeline instead of the xyz2mol pipeline, results are presented in Figure~\ref{fig:validity_molsize2}. 
We found an even less pronounced correlation: the Pearson correlation coefficient was 0.07, compared to the previous 0.12 observed with the xyz2mol pipeline. 

\FloatBarrier

\subsection{Impact of integration steps on baseline validity}\label{subsec:quality_runtime_tradeoff}

We investigate the impact of the number of integration steps on the molecular validity of the SemlaFlow \citep{Irwin2025} and FlowMol3 \citep{Dunn2025FlowMol} baselines. 
We report the average runtime and performance metrics over three independent runs for both models with three choices for the number of integration steps: a) the number recommended by the authors of SemlaFlow and FlowMol3, b) 60 (same as NEAT), and c) 15 (resulting in runtimes matching NEAT’s). 
In the latter setting, SemlaFlow and FlowMol3 achieve xyz2mol validities of 80.4\% and 55.5\%, respectively, compared to NEAT’s xyz2mol validity of 93.9\%. 
The results are summarized in Table~\ref{tab:quality_runtime_tradeoff}.

\begin{table}[H]
    \centering
    \caption{Average runtime and performance metrics for both SemlaFlow and FlowMol3 with three choices for the number of integration steps: a) the number recommended by the authors of SemlaFlow and FlowMol3, b) 60 (same as NEAT), and c) 15 (resulting in runtimes matching NEAT’s). Mean $\pm$ one standard deviation are reported over 3 independent runs.}
        \begin{tabular}{lcccc}
            \toprule
            Model & Steps & Runtime & \makecell{xyz2mol\\valid} & \makecell{xyz2mol\\unique} \\
            \midrule
            SemlaFlow & 100 & 1064 & $86.6 \pm 0.5$ & $86.5 \pm 0.5$ \\
            SemlaFlow & 60 & 662 & $85.5 \pm 0.1$ & $85.5 \pm 0.1$ \\
            SemlaFlow & 15 & 188 & $80.4 \pm 0.4$ & $80.4 \pm 0.4$ \\
            FlowMol3 & 250 & 3420 & $95.8 \pm 0.7$ & $95.7 \pm 0.7$ \\
            FlowMol3 & 60 & 855 & $92.2 \pm 0.3$ & $91.9 \pm 0.3$ \\
            FlowMol3 & 15 & 236 & $55.5 \pm 0.2$ & $55.5 \pm 0.2$ \\
            \rowcolor{gray!15} NEAT & 60 & 208 & $93.9 \pm 0.2$ & $91.6 \pm 0.2$ \\
            \bottomrule
        \end{tabular}
    \label{tab:quality_runtime_tradeoff}
\end{table}

\FloatBarrier

\section{FlowMol3 inpainting procedure}\label{sec:inpainting}

FlowMol3 \citep{Dunn2025FlowMol} represents a molecule as a graph $g=(X,A,C,E)$ with atom positions $X$, atom type $A$, charges $C$, and edges $E$. 
Given a fixed prefix $g^{\text{prefix}}_1$ with $n$ atoms, we sample a corresponding prefix state $g^{\text{prefix}}_0$ from the prior and construct the same straight interpolation path used during FlowMol3 training:

\begin{equation}
    g^{\text{prefix}}_t = (1-t) g^{\text{prefix}}_0 + t g^{\text{prefix}}_1
\end{equation}

At inference, we initialize the full molecule $g_0$ from the prior and define a binary mask $M$, where $M=1$ denotes the prefix atoms to be kept fixed and $M=0$ denotes the remaining atoms to be generated (edges are handled analogously). 
During integration, after each model update to $g_t$, we overwrite the masked prefix coordinates/features with the corresponding point on the prefix interpolation path:

\begin{equation}
    g_t[M] \leftarrow g^{\text{prefix}}_t
\end{equation}

Thus, the prefix follows the prescribed noised-to-clean path while the unmasked part is generated conditionally on it through the model dynamics. 
This provides a direct inference-time inpainting baseline for FlowMol3 under the same prefix constraints as our method, without requiring any additional training. 
For each of the 100 prefixes, we sample 1000 molecules. The results are reported in Table~\ref{tab:results_prefix}.

\section{Visualizations}\label{sec:visualizations}

\subsection{Examples of generated molecules}\label{subsec:examples_generated_mols}
\begin{figure}[htb]
    \centering
    \includegraphics[width=1.\linewidth]{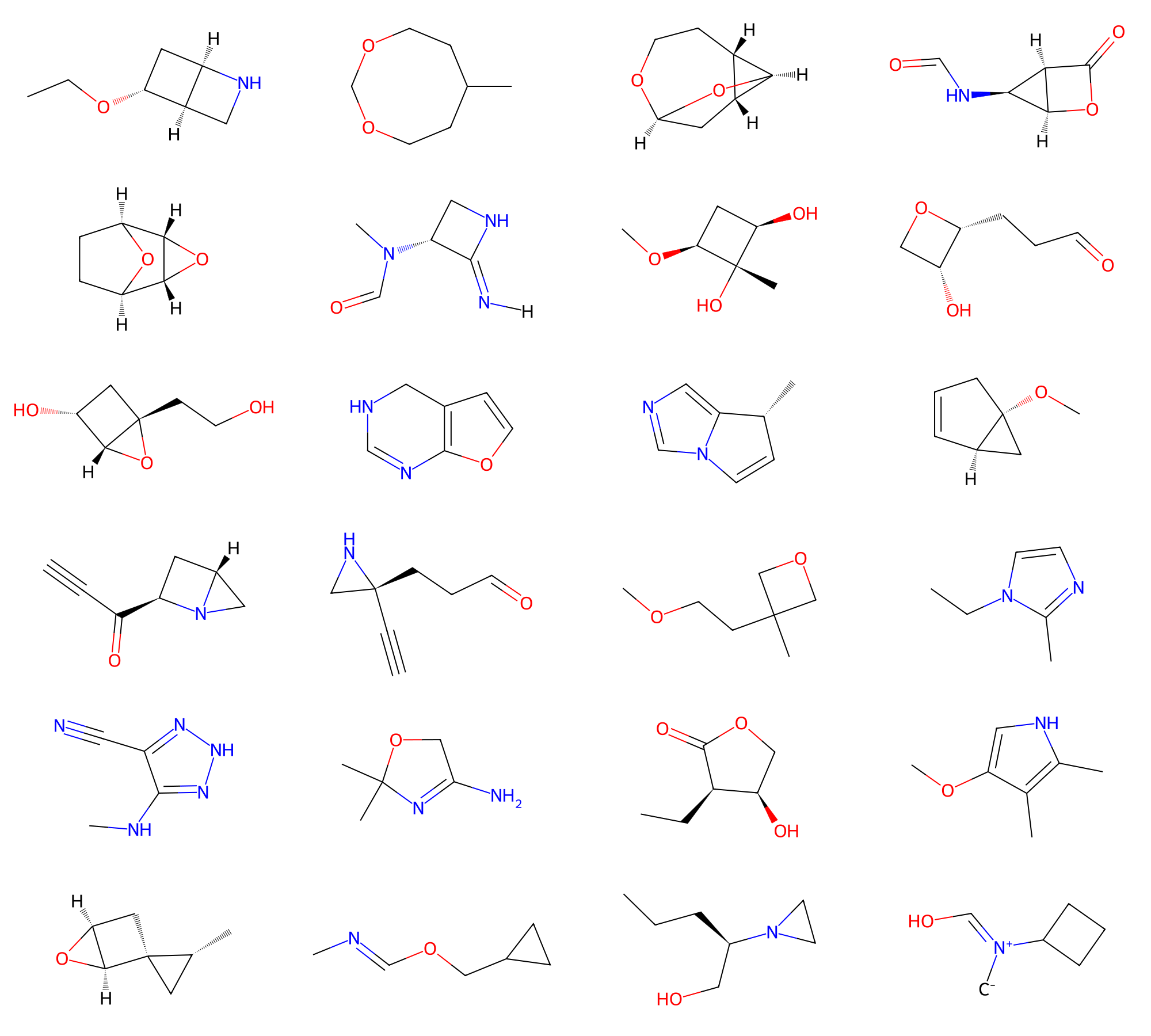}
    \caption{Randomly selected molecules generated by NEAT trained on the QM9 dataset. Implicit hydrogen atoms have been removed for the sake of clarity. 3D plots of the same molecules are shown in the next figure.}
    \label{fig:examples_qm9_2d}
\end{figure}

\begin{figure}[htb]
    \centering
    \includegraphics[width=0.8\linewidth]{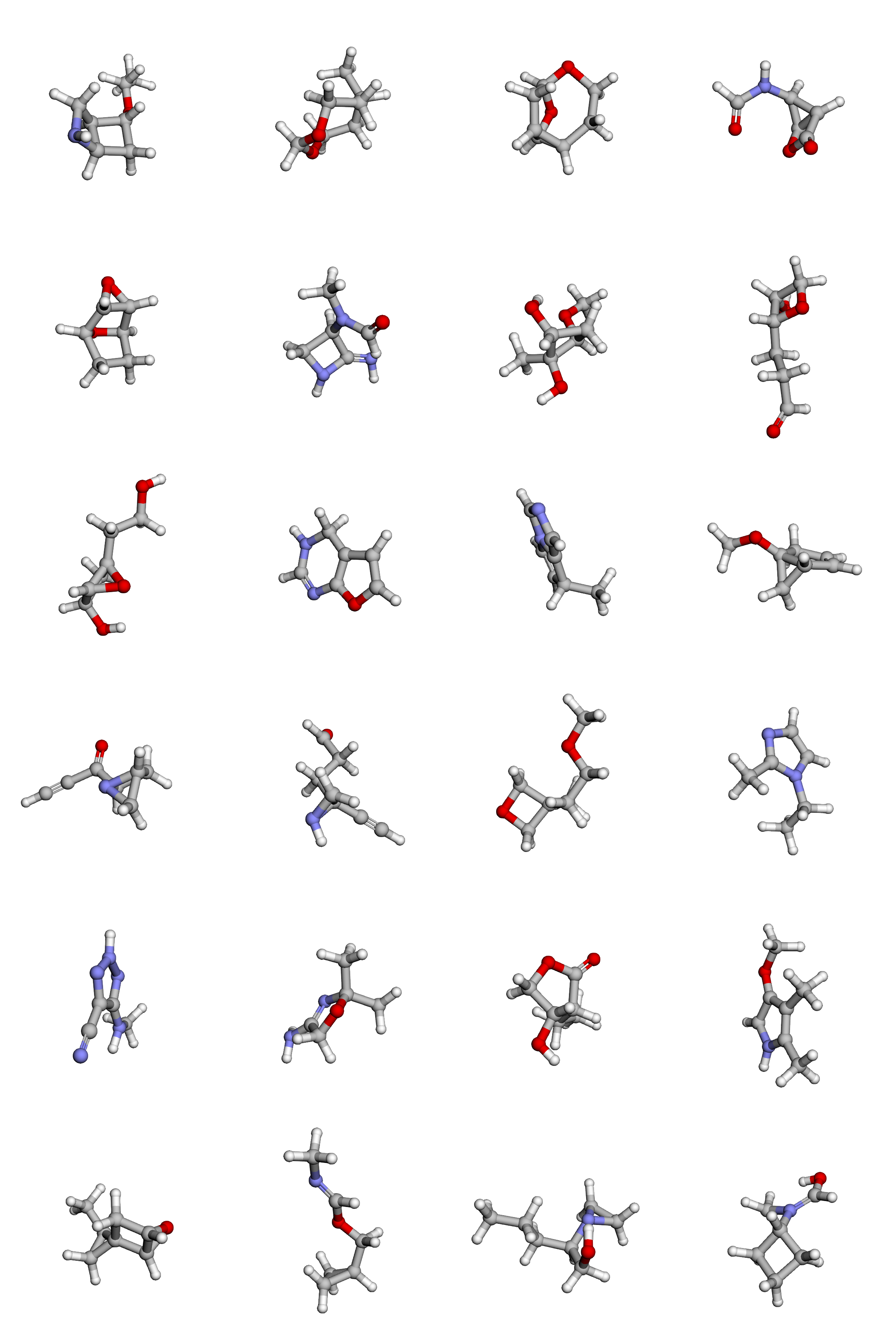}
    \caption{Randomly selected molecules generated by NEAT trained on the QM9 dataset (white: hydrogen, gray: carbon, blue: nitrogen, red: oxygen, brown: fluorine). 2D plots of the same molecules are shown in the previous figure.}
    \label{fig:examples_qm9_3d}
\end{figure}

\begin{figure}[htb]
    \centering
    \includegraphics[width=1.\linewidth]{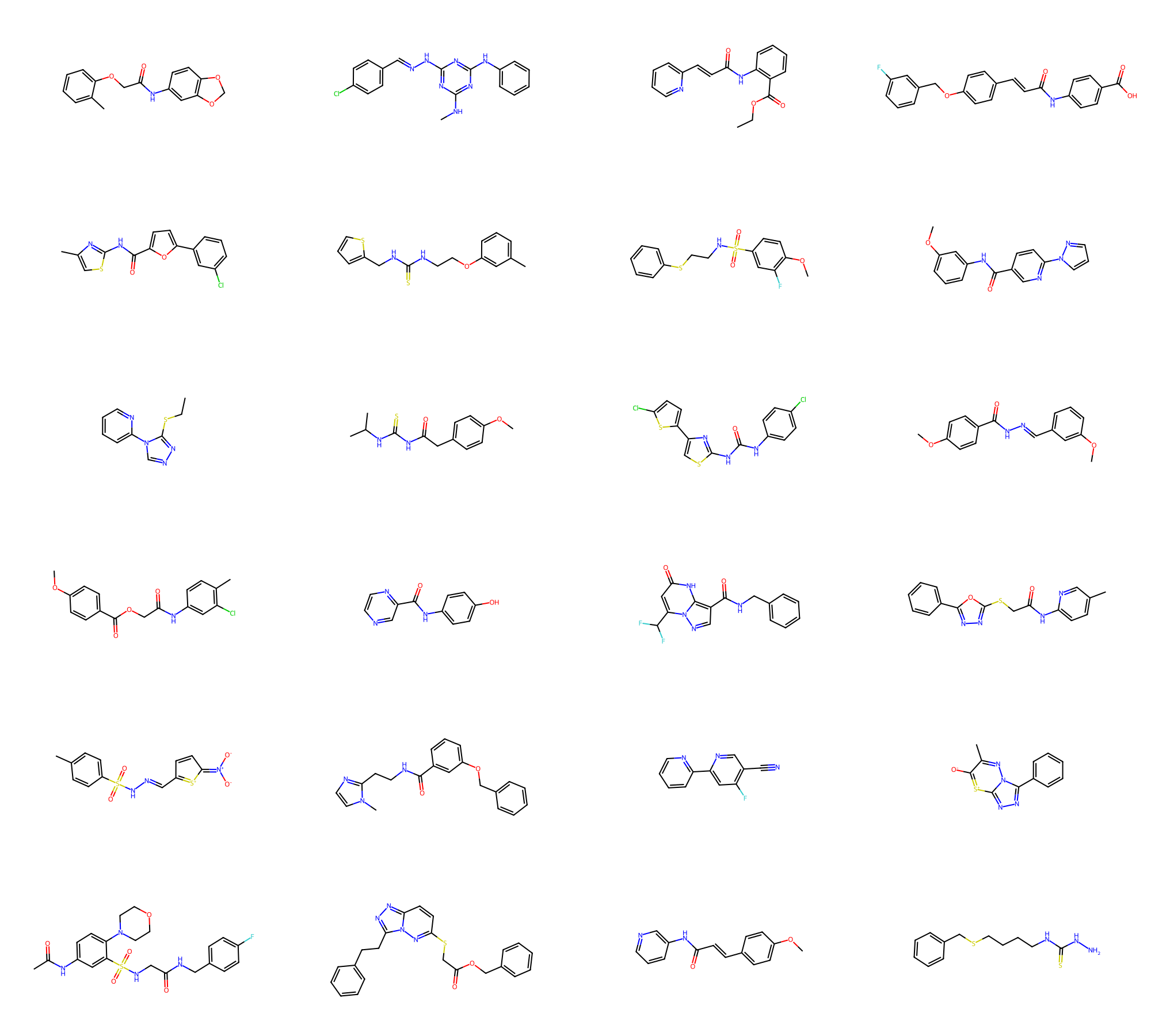}
    \caption{Randomly selected molecules generated by NEAT trained on the GEOM-Drugs dataset. Implicit hydrogen atoms have been removed for the sake of clarity. 3D plots of the same molecules are shown in the next figure.}
    \label{fig:examples_geom_2d}
\end{figure}

\begin{figure}[htb]
    \centering
    \includegraphics[width=0.8\linewidth]{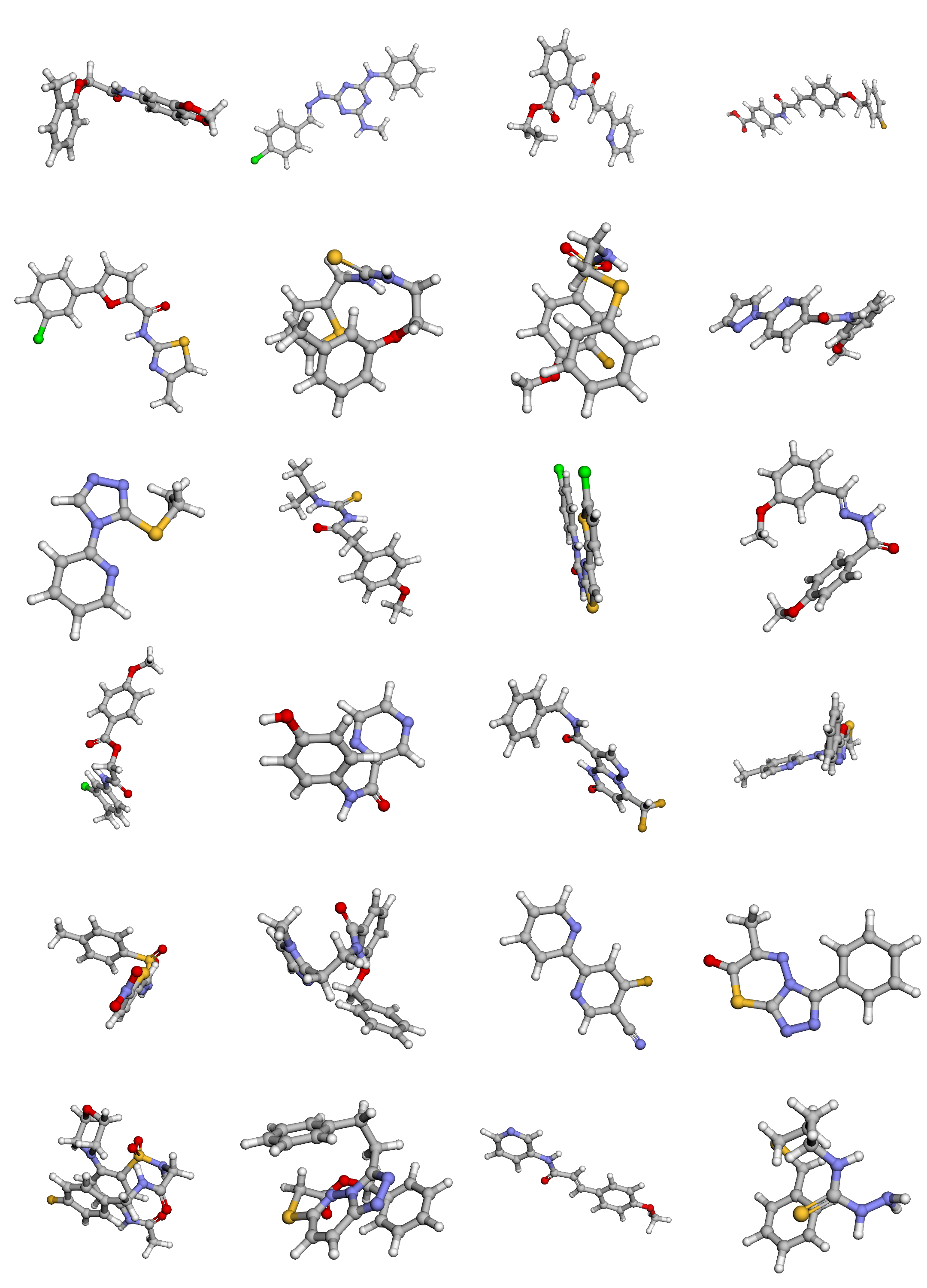}
    \caption{Randomly selected molecules generated by NEAT trained on the GEOM-Drugs dataset (white: hydrogen, gray: carbon, blue: nitrogen, red: oxygen, yellow: sulfur, brown: fluorine, green: chlorine). 2D plots of the same molecules are shown in the previous figure.}
    \label{fig:examples_geom_3d}
\end{figure}

Figure~\ref{fig:examples_qm9_2d}, \ref{fig:examples_qm9_3d}, \ref{fig:examples_geom_2d}, and \ref{fig:examples_geom_3d} show 2D and 3D plots of randomly selected molecules generated by NEAT trained on the QM9 and on the GEOM-Drugs datasets.

\subsection{Prefixes}\label{subsec:prefixes}
\begin{figure}[htb]
    \centering
    \includegraphics[width=1.0\linewidth]{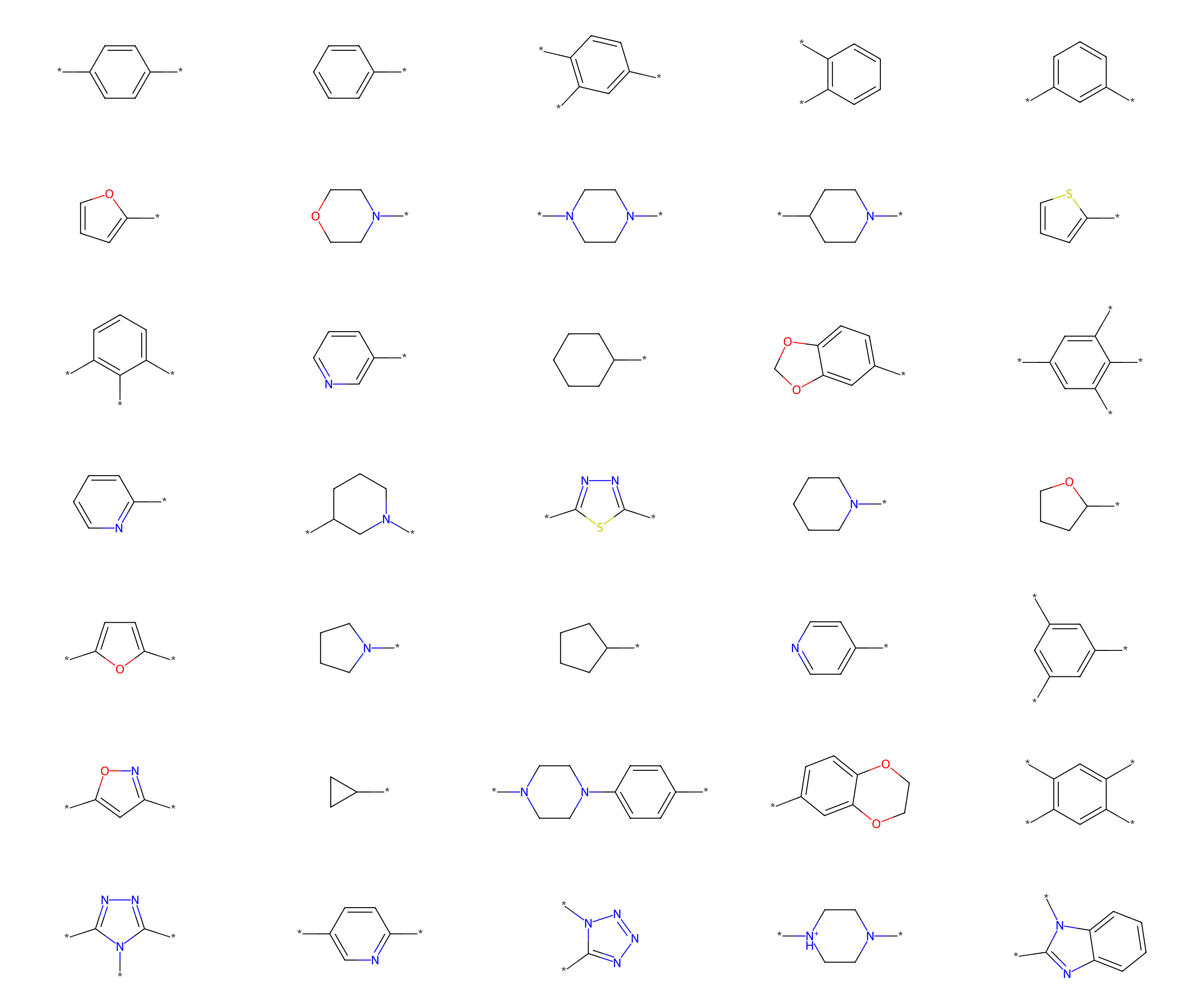}
    \caption{Prefixes used for the prefix completion task (1/3).}
    \label{fig:full_prefixes_1}
\end{figure}

\begin{figure}[htb]
    \centering
    \includegraphics[width=1.0\linewidth]{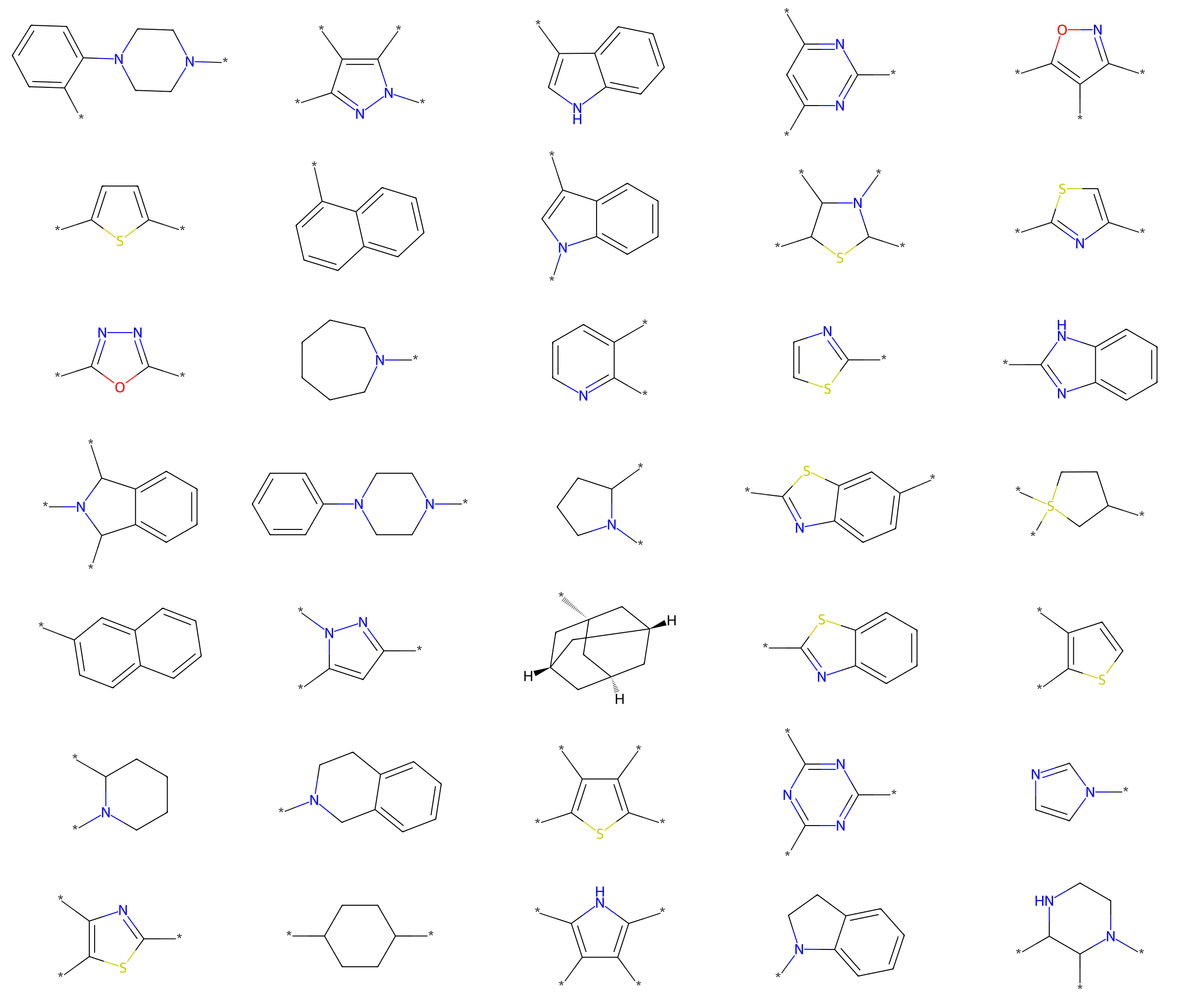}
    \caption{Prefixes used for the prefix completion task (2/3).}
    \label{fig:full_prefixes_2}
\end{figure}

\begin{figure}[htb]
    \centering
    \includegraphics[width=1.0\linewidth]{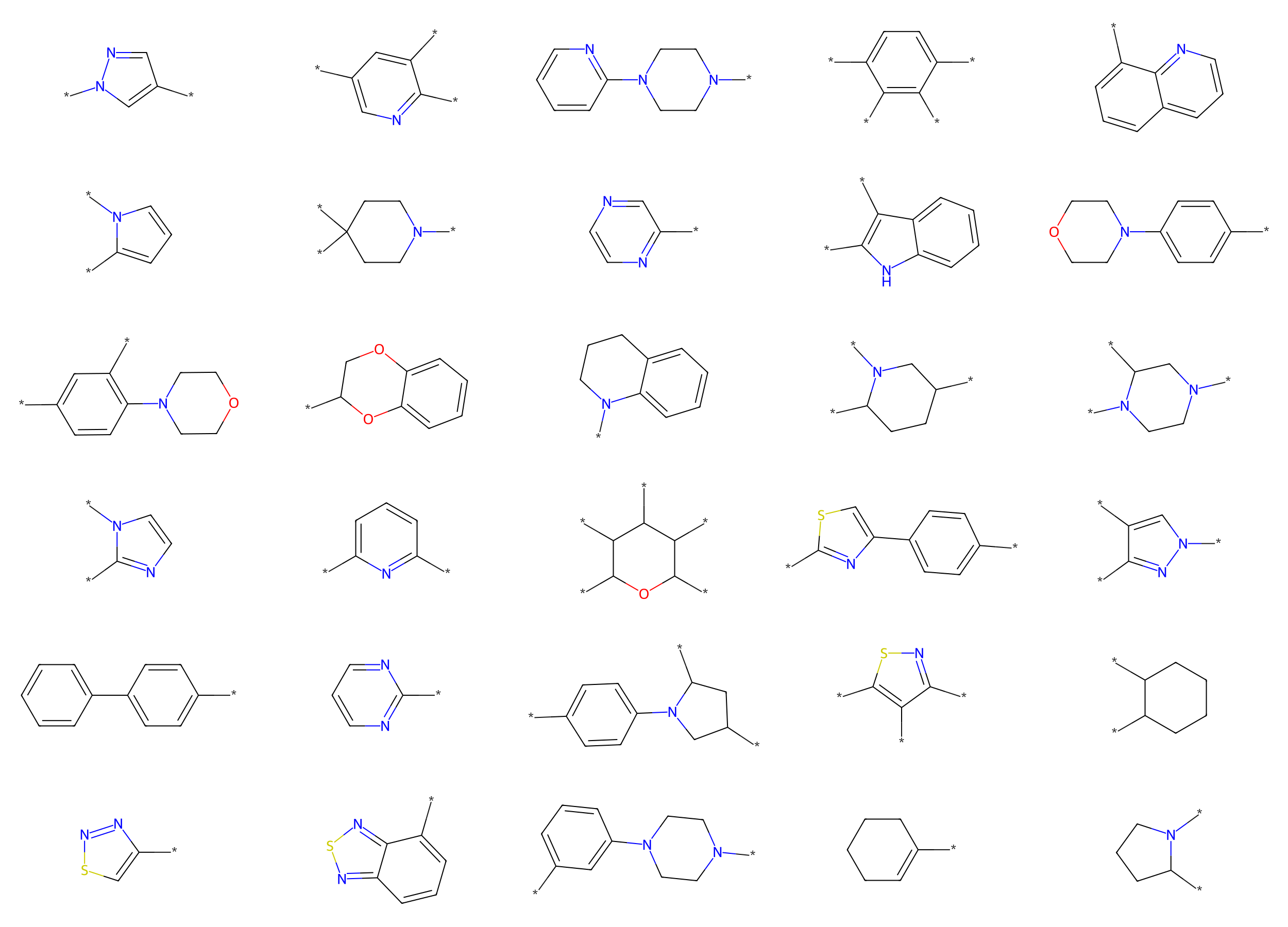}
    \caption{Prefixes used for the prefix completion task (3/3).}
    \label{fig:full_prefixes_3}
\end{figure}

The complete list of prefixes extracted from the GEOM dataset is shown in Figure~\ref{fig:full_prefixes_1}, \ref{fig:full_prefixes_2}, and \ref{fig:full_prefixes_3}.

\subsection{Vector field visualization}\label{subsec:vector_field}
\begin{figure}[htb]
    \centering
    \includegraphics[width=1.0\linewidth]{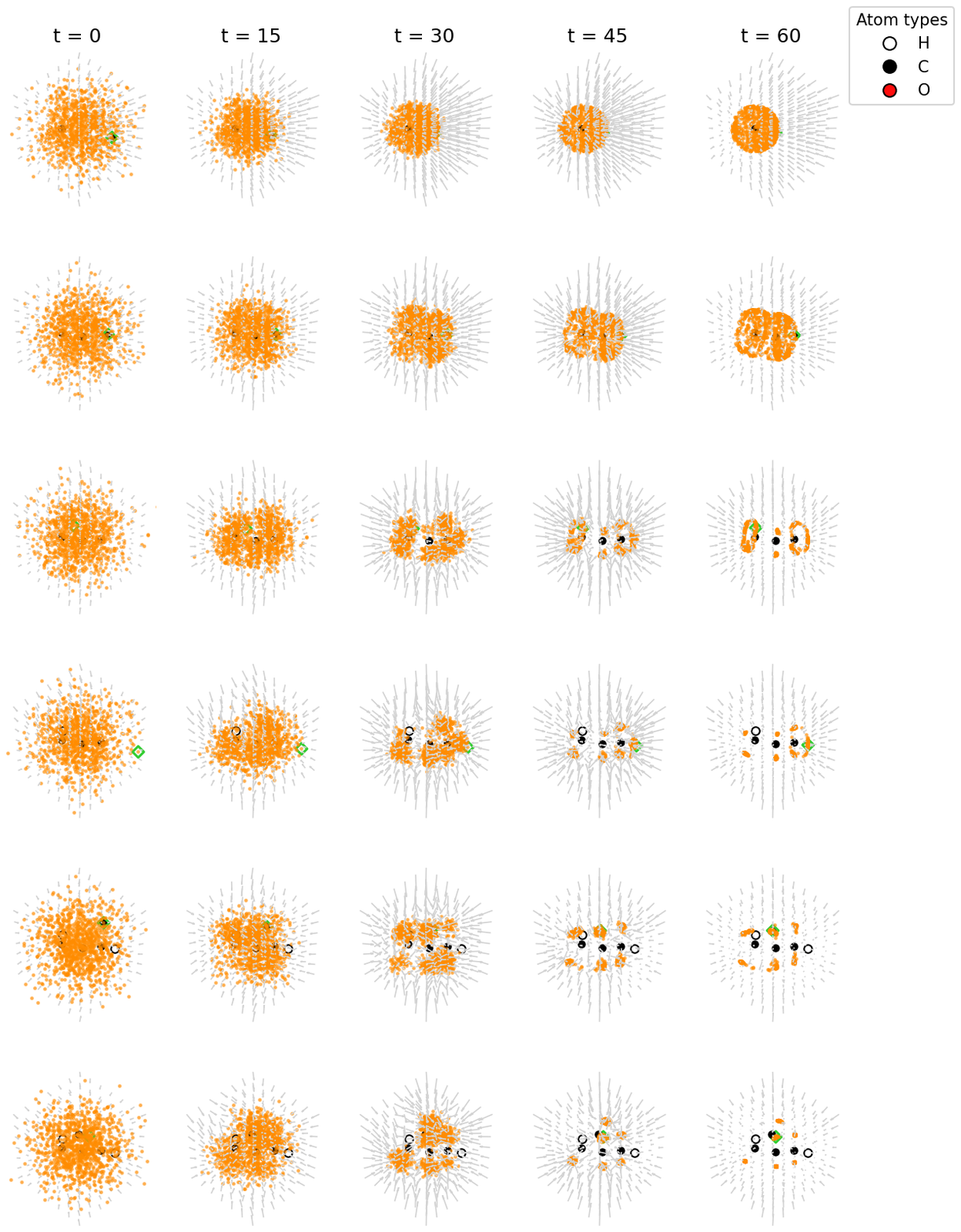}
    \caption{Iterative build of a molecule with NEAT, trained on QM9 (1/3).}
    \label{fig:molecule_build_1}
\end{figure}

\begin{figure}[htb]
    \centering
    \includegraphics[width=1.0\linewidth]{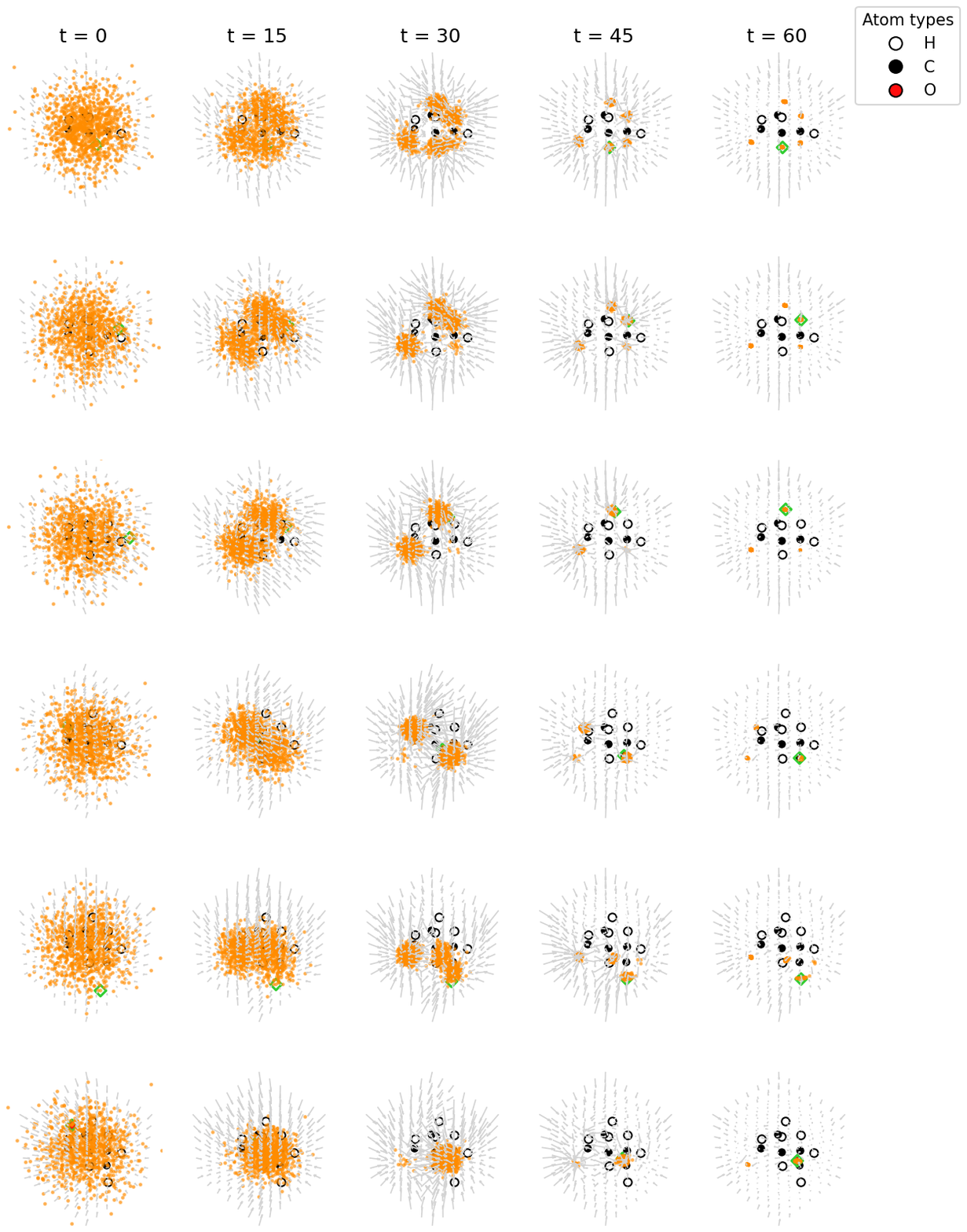}
    \caption{Iterative build of a molecule with NEAT, trained on QM9 (2/3).}
    \label{fig:molecule_build_2}
\end{figure}

\begin{figure}[htb]
    \centering
    \includegraphics[width=1.0\linewidth]{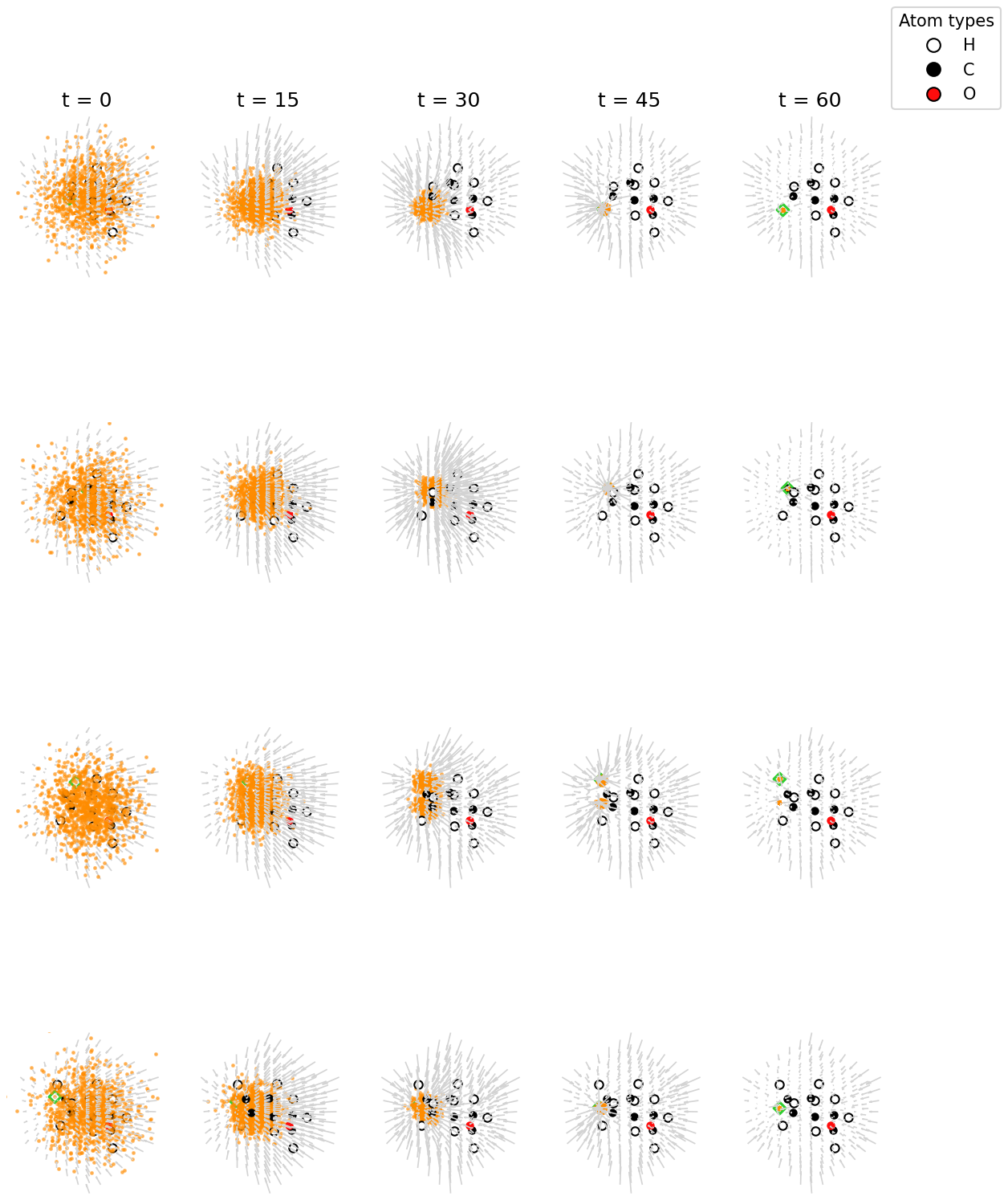}
    \caption{Iterative build of a molecule with NEAT, trained on QM9 (3/3).}
    \label{fig:molecule_build_3}
\end{figure}

Figure~\ref{fig:molecule_build_1}–\ref{fig:molecule_build_3} illustrate iterative molecular construction by the NEAT model trained on QM9. 
Each row corresponds to one construction step. 
From left to right, panels show Gaussian-initialized points (orange) advected by the NEAT velocity field (gray arrows, with length indicating speed). 
Trajectories are integrated using explicit Euler with a uniform time step for $N=60$ steps. 
The green diamond marks the point selected as the atom position for the next iteration. 
We use QM9 for visualization, as examples from the larger GEOM dataset become visually cluttered.

\newpage
\clearpage
\section*{NeurIPS Paper Checklist}

\begin{enumerate}

\item {\bf Claims}
    \item[] Question: Do the main claims made in the abstract and introduction accurately reflect the paper's contributions and scope?
    \item[] Answer: \answerYes{} 
    \item[] Justification: Our main contributions are elucidated in the abstract and the introduction.

\item {\bf Limitations}
    \item[] Question: Does the paper discuss the limitations of the work performed by the authors?
    \item[] Answer: \answerYes{} 
    \item[] Justification: The limitations of our work are discussed in the main text in a dedicated paragraph and further elaborated in the Appendix.

\item {\bf Theory assumptions and proofs}
    \item[] Question: For each theoretical result, does the paper provide the full set of assumptions and a complete (and correct) proof?
    \item[] Answer: \answerNA{} 
    \item[] Justification: The paper does not include theoretical results.

\item {\bf Experimental result reproducibility}
    \item[] Question: Does the paper fully disclose all the information needed to reproduce the main experimental results of the paper to the extent that it affects the main claims and/or conclusions of the paper (regardless of whether the code and data are provided or not)?
    \item[] Answer: \answerYes{} 
    \item[] Justification: The pseudo-code for our proposed method and a detailed description of the model architecture are given in the "Methods" section of the paper. The experimental setting is described in the "Experiments" section of the paper. Additional information on the model architecture and the experimental setting are provided in the "Implementation details" section of the Appendix. Additionally, the submission includes an anonymized link to our code repository and pretrained model weights. We will include a de-anonymized link to the official code repository upon acceptance. All datasets used are publicly available. We trust that the provided information is complete and sufficient to reproduce all results presented in this work.

\item {\bf Open access to data and code}
    \item[] Question: Does the paper provide open access to the data and code, with sufficient instructions to faithfully reproduce the main experimental results, as described in supplemental material?
    \item[] Answer: \answerYes{} 
    \item[] Justification: This submission includes an anonymized link to our code repository, including pretrained model weights. We will include a de-anonymized link to the official code repository upon acceptance. Our repository includes scripts for data and model download, data preprocessing, model training, model inference, and evaluation of the results. System specification are detailed in the Appendix.

\item {\bf Experimental setting/details}
    \item[] Question: Does the paper specify all the training and test details (e.g., data splits, hyperparameters, how they were chosen, type of optimizer) necessary to understand the results?
    \item[] Answer: \answerYes{} 
    \item[] Justification: We provide comprehensive information on the experimental setting in the "Experiments" section of the core of the paper and present detailed information in the "Implementation details" section of the Appendix. We trust that the provided information, together with the released code and model weights, is complete and sufficient to reproduce all results presented in this work.

\item {\bf Experiment statistical significance}
    \item[] Question: Does the paper report error bars suitably and correctly defined or other appropriate information about the statistical significance of the experiments?
    \item[] Answer: \answerYes{} 
    \item[] Justification: In our performance evaluation, we report the mean and 1-sigma standard deviation interval over three inference runs with different random seed initialization. When the source code and trained model weights were available, we re-evaluated baselines using our evaluation pipeline to maximize compatibility. All this is stated in the paper.

\item {\bf Experiments compute resources}
    \item[] Question: For each experiment, does the paper provide sufficient information on the computer resources (type of compute workers, memory, time of execution) needed to reproduce the experiments?
    \item[] Answer: \answerYes{} 
    \item[] Justification: The paper specifies the hardware used for training and inference, including GPU models, memory, and processors. It also specified the version of the software and libraries employed. The size, computational cost, and runtime of the proposed model are discussed in the main part of the paper. 
    
\item {\bf Code of ethics}
    \item[] Question: Does the research conducted in the paper conform, in every respect, with the NeurIPS Code of Ethics \url{https://neurips.cc/public/EthicsGuidelines}?
    \item[] Answer: \answerYes{} 
    \item[] Justification: The research did not involve human subjects or participants. The datasets do not contain personally identifiable information and are still available for use. The QM9 and GEOM-Drugs datasets that were used in this work have defined licensing agreements, which we respected. We discussed the societal impact and potential harmful consequences of our work in the "Broader societal impact" section of the Appendix.

\item {\bf Broader impacts}
    \item[] Question: Does the paper discuss both potential positive societal impacts and negative societal impacts of the work performed?
    \item[] Answer: \answerYes{} 
    \item[] Justification: We discussed the societal impact and potential harmful consequences of our work and proposed adequate mitigation measures in the "Broader societal impact" section of the Appendix.
    
\item {\bf Safeguards}
    \item[] Question: Does the paper describe safeguards that have been put in place for responsible release of data or models that have a high risk for misuse (e.g., pre-trained language models, image generators, or scraped datasets)?
    \item[] Answer: \answerNA{} 
    \item[] Justification: We do not foresee high risk of misuse for our model.

\item {\bf Licenses for existing assets}
    \item[] Question: Are the creators or original owners of assets (e.g., code, data, models), used in the paper, properly credited and are the license and terms of use explicitly mentioned and properly respected?
    \item[] Answer: \answerYes{} 
    \item[] Justification: The original owners of the QM9 and GEOM-Drugs datasets used in this work are properly referenced and credited and the licenses of both datasets are explicitly mentioned in the "Data processing" section of the Appendix. We do not repackage the datasets.

\item {\bf New assets}
    \item[] Question: Are new assets introduced in the paper well documented and is the documentation provided alongside the assets?
    \item[] Answer: \answerYes{} 
    \item[] Justification: The paper introduces new assets in the form of code and trained model weights. We provide an anonymized links to our code repository and pretrained model weights. We will include a de-anonymized link to the official code repository / model weights upon acceptance.

\item {\bf Crowdsourcing and research with human subjects}
    \item[] Question: For crowdsourcing experiments and research with human subjects, does the paper include the full text of instructions given to participants and screenshots, if applicable, as well as details about compensation (if any)? 
    \item[] Answer: \answerNA{} 
    \item[] Justification: This work does not involve crowdsourcing nor research with human subjects.

\item {\bf Institutional review board (IRB) approvals or equivalent for research with human subjects}
    \item[] Question: Does the paper describe potential risks incurred by study participants, whether such risks were disclosed to the subjects, and whether Institutional Review Board (IRB) approvals (or an equivalent approval/review based on the requirements of your country or institution) were obtained?
    \item[] Answer: \answerNA{} 
    \item[] Justification: This work does not involve crowdsourcing nor research with human subjects.

\item {\bf Declaration of LLM usage}
    \item[] Question: Does the paper describe the usage of LLMs if it is an important, original, or non-standard component of the core methods in this research? Note that if the LLM is used only for writing, editing, or formatting purposes and does \emph{not} impact the core methodology, scientific rigor, or originality of the research, declaration is not required.
    \item[] Answer: \answerNA{} 
    \item[] Justification: LLMs were used only for spell-checking, editing of self-written text, formatting plots and implementing standard elements of code. They were not involved in the development of the core methods presented in this work.

\end{enumerate}

\end{document}